%% file: main_neurips2026_arxiv.tex
\documentclass{article}

\usepackage[preprint]{neurips_2026}
\input{math_commands.tex}

\usepackage[utf8]{inputenc} 
\usepackage[T1]{fontenc}    
\usepackage{hyperref}       
\usepackage{url}            
\usepackage{booktabs}       
\usepackage{amsfonts}       
\usepackage{nicefrac}       
\usepackage{microtype}      
\usepackage{xcolor}         
\usepackage{multirow}       
\usepackage{makecell}
\usepackage{graphicx}
\usepackage{enumitem}

\usepackage{cleveref}
\crefname{figure}{Fig.}{Figs.}
\Crefname{figure}{Fig.}{Figs.} 

\usepackage{titletoc}
\usepackage{hyperref}             
\usepackage{etoc}
\definecolor{rosy}{RGB}{220,20,120 }
\definecolor{NavyBlue}{RGB}{68,139,180}
\definecolor{ForestGreen}{RGB}{74,136,91}
\hypersetup{colorlinks=true, allcolors=rosy}

\usepackage{titlesec}
\titlespacing*{\paragraph}
  {0pt}    
  {1ex plus .1ex minus .2ex}  
  {1em}    
\usepackage[autostyle, english=american]{csquotes}
\MakeOuterQuote{"}   

\newcommand{\x}{\mathbf{x}}

\newcommand{\todo}[1]{\textcolor{magenta}{(todo: #1)}}

\newcommand{\qinv}[1]{\textcolor{orange}{#1}}
\newcommand{\rinv}[1]{\textcolor{red}{#1}}
\newcommand{\nov}[1]{\textcolor{ForestGreen}{#1}}
\newcommand{\mem}[1]{\textcolor{violet}{#1}}

\title{The two clocks and the innovation window:\\When and how generative models learn rules}

\author{
  Binxu Wang
  \thanks{Corresponding to \texttt{binxu\_wang@hms.harvard.edu}.}
  \ \thanks{An earlier version of this work was presented as an oral presentation at the NeurIPS 2025 SPIGM Workshop, \url{https://openreview.net/forum?id=LjqX8OhPPi}}
  \qquad
  Emma Lucia Byrnes Finn
  \qquad
  Bingbin Liu
  \\
  \texttt{binxu\_wang@hms.harvard.edu} \qquad\texttt{efinn@college.harvard.edu}
  \qquad\texttt{bliu@g.harvard.edu} \\
  Kempner Institute at Harvard University
}

\begin{document}

\maketitle

\begin{abstract}
Generative models trained on finite data face a fundamental tension: their score-matching or next-token objective converges to the empirical training distribution rather than the population distribution we seek to learn.
Using rule-valid synthetic tasks, we trace this tension across two training timescales: $\trule$, the step at which generations first become rule-valid, and $\tmem$, the step at which models begin reproducing training samples.
Focusing on parity and extending to other binary rules and combinatorial puzzles,
we characterize how these two clocks $\trule$, $\tmem$ depend on key aspects of the learning setup.
Specifically, we show that $\trule$ increases with rule complexity and decreases with model capacity, while $\tmem$ is approximately invariant to the rule and scales nearly linearly with dataset size $N$.
We define the \emph{innovation window} as the interval $[\trule, \tmem]$.
This window widens with increasing $N$ and narrows with rule complexity, and may vanish entirely when $\trule \geq \tmem$.
The same two-clock structure arises in both diffusion (DiT) and autoregressive (GPT) models, with architecture-dependent offsets.
Dissecting the learned score of DiT models reveals a corresponding evolution of the optimization landscapes, where attractors emerge at both timescales: rule-valid samples' basins expand substantially around $\trule$, while training samples' basins begin to dominate around $\tmem$.
Together, these results yield a unified and predictive account of when and how generative models exhibit genuine innovation.
\end{abstract}

\section{Motivation}

Modern generative models produce strikingly realistic images, audio, and text by learning from finite data. Yet realism alone does not tell us whether a model has learned the latent rules defining a distribution. Memorization can be tested by comparing against the training data. But the stronger question---whether a model has learned the true data-generating distribution---is much harder to answer in natural data. 
Therefore, we turn to synthetic distributions governed by latent rules, where rule satisfaction, novelty, and memorization can be measured precisely. 

Our main testbed is group-level parity on binary images: each image is divided into groups of size $G$, and each group must satisfy an even-parity constraint. Increasing $G$ increases the interaction order among bits, giving a direct knob for rule complexity. Because both the full rule-valid set and the finite training set are known, we can separately quantify rule accuracy, novelty, and exact memorization.

This setup lets us ask: when do generative models learn the rule, and when do they begin to reproduce training samples? Tracking training dynamics reveals two separable timescales. We define $\trule$ as the step at which generations first become rule-valid, and $\tmem$ as the step at which they begin to match training samples. These are the two \emph{clocks} of generative training. 
Their relative timing defines an \emph{innovation window}: the interval $[\trule,\tmem]$ during which generations are rule-valid but non-memorized. 

We next ask what controls these two timescales (Sec.~\ref{sec:two_clocks}). Rule complexity $G$ strongly delays $\trule$; dataset size $N$ scales $\tmem$ nearly linearly; and model capacity shifts both clocks. Thus, increasing $G$ can close the innovation window by pushing rule learning past memorization, whereas increasing $N$ can reopen it by postponing memorization.

We further dissect diffusion training to understand how these clocks arise (Sec.~\ref{sec:dissect}). By tracking fixed initial noises, evaluating denoising losses across noise scales, and visualizing learned vector fields on slices of the Boolean hypercube, we find that rule learning and memorization correspond to distinct changes in the attractor landscape. Early in training, the model learns attractors at Boolean vertices. Around $\trule$, rule-valid basins expand and suppress rule-violating corners. Around $\tmem$, training-sample basins expand further and invade nearby held-out valid regions.

Finally, we show that the two-clock structure extends beyond diffusion models and parity. Autoregressive models (e.g., GPT) also exhibit the two $\trule$,$\tmem$ timescales with their distinct scaling behaviors (Sec.~\ref{sec:gpt}). Other rule families, including count-based rules and categorical puzzles (e.g., Sudoku) show similar dynamics (Sec.~\ref{sec:beyond_parity}). 
Together, these results provide a dynamical account of when and how generative models generalize, memorize, and genuinely innovate.

\vspace{-0.5em}
\section{Generative modeling of parity: rule learning and memorization}
\label{sec:gen_model}

\begin{figure}[!tp]
    \centering
    \vspace{-13pt}
    \includegraphics[width=\textwidth]{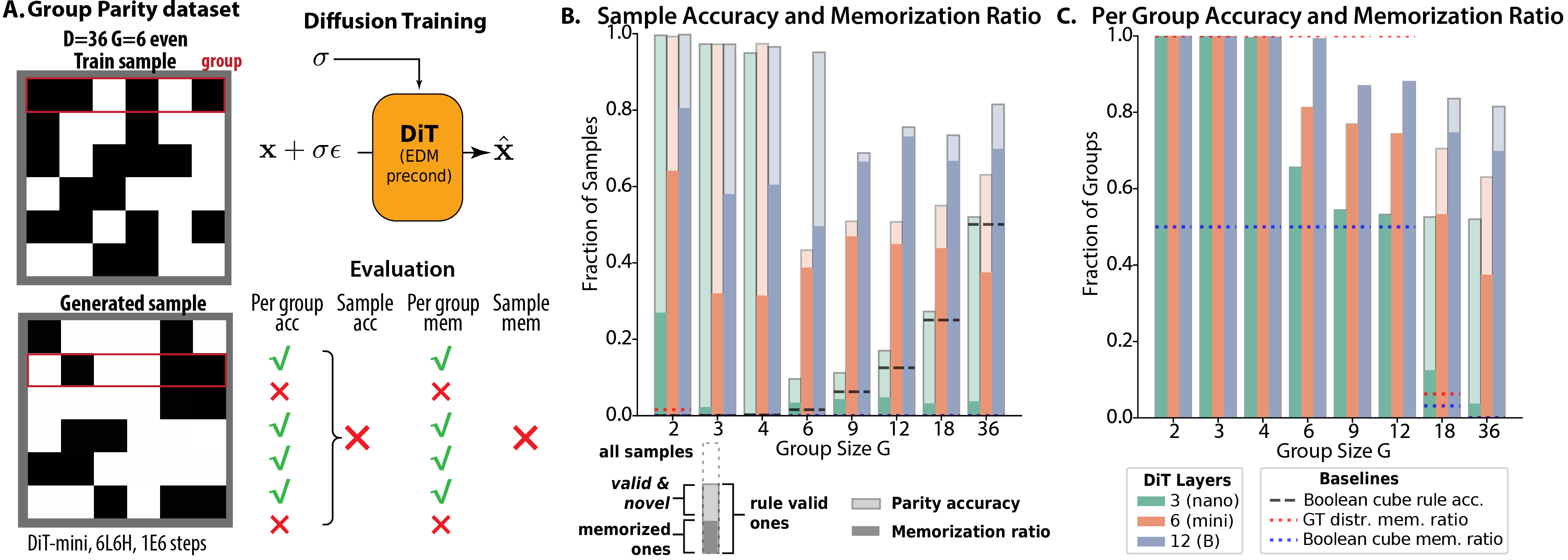}
    \vspace{-13pt}
    \caption{
    \textbf{Rule learning and memorization for group-parity.}
    \textbf{A.} Structure of the Group Parity dataset and evaluation setup. Each $D$-dim binary image is divided into groups of size $G$ (here $D=36$, $G=6$), with each \emph{group} satisfying even parity.
    We train DiT models on datasets of $N=4096$ samples and evaluate model generations in term of accuracy and memorization ratio, at both (\textbf{B}) sample and (\textbf{C}) group level, across different model sizes. 
    In \textbf{B}-\textbf{C}, pale bars denote rule accuracy, and dark bars denote the portion of generations that are seen during training (i.e., memorized and valid).
    The black dashed lines mark the random baseline accuracy, 
    while blue and red dashed lines mark the baseline memorization ratios for the Boolean cube and ground-truth distribution.
}
    \vspace{-10pt}
    \label{fig:dataset_eval_model_cmp}
\end{figure}

\vspace{-0.5em}
We focus on group-level parity as a case study; extensions to other tasks are discussed in \Cref{sec:beyond_parity}.

\paragraph{Group-level parity}
Define $\mathcal{S}^+_d=\{\x\mid \prod^d_{i=1}x_i=1,x_i\in\{-1,1\}\}$ to be the set of points with \textit{even parity} on a $d$-dimensional boolean hypercube $\{-1,1\}^d$ (\Cref{fig:dataset_eval_model_cmp}\textbf{A}).
For example, $\mathcal{S}^+_3=\{(1,1,1),(-1,-1,1),(-1,1,-1),(1,-1,-1)\}$;
note that $|\mathcal{S}^+_d|=2^{d-1}$.
Define $\mathcal{P}_d^+(\x)=|\mathcal{S}^+_d|^{-1}\sum_{\mathbf{y}\in\mathcal{S}^+_d }\delta(\x-\mathbf{y})$ to be the mixture of delta measures at all points of the set $\mathcal{S}^+_d$.
The \emph{group-level parity} is defined as $(\mathcal{S}^+_{d})^m=\mathcal{S}^+_d\times...\times\mathcal{S}^+_d\subset \{-1,1\}^{md}$, where the $d$ bits in each of the $m$ groups satisfy even parity, with $|(\mathcal{S}^+_{d})^m|=(2^{d-1})^m$.
Denote $(\mathcal{P}^+_{d})^m$ the uniform measure over the rule-valid set $(\mathcal{S}^+_{d})^m$,
and define $\mathcal U_d$ as the uniform measure on $d$-dimensional Boolean hypercube.

\paragraph{Dataset design}
We construct samples $\x\in\mathbb R^D$ with $D=36$.
Each $\x$ is divided into $D/G$ groups of size $G$ that each satisfies even parity, i.e., $\x\in (\mathcal{S}^+_{G})^{D/G}$.
To generate $\x$, we first sample each group i.i.d. from
$\sim\mathcal{P}_G^+$, and then concatenate the $D/G$ groups to form $\x$ (Fig.~\ref{fig:dataset_eval_model_cmp}\textbf{A}).
The training set contains $N$ unique samples obtained by rejection sampling, while individual groups may repeat across samples.
The key design parameter for the dataset are hence $D,G,N$.
For diffusion training, we reshape each sample to a $6\times 6$ single channel image.

\paragraph{Model architecture}
We use Gaussian diffusion models and treat the dataset as lying in the continuous space $\mathbb R^D$.
Specifically, we use the continuous-time EDM framework \citep{karras2022elucidatingDesignSp}, $x_0$ prediction, and use diffusion transformer (DiT) \citep{peebles_scalable_2023} as the function approximator with EDM preconditioning.
The baseline version is \texttt{DiT-mini} with 6 layers, 6 heads and hidden dimension 384, and we vary the model depth (in $\{3,6,12\}$) to examine the effect of model capacity.
We use patch size 1 to maximize the capacity for modeling relation between bits.
The DiTs are trained with Adam for $10^6$ steps, without weight decay, with a constant learning rate $10^{-4}$ and batch size 256. 
Training details are deferred to App.~\ref{app:train_model_method}.

\paragraph{Evaluation} 
Throughout training, we periodically generate samples using Heun's second-order deterministic sampler~\citep{karras2022elucidatingDesignSp} and evaluate quantization, rule satisfaction, and memorization (details in App.~\ref{app:eval_method}).
We first measure proximity to the Boolean cube by $d_{\ell_\infty} (\x)=\max_i||x_i|-1|$, 
and call a sample invalid (quantization error) at threshold $\epsilon$ if $d_{\ell_\infty}(\x)>\epsilon$. 
After binarizing each element to $\{-1,1\}$, we compute parity \textit{accuracy} at both the group and sample levels (Fig.~\ref{fig:dataset_eval_model_cmp}\textbf{A}).
Group-level accuracy has chance level $1/2$, while sample-level accuracy requires all $D/G$ groups to be correct and therefore has chance level $2^{-D/G}$ (Fig.~\ref{fig:dataset_eval_model_cmp}\textbf{B}, black dashed). 

We also report group- and sample-level \textit{memorization ratios}, defined as the fractions of generated groups or full samples that exactly match those in the training set.
If the model samples uniformly from the true parity-constrained distribution $(\mathcal{S}^+_G)^{D/G}$, the expected sample-level memorization ratio is $N/(2^{G-1})^{D/G} = N\cdot 2^{-\frac{G-1}{G}D}$. 
If it samples uniformly from the full Boolean cube, the corresponding baseline is $N\cdot 2^{-D}$.

\paragraph{General performance}
Using these metrics, we find that DiTs learn the parity-constrained distribution well only for small $G$.
At the sample level, all models achieve near-perfect rule accuracy for easy rules $G\leq4$, but the accuracy deteriorates as $G$ increases; larger models extend this learnable regime but do not remove the degradation (Fig.~\ref{fig:dataset_eval_model_cmp}\textbf{B}).
Group-level evaluation shows the same trend more directly: parity accuracy remains near perfect for small $G$, but approaches the chance level $0.5$ for larger $G$, especially in smaller models (Fig.~\ref{fig:dataset_eval_model_cmp}\textbf{C}).
Meanwhile, memorization takes up an increasing fraction of the rule-valid generations at larger $G$, indicating that apparent performance increasingly reflects exact retrieval of training examples rather than rule generalization.

\paragraph{Combinatorial creativity}
We find little evidence for novelty at the group level.
For $G \leq 12$, most rule-valid generated groups are memorized training groups; for $G>12$, the residual gap between group accuracy and group memorization is largely explained by chance parity satisfaction ($1/2$; Fig.~\ref{fig:dataset_eval_model_cmp}\textbf{C}).
Thus, novel valid samples arise primarily by recombining memorized groups into unseen full samples, reflecting a form of \textit{combinatorial creativity}.

\section{Two clocks of generative training}
\label{sec:two_clocks}
The above analysis, performed at the end of training,
shows that increasing the rule complexity $G$ shifts DiT generations from novel generalization toward memorization-dominated performance. 
But this static view does not explain \emph{when} rule learning and memorization emerge during training, or \emph{why} some settings produce novel valid samples while others do not. 
We therefore track rule accuracy and memorization over training and find two distinct timescales: $\trule$, when generated samples first become rule-valid, and $\tmem$, when they begin to recall training samples. 
We call these the two \textit{clocks} of generative training.
Their gap defines an ``innovation window'', in which the model generates rule-valid but non-memorized samples.


\begin{figure}[!tp]
    \centering
    \vspace{-16pt}
    \includegraphics[width=\linewidth]{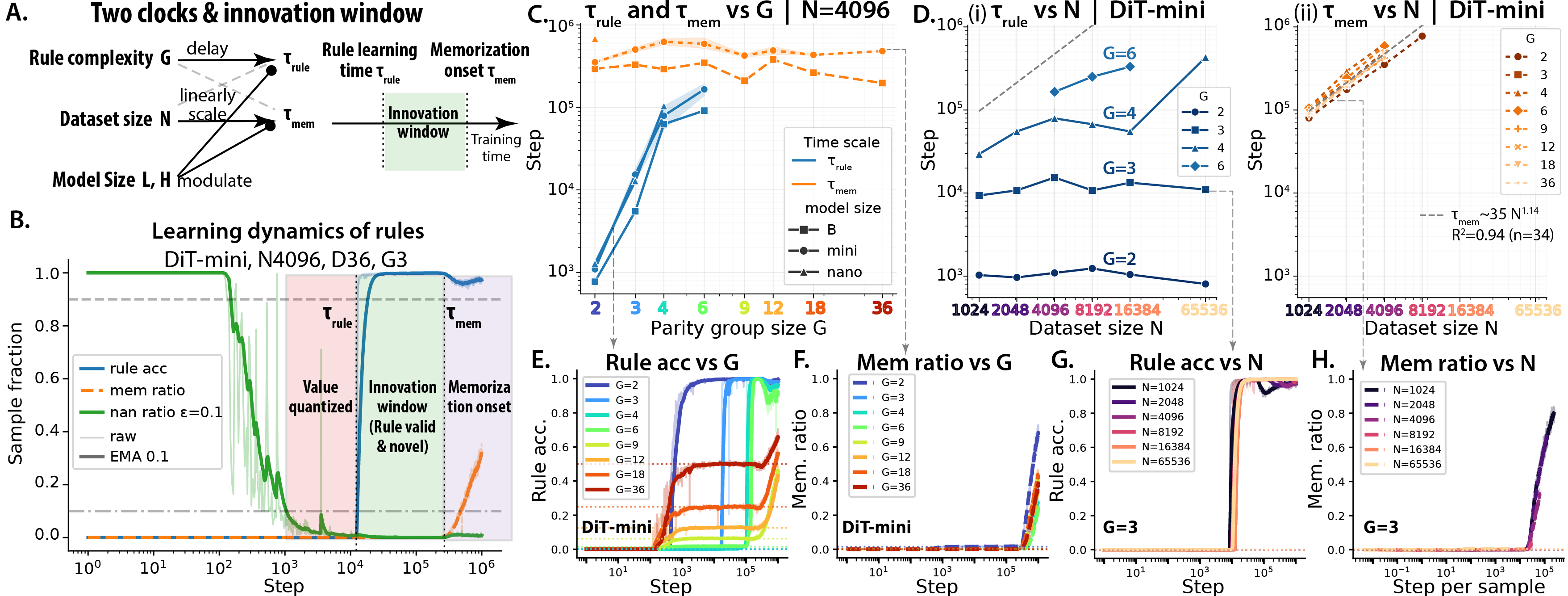}
    \vspace{-12pt}
    \caption{\textbf{Two clocks $\trule$ and $\tmem$ scale differently with rule, data, and model.}
    \textbf{A.}~Schematic: rule complexity $G$ delays $\trule$, dataset size $N$ scales $\tmem$ near-linearly, model size $(L,H)$ modulates both. The interval $[\trule, \tmem]$ is the \emph{innovation window} of valid novel generation.
    \textbf{B.}~Single-run learning dynamics (DiT-mini, $N{=}4096$, $G{=}3$): rule accuracy (blue), memorization ratio (orange dashed), and NaN/quantization error ratio with $\epsilon{=}0.1$ (green); EMA over raw traces. 
    Vertical dotted lines: $\trule$ (first step rule-acc $>0.9$) and $\tmem$ (first step mem-ratio $>0.1$). 
    Shaded phases: value quantization, innovation window, memorization onset.
    \textbf{C.}~$\trule$ (blue) and $\tmem$ (orange dashed) vs.~$G$ at $N{=}4096$ for three model sizes (B, mini, nano); $\trule$ shown only for learnable $G$, i.e., $G \leq 6$.
    \textbf{D.}~$(\textit{i})$ $\trule$ vs.~$N$ for $G \in \{2,3,4,6\}$. $(\textit{ii})$ $\tmem$ vs.~$N$ across all $G$, with power-law fit $\tmem \approx 35\,N^{1.14}$ ($R^2{=}0.94$, $n{=}34$).
    \textbf{E,F.}~Rule accuracy and memorization ratio vs.~step (DiT-mini, $N{=}4096$), one curve per $G$.
    \textbf{G,H.}~Rule accuracy vs.~step and memorization ratio vs.~steps-per-sample (i.e. step $\times$ batch size / $N$), one curve per $N$ (DiT-mini, $G{=}3$).}
    \vspace{-8pt}
    \label{fig:two_clocks_and_scaling}
\end{figure}

\paragraph{Two clocks in one run}
Let's follow a training run with DiT-mini, $G{=}3$, $N{=}4096$
(Fig.~\ref{fig:two_clocks_and_scaling}\textbf{B}). 
First, the invalid-value fraction rapidly decreases, indicating that the model has learned to sample from the Boolean cube.
Next, the rule-learning transition $\trule$ occurs, marked by a sharp rise in sample-level rule accuracy from chance to near-perfect performance. 
Memorization remains near baseline level at this point and
starts climbing around the much later $\tmem$, accompanied by a small uptick in the fraction of invalid samples and a small dip in rule accuracy. 
We operationally define $\trule$ as the first step at which sample-level rule accuracy reliably exceeds $0.9$, and $\tmem$ as the first step at which sample memorization fraction reliably exceeds $0.1$.
The interval $[\trule, \tmem]$ is referred to as the \emph{innovation window}, during which the model generations are rule-valid yet novel.

This temporal separation reinforces prior observations that early stopping preserves the generalizing solution before memorization dominates~\citep{bonnaire2025memorize};
our contribution is to systematically study $\trule$ and $\tmem$ as a function of rule complexity $G$, dataset size $N$, and model capacity (Fig.~\ref{fig:two_clocks_and_scaling}\textbf{A}), which we discuss in the following.

\paragraph{Determinants of rule-learning time \texorpdfstring{$\trule$}{}}

We find that \textit{rule complexity monotonically delays $\trule$}.
For small $G$, where the rule-learning transition occurs reliably, $\trule$ increases from approximately $10^3$ steps at $G=2$ to $10^4$ at $G=3$ and $10^5$ at $G=4$; thus, each unit increase in $G$ delays rule learning by roughly an order of magnitude (Fig.~\ref{fig:two_clocks_and_scaling}\textbf{C}, blue).
At $G=6$, the transition becomes unreliable, likely because $\trule$ approaches $\tmem$ and the innovation window begins to close. 
For all model sizes, rules with $G \geq 9$ are not acquired within $10^6$ steps, and their final accuracy can be fully attributed to the later memorization process, which matches the known difficulty of learning high-degree parity~\citep{Kearns98,barak2022hidden,abbe2024generalization}. 
In contrast, \textit{$\trule$ is only mildly modulated by dataset size $N$}: increasing $N$ from $1024$ to $65536$ leads to indistinguishable rule learning curves at $G=2,3$ and mildly increases $\trule$ for $G=4,6$ (Fig.~\ref{fig:two_clocks_and_scaling}\textbf{D}\textit{(i)}, \textbf{G}). 
Finally, \textit{increasing model capacity accelerates rule learning $\trule$ but does not remove the complexity barrier}
\footnote{See \Cref{sec:related_work} for more discussions on related work.}. 
Given a fixed $G$, larger DiTs have an earlier rule transition $\trule$ (Fig.~\ref{fig:two_clocks_and_scaling}\textbf{C}, comparing DiT-B/mini/nano). 
However, even the largest 12-layer model we test fails at $G\geq9$, showing the learnability frontier. 

\paragraph{Determinants of memorization time \texorpdfstring{$\tmem$}{}}

In contrast to $\trule$, \textit{$\tmem$ is approximately invariant to the rule complexity $G$}:
memorization occurs after a similar number of optimization steps regardless of the rule being acquired or not (Fig.~\ref{fig:two_clocks_and_scaling}\textbf{C}, orange; Fig.~\ref{fig:two_clocks_and_scaling}\textbf{F}). 
Rather, there is a \textit{clear scaling of $\tmem$ in the training set size $N$:} increasing $N$ shifts $\tmem$ nearly linearly, with empirical fit $\tmem \approx 35\,N^{1.14}$ ($R^2 = 0.94$, $n = 34$ across all $G$ at DiT-mini scale; Fig.~\ref{fig:two_clocks_and_scaling}\textbf{D}\textit{(ii)}; see App.~\ref{app:tau_mem_scaling_extended} for scaling per $G$ and model).
Consistently, when plotted against \emph{steps per training example} (step $\times$ batch / $N$), the memorization trajectories across $N$ collapse onto a common curve (Fig.~\ref{fig:two_clocks_and_scaling}\textbf{H}, c.f. \cite{bonnaire2025memorize}), suggesting that memorization is governed by \textit{gradient updates per example}, with onset at around $10^4$ steps-per-example in our setup. 
Finally, increasing model capacity also accelerates $\tmem$ (Fig.~\ref{fig:two_clocks_and_scaling}\textbf{C}, orange), with DiT-nano having nearly no memorization within $10^6$ steps. 
This earlier onset also leads to higher memorization ratio, and higher memorization-driven accuracy at final evaluation (Fig.~\ref{fig:dataset_eval_model_cmp}\textbf{B,C} across depths). 

\paragraph{The innovation window}
These scaling relationships (Fig.~\ref{fig:two_clocks_and_scaling}\textbf{A}) together provide one explanation of why models have a hard time learning higher $G$ rules: at a fixed model size and $N$, 
higher $G$ significantly scales up $\trule$ while keeping $\tmem$ stable; thus, extrapolating this trend, the hypothetical $\trule$ for higher $G$ will hit the $\tmem$ ceiling, resulting in models only reaching the memorization solution and closing the innovation window (Fig. \ref{fig:two_clocks_and_scaling}\textbf{C}). 
In contrast, increasing dataset size $N$ delays $\tmem$ and opens up the innovation window for learning harder rules: for example, $G=6$ is not learned at $N=1024,2048$ as memorization happens first, but it is learned for larger $N=4k,8k,16k$ as $\tmem$ ceiling goes up (Fig. \ref{fig:two_clocks_and_scaling}\textbf{D}(i)). 
Further, a higher model capacity accelerates both $\trule$ and $\tmem$, thus not necessarily increasing the innovation window (Fig. \ref{fig:two_clocks_and_scaling}\textbf{C}). 

%


\section{Dissecting rule learning and memorization dynamics}
\label{sec:dissect}

\begin{figure}[!tp]
    \centering
        \vspace{-20pt}
\includegraphics[width=\textwidth]{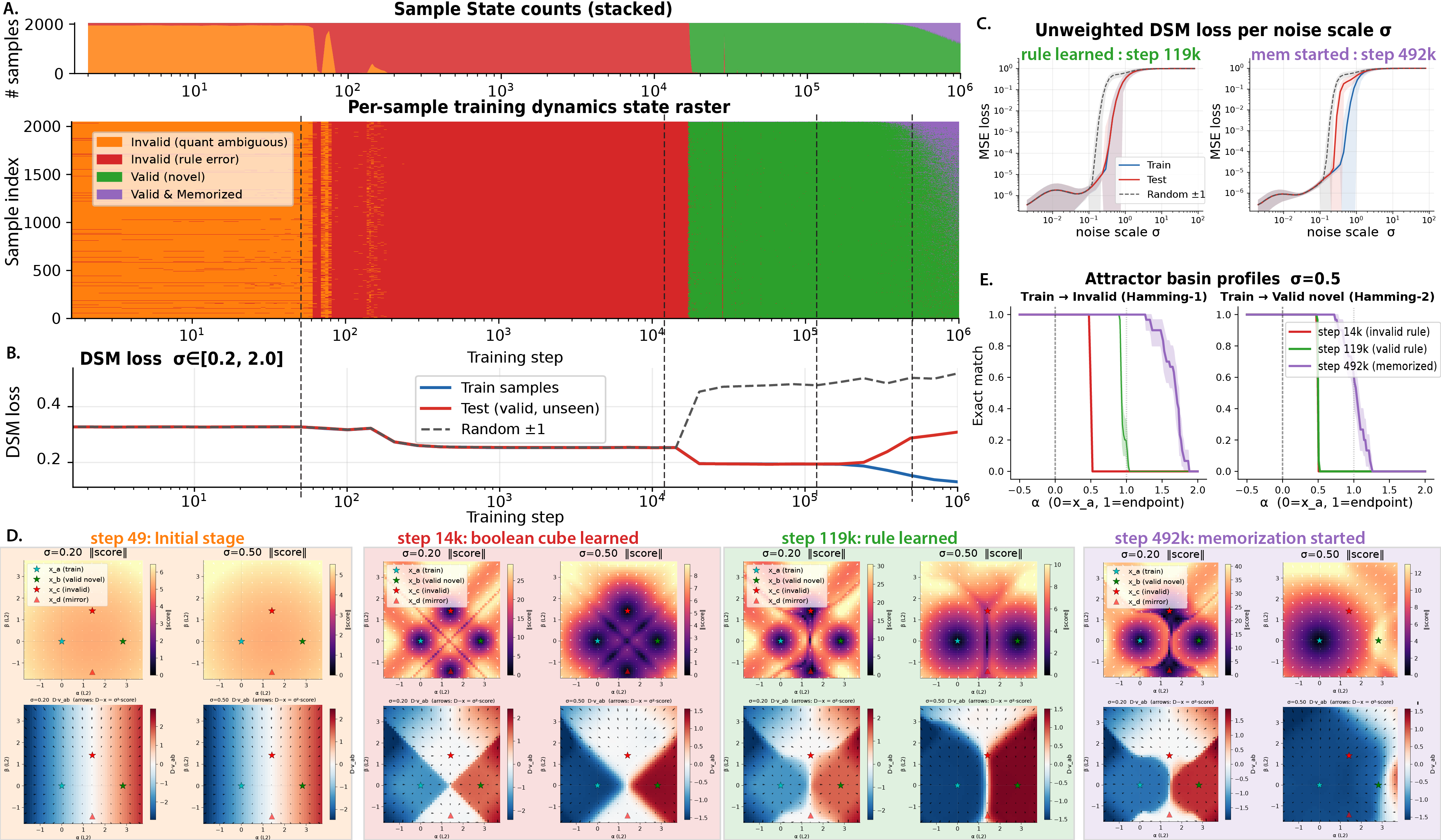}
        \vspace{-12pt}
        \caption{
        \textbf{Dissecting rule learning and memorization dynamics via multiple perspectives}.
        \textbf{A.}~\emph{Per-sample training-dynamics state raster} for 2000 fixed initial noises. Samples are classified into four states across training steps: \textcolor{orange}{invalid (quantization-ambiguous)}, \textcolor{red}{invalid (rule-violating)}, \nov{valid \& novel}, and \textcolor{violet}{valid \& memorized},
        shown as (\textit{Top}) counts or (\textit{Bottom}) per-seed evolution throughout training. The transition matrices are showed in Fig.~\ref{suppfig:state_transition_markov}.
        \textbf{B.}~\emph{DSM loss averaged over $\sigma\in[0.2,2.0]$}, evaluated on training samples (blue), held-out rule-valid samples (red), and random Boolean-cube samples (dashed gray).
        \textbf{C.}~\emph{Unweighted DSM loss}, as a function of $\sigma$ at (\textit{left}) the rule-learning step (119k) and (\textit{right}) the memorization-onset step (492k). 
        All views show a sharp, synchronized rule-learning transition and a gradual, stochastic memorization onset.
        To understand why this happens, we provide the
        \textbf{D.}~\emph{vector-field and landscape view}, which shows (\textit{top}) score magnitude $\|s_\theta\|$ and (\textit{bottom}) projected denoising velocity $D{-}v_{ab}$ at $\sigma\in\{0.20,0.50\}$.
        The visualization shows a projection on a 2D slice of the boolean hypercube containing four reference points (\Cref{sec:vec_field}), where the basin of the training sample $x_a$ keeps expanding during training.
        This is further supported by \textbf{E.}~\emph{Attractor basin profiles} at $\sigma{=}0.5$,
        which shows fractions of samples getting denoised to $x_a$ when moving along a linear interpolations from $x_a$ ($\alpha{=}0$) to (\textit{left}) invalid neighbors (i.e., $x_c$ or $x_d$) or (\textit{right}) valid neighbors (i.e., $x_b$):
        $x_a$ and $x_c,x_d$ are equally attractive basins in the landscape before rule learning (step 14k), and such symmetry is broken after rule learning (step 119k).
        $x_a, x_b$ remain equally attractive at this time, but $x_a$ becomes the dominating attractor after memorization onset (step 492k).
        }
        \vspace{-10pt}
    \label{fig:diffusion_rule_mem_dissection}
\end{figure}

\Cref{sec:two_clocks} provides statistical observations for the rule-learning time $\trule$ and memorization time $\tmem$;
but what underlying mechanisms govern them?
This section provides an in-depth analysis from three complementary perspectives: (1) per-sample training dynamics (Fig.~\ref{fig:diffusion_rule_mem_dissection}\textbf{A}), (2) loss objective, (Fig.~\ref{fig:diffusion_rule_mem_dissection}\textbf{B,C}), and (3) the vector field and energy landscape (Fig.~\ref{fig:diffusion_rule_mem_dissection}\textbf{D,E}).

These three views jointly present the following picture.
The rule learning and memorization are associated with a critical noise scale of $\sigma\approx1$ (Fig.~\ref{fig:diffusion_rule_mem_dissection}\textbf{B}).  
First, the model learns to generate on the Boolean hypercube by turning each vertex into an attractor at these noise scales (Fig.~\ref{fig:diffusion_rule_mem_dissection}\textbf{D}, step 14k). 
After rule learning transition, the rule-valid samples expand their attractor basins over the rule-violating ones (Fig.~\ref{fig:diffusion_rule_mem_dissection}\textbf{D}, step 119k).
Finally, after memorization onset, basins of the trained samples expand further and eventually dominate the landscape
(Fig.~\ref{fig:diffusion_rule_mem_dissection}\textbf{D,E}, step 492k).
We next discuss each perspective in detail.

\vspace{-5pt}
\subsection{Per-sample training dynamics of rule learning and memorization}
With a deterministic sampler, diffusion samples are endpoints of probability-flow ODE trajectories that depend continuously on the learned velocity field, and hence on the network parameters $\theta$. 
Thus, the generated samples change continuously across training steps, and we can read out the properties of the underlying landscape from these dynamics.
Starting from fixed initial noises, we track the corresponding generated samples across training steps. 
We classify the generations into four categories: (1) \qinv{invalid values (quantization-ambiguous)}, (2) \rinv{valid values but rule-violating}, (3) \nov{rule-valid and novel}, and (4) \mem{valid and memorized from training set}.

We then visualize the state transition between these four categories as a function of training steps.
As shown in the raster (Fig.~\ref{fig:diffusion_rule_mem_dissection}\textbf{A}, bottom),
the \rinv{rule-violating} to \nov{rule-valid transition} is sharp and synced across all initial seeds, while the later memorization happens more gradually and stochastically.
In the late phase, individual samples consist entirely of valid values, but can flicker between being \nov{rule-valid} or \rinv{rule-violating}, or being \nov{valid-novel} or \mem{valid-memorized}. 
Modeling the Markov transition probability between the four categories in the late phase, we see that the \rinv{rule-violating state} is non-attractive where samples tend to flip back to valid quickly, while the \mem{valid-memorized} state is sticky where samples have a higher probability to stay (Fig.~\ref{suppfig:state_transition_markov}). 

During late training ($10^5$--$10^6$ steps), generated samples continue to move closer to the training set in nearest-neighbor Hamming distance, even before $\tmem$ and even when restricted to rule-valid, non-memorized generations (Fig.~\ref{suppfig:haming_over_training}).
This is consistent with the intuition that the training samples are increasingly "attractive" whose basins devour more samples originally converging elsewhere. 
A more detailed analysis of per-sample training dynamics process can be found in App.~\ref{app:per_sample_dynamics_analysis}. 

\vspace{-8pt}
\subsection{A loss perspective of rule learning and memorization}
DiT training optimizes for the denoising score matching (DSM) objective $\mathcal{L}_\sigma = \mathbb{E}_{\x_0\sim p_0} \|\mathbf{D}_\theta(\x_0+\sigma \mathbf{z},\sigma)-\x_0\|^2$ integrated over some noise scale $\int_\sigma \mathcal{L}_\sigma d\sigma$. 
Here, we break down the aggregation and track the DSM loss at individual noise scales across training steps (Fig.~\ref{fig:diffusion_rule_mem_dissection}\textbf{B},\textbf{C}, App.~\ref{app:dsm_per_noise_scale}).  

We evaluate the DSM loss separately for I) training samples, II) held-out rule-valid samples, and III) random samples from the boolean hypercube,
which allows to interpret what the learned score captures.
Before the rule-learning transition, the DSM losses are indistinguishable between training samples and random Boolean samples (\Cref{fig:diffusion_rule_mem_dissection}\textbf{B}, $\tau<14$k), suggesting that the model is basically learning the score of the Boolean hypercube. 
After $\trule$, the DSM loss on training samples becomes lower than the hypercube (\Cref{fig:diffusion_rule_mem_dissection}\textbf{B}, $\tau>14$k).
Further, around $\tmem$, the DSM loss on training samples and held-out samples diverge from each other (\Cref{fig:diffusion_rule_mem_dissection}\textbf{B}, $\tau>492$k; \cite{bonnaire2025memorize}).

Notably, after $\tmem$, the DSM loss separates between training and held-out samples only over an intermediate range of noise scales, $\sigma \in [0.2,2.0]$ (\Cref{fig:diffusion_rule_mem_dissection}\textbf{C}), recapitulate findings in \cite{dodson2026two}.
This range matches the geometry of the Boolean hypercube: nearest Boolean vertices are separated by Euclidean distance $2$, while nearest even-parity vertices are separated by $2\sqrt{2}$.
Thus, memorization emerges when the denoising scale is comparable to the distance between neighboring data points, making it a property of the learned vector field at specific intermediate distances from the training samples.
We next analyze this vector-field mechanism in detail in \Cref{sec:vec_field}.


\subsection{A vector field and landscape view of rule learning and memorization}
\label{sec:vec_field}
We directly visualize the learned vector field at different training steps $\tau$ and noise scales $\sigma$ (Fig.~\ref{fig:diffusion_rule_mem_dissection}\textbf{D}),
focusing on the critical noise scale around the phase transition times. 
We visualize a 2d slice of the learned high-dimensional vector field.
The slice is chosen to pass through four Boolean vertices on a face of the Boolean hypercube: a training sample $\x_a$, a rule-valid held-out sample $\x_b$ at Hamming distance $2$ from $\x_a$, and two rule-violating odd-parity samples $\x_c,\x_d$, each at Hamming distance $1$ from both $\x_a$ and $\x_b$. %
The visualization (App.~\ref{app:vector_field_basin}) reveals the following two clear phenomena.  

\paragraph{Rule learning transition}
Before rule learning, all four points are attractors with similar sized, symmetric basins (Fig.~\ref{fig:diffusion_rule_mem_dissection}\textbf{D}, step 14k).
After rule learning, suddenly, the attractor basins for rule-valid samples $x_a,x_b$ expand sharply, shrinking the coverage of the basins of rule-violating samples $x_c,x_d$ (Fig.~\ref{fig:diffusion_rule_mem_dissection}\textbf{D}, step 119k).
As a results, much fewer samples can fall on to the rule-violating corners of the cube.
Note that at this moment, the basins of $x_a,x_b$ are still roughly symmetric, consistent with the identical DSM losses on training and held-out samples (Fig.~\ref{fig:diffusion_rule_mem_dissection}\textbf{C}, left).
This configuration is stable for a long time, until the memorization onset.

\paragraph{Memorization onset}
The onset happens more gradually than the rule learning transition.
At the onset, the basin of the training sample $x_a$ continues to expand and invades the basin of $x_b$, as evidenced by the shift of the decision boundary towards $x_b$ (Fig.~\ref{fig:diffusion_rule_mem_dissection}\textbf{D}, step 492k).
This is consistent with the post-$\tmem$ divergence of DSM loss between training and held-out samples (Fig.~\ref{fig:diffusion_rule_mem_dissection}\textbf{B,C}): the learned vector field becomes a poorer denoiser for held-out samples, pointing toward the memorized training point $\x_a$ even when the noised samples are closer to $\x_b$. 
This attractor basin of $x_a$ keeps growing and eventually devours the odd-parity samples $x_c,x_d$, as quantified by the basin-profile measured along 1d interpolation between $x_a\to x_c$ and $x_a\to x_b$ (Fig.~\ref{fig:diffusion_rule_mem_dissection}\textbf{E}). 
Notably, this also explains the increasing invalid sample ratio in the late phase (Fig.~\ref{fig:diffusion_rule_mem_dissection}\textbf{A}, Fig.~\ref{fig:two_clocks_and_scaling}\textbf{B}, late phase): while the trajectory points towards the memorized samples, a noised sample may accidentally pass by the rule-violating samples $x_c,x_d$ along the trajectory, and get trapped by them as the noise scale lowers --- at lower noise scales, the score is still very much that of the boolean hypercube, so every corner of the cube is still an attractor (Fig.~\ref{fig:diffusion_rule_mem_dissection}\textbf{C}).
This is also consistent with our previous observation that the late-phase invalid samples do not form a stable state, but keep flipping back to valid samples (Fig.~\ref{fig:diffusion_rule_mem_dissection}\textbf{A}, bottom).

\section{Generalization to broader settings}
\label{sec:generalize}

Our investigation has so far focused on diffusion model learning parity.
This section shows that the observations therein hold more broadly across architectures and tasks.

\subsection{Architecture generalization to autoregressive models}
\label{sec:gpt}

\begin{figure}[t]
    \centering
    \vspace{-20pt}
    \includegraphics[width=0.99\linewidth]{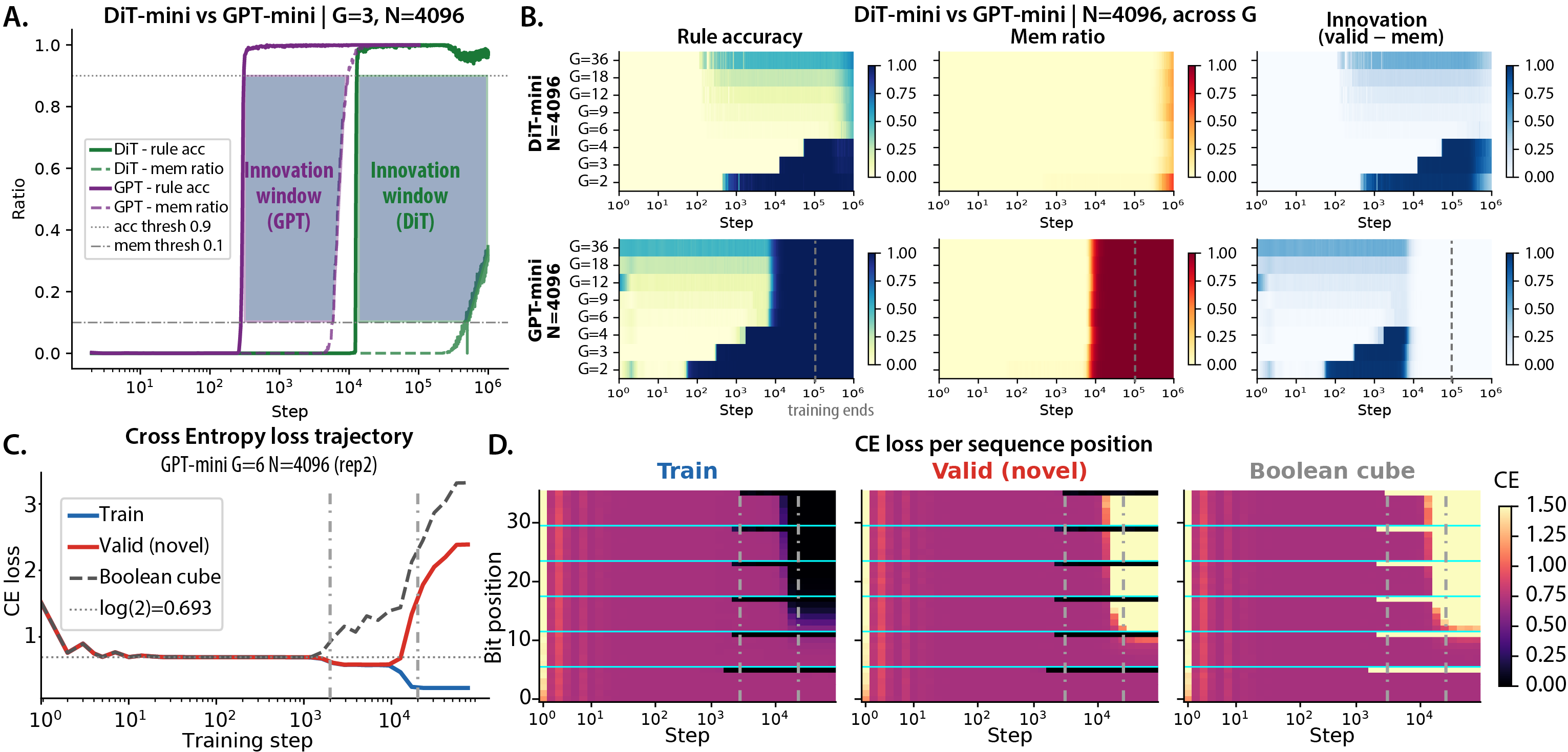}
    \vspace{-8pt}
    \caption{\textbf{Diffusion (DiT) and autoregressive (GPT) transformers share the two-clock structure with different timescales.}
    \textbf{A.}~Sample-level rule accuracy (solid) and memorization ratio (dashed) vs.~training step for DiT-mini (green) and GPT-mini (purple) at $G{=}3$, $N{=}4096$. Shaded: innovation window where rule accuracy $>0.9$ and memorization $<0.1$.
    \textbf{B.}~Per-architecture heatmaps across $G \in \{2,3,4,6,9,12,18,36\}$ at $N{=}4096$: rule accuracy (left), memorization ratio (middle), innovation $\text{rule\_acc}-\text{mem\_ratio}$ (right). Top row: DiT-mini; bottom row: GPT-mini; vertical grey dashed line marks the GPT training cutoff.
    \textbf{C.}~GPT-mini cross-entropy loss trajectory with $G{=}6$, $N{=}4096$ (run~2), on training samples (blue), held-out rule-valid samples (red), and random Boolean-cube samples (dashed grey). Reference: $\log(2)\!\approx\!0.693$.
    \textbf{D.}~GPT-mini per-bit-position cross-entropy heatmaps for the same three datasets (Train, Valid novel, Boolean cube). Cyan horizontal lines: $G{=}6$ group boundaries; vertical dashed lines: $\trule$ and $\tmem$.}
    \label{fig:dit_vs_gpt}
\end{figure}

One may wonder whether observations in previous sections are specific to diffusion models or are a general property of generative models.
We investigate this question by training GPT2-style transformers, whose architecture
matches with DiT
\footnote{When having the same architecture, causal attention strictly weakens expressivity~\citep{chen25multilayer}.
Our later discussion shows that GPT learns faster despite the capacity disadvantage.}
but using autoregressive modeling objective on the same group-parity datasets.
We provide hyperparameter ablation in \Cref{app:optim_ablation}.

We find a similar two clocks structure for autoregressive models: for simpler rules $G\leq6$, the GPT experiences a similar rule learning transition and a later memorization ``transition'' (Fig.~\ref{fig:dit_vs_gpt}\textbf{A}). 
Similar to diffusion (Fig.~\ref{fig:two_clocks_and_scaling}\textbf{A}), $\trule$ for GPT is delayed by rule complexity $G$, while $\tmem$ is relatively consistent across $G$ (Fig.~\ref{fig:dit_vs_gpt}\textbf{B}) and linearly modulated by the sample size $N$ ($\tmem\approx2.1N^{0.97}, R^2=0.94,n=34$, GPT-mini, all $G$, Tab.~\ref{tab:mem_scaling}, Fig.~\ref{suppfig:tau_mem_vs_N_scaling},~\ref{suppfig:gpt_tau_rule_mem_scaling}). 
Unlike diffusion though, GPT learns and memorizes faster than DiT of a matching size,
leading to a shorter ``innovation window'' in between (Fig.~\ref{fig:dit_vs_gpt}\textbf{A,B}).
In addition, the transition to memorization is more sudden than DiTs (Fig.~\ref{fig:dit_vs_gpt}\textbf{A}): at $N=4096$ GPT memorization rise to 1.0 in a few thousands of steps; in contrast, DiT memorization didn't saturate across $8\times 10^5$ steps. 
We hypothesize that this is because autoregressive training repeatedly sees the same prefixes in training, whereas the noises in DiT training effectively serve as data augmentation that slows down both learning and memorization.
This is also related to prior work's observation that diffusion can be more data-efficient~\citep{prabhudesai2025diffusion,ni2025diffusion}.

Mechanistically, autoregressive models exhibit similar loss dynamics, as evidenced by the cross entropy loss for next-token prediction (NTP) against I) training samples, II) held-out rule-valid samples, and III) random boolean samples.
The CE loss against random samples diverges from the other two at the rule learning transition; while the training and held-out samples losses diverge at the memorization onset time (Fig.~\ref{fig:dit_vs_gpt}\textbf{C}), mirroring the DSM-loss split observed in diffusion (cf.~Fig.~\ref{fig:diffusion_rule_mem_dissection}\textbf{B}).
Interestingly, the two NTP loss divergences have different positional composition. 
For rule learning, the hypercube vs training losses diverge at 
the last bit in each group (Fig.~\ref{fig:dit_vs_gpt}\textbf{D}), i.e., $i\mod G=0$, which are deterministic conditioned on the autoregressive prefix.
For memorization, the train-test divergence happens at all position past a certain position around $11$ (Fig.~\ref{fig:dit_vs_gpt}\textbf{D}). 
This creates an intriguing parallel: learning and memorization manifest at specific noise scales for diffusion (Fig.~\ref{fig:diffusion_rule_mem_dissection}\textbf{C}), and at specific token positions and prefix lengths in autoregressive models.

\subsection{Task generalization beyond parity}
\label{sec:beyond_parity}


We start with a broader choice of binary rules which also require global coupling.
Inspired by the counting problem in T2I diffusion~\citep{mou2023t2i0adapter0}, we consider the following rules:
1) \textit{exact-$K$}, where each sample has exactly $K$ bits positive; 
2) \textit{row-$K$}, where each row has exactly $K$ bits positive; 
and more in App.~\ref{app:beyond_parity_rules}. Example training samples are shown in Fig.~\ref{suppfig:rule_sample_montage}.  
Notably, exact-$K$ and row-$K$ rule has a rapid $\trule$ comparable to parity with $G\in[2,3]$; further their $\trule$ is insensitive to the choice of $K$ (Fig.~\ref{suppfig:beyond_parity_tau_mem_rule_scaling}\textbf{B}). Further, the memorization onset $\tmem$ is insensitive to the types of rules, showing robustness of our result. 


Beyond learning binary rule, we experiment with Latin square and Sudoku (6x6), which are more natural and complex with multi-valued inputs (Fig.~\ref{suppfig:rule_sample_montage_latinsquare}). 
We find DiTs can learn to generate substantial fraction of novel and valid samples (Fig.~\ref{suppfig:beyond_parity_latinsq_sudoku}), but the max accuracy is farther from 1.0 than previous rules. We think the difficulty comes from the nestedness of constraints: even though uniqueness is equivalent to row-$K=1$, 
Lifting the column uniqueness constraint, the row-only Latin square can be learned rapidly, and accuracy rise to perfection within 1k steps (Fig.~\ref{suppfig:beyond_parity_latinsq_sudoku_tau_rule_mem}), thus the $\trule$ is indeed comparable to row-$K$. 
A full investigation of these rules is deferred to future work. 

\paragraph{An energy perspective for learning speed across tasks}
Above, we show that different rules exhibit markedly different learning times. 
A revealing contrast is parity versus \textit{exact-$K$}: exact-$K$ imposes a global constraint over all $36$ bits, yet is learned as quickly as the easiest parity cases ($G=2$), and its $\trule$ is largely insensitive to $K$ (Fig.~\ref{suppfig:beyond_parity_tau_mem_rule_scaling}\textbf{A,B}).
Thus, globality alone does not determine rule-learning difficulty; the form of the constraint matters.

To see this, write a binary-rule energy as
{\small
\begin{align}
\label{eq:energy_d_bit_binary}
E_d(\x)
=
\frac{1}{2}\sum_{i=1}^d(x_i^2-1)^2
+
\lambda_p E_f(\x),
\qquad \lambda_p>0,
\end{align}
}
where the first term encourages $\x$ to lie on the Boolean cube and $E_f$ encodes the rule.
For the corresponding density $p_d(\x;\beta)\propto\exp[-\beta E_d(\x)]$, the low-temperature limit $\beta\to\infty$ concentrates on the rule-valid set $\mathcal P_d^+$, and the optimal score satisfies
\begin{align}
\label{eq:parity_d_bit_score}
\nabla\log p_d(\x;\beta)
=
-\beta\nabla_\x E_d(\x),
\qquad
\nabla_{x_i}E_d(\x)
=
2x_i(x_i^2-1)
+
\lambda_p\nabla_{x_i}E_f(\x).
\end{align}
The first, bit-local term explains the rapid learning of the Boolean hypercube (Fig.~\ref{fig:two_clocks_and_scaling}\textbf{B}); the second, rule-specific term accounts for the task-dependent learning difficulty.

For parity, one energy model reads 
$E_{G\text{-parity}}(\x)=\frac{1}{2}\sum_{i\in[d/G]}\big(\prod_{j\in[G]}x_{ij}-1\big)^2$,
whereas for exact-$K$,
$E_{\text{exact-}K}(\x)=\frac{1}{2}\big(\sum_{i=1}^d x_i-K\big)^2$.
Parity therefore requires degree-$G$ multiplicative interactions in the rule-specific score, while exact-$K$ is a low-degree count constraint.
This is consistent with the known difficulty of learning high-degree parity-like functions~\citep{abbe2023sgd,abbe2024generalization}.

\section{Discussion}

Generative models are often evaluated by the realism and diversity of their samples, but genuine generalization also requires learning the latent rules that define a distribution. 
In this work, using controlled rule, we identify two distinct clocks of training: $\trule$, the onset of rule-valid generation, and $\tmem$, the onset of memorization.
Across diffusion and autoregressive models, we find these two clocks are governed by different factors: $\trule$ is primarily controlled by rule complexity, whereas $\tmem$ scales with dataset size, both modulated by model capacity. Their separation defines an \emph{innovation window}, during which models generate valid but non-memorized samples. 
By dissecting diffusion training dynamics, losses, and learned vector fields, we further show that these clocks correspond to distinct changes in the learned attractor landscape. 
Together, these results provide a dynamical account of when and how generative models generalize, memorize, and genuinely innovate.



\paragraph{Implications for natural data.}
Our findings suggest that relations involving many-way interactions are inherently difficult for current diffusion architectures.
Such interactions may underlie the difficulty of learning general abstract reasoning rules. 
In addition to tasks in \Cref{sec:beyond_parity}, another example studied in prior work is the RAVEN progression matrices, where~\cite{wang_diverse_2024} showed that XOR-type relations over multiple attributes are especially hard for diffusion models.
More broadly, our results imply that scientific or physical constraints depending on large-scale multiplicative structure, such as conservation laws involving many coupled quantities, may not be faithfully learned without targeted architectural or training interventions.

\paragraph{Pathways to improved rule learning.}
One is to modify the architecture to enable \emph{global broadcasting} of information—through dedicated register tokens~\cite{darcet2023ViTregister}, global memory units, or structured multiplicative interactions—so that the model can aggregate and disseminate the features required for large-$G$ parity in a single step. Another is to enrich training with auxiliary objectives that explicitly require detecting and representing parity-like dependencies, such as masked group-parity prediction, to encourage the formation of suitable internal representations \cite{yu2024repa}. Finally, a curriculum that gradually increases $G$ during training could scaffold the acquisition of higher-order rules, allowing the network to build on simpler cases before tackling more complex ones.

\paragraph{Broader outlook.}
Although tasks focused in this work are synthetic, they isolate a fundamental limitation: global rules with high interaction order are not well aligned with the inductive biases of current diffusion transformers. Addressing this limitation is critical for applications where rule adherence is as important as perceptual fidelity, including symbolic reasoning, structured design, and scientific modeling. Our group-parity testbed provides a controlled setting in which to explore both the failure modes and potential solutions, and offers a stepping stone toward architectures that can internalize and apply abstract, combinatorial rules from data.

\begin{ack}
B.W. and B.L. were supported by the Kempner Fellowship from the Kempner Institute at Harvard. This work was made possible in part by a gift from the Chan Zuckerberg Initiative Foundation to establish the Kempner Institute for the Study of Natural and Artificial Intelligence, and by the generous computing resources provided by the Kempner Institute and Harvard Research Computing. E.F. would also like to thank the CRISP Lab and the Calvin Coolidge Presidential Foundation for their support during her undergraduate career.

We thank the organizers of the NeurIPS 2025 Workshop on Structured Probabilistic Inference and Generative Modeling for accepting an earlier version of this work as an oral presentation, and for the helpful discussions it generated. We are grateful to T. Anderson Keller for discussions on the connections between various energy functions and their corresponding score architectures, which inspired our focus on the parity task. We thank Alex Damian for insightful discussions on the parametric energy model of parity and for pointing us to relevant literature on learning dynamics. Finally, we thank Shane Jiaqi Shang and Haim Sompolinsky for early discussions on rule learning in diffusion models and transformers — particularly regarding the RAVEN task — from which our work draws much of its inspiration. 

\end{ack}

\bibliographystyle{plainnat}
\bibliography{iclr2025_conference}

\newpage
\appendix
\tableofcontents

\input{related_work}

\section{Experimental Details}
\label{app:experimental_details}

\subsection{DiT architecture and training}\label{app:train_model_method}

\paragraph{Architecture.}
We use a 2d Diffusion Transformer (DiT)~\citep{peebles2023scalable} for images.
The input $\mathbf{x}\in\{-1,+1\}^D$ ($D{=}36$) is reshaped to a $6{\times}6$ single-channel image and tokenized with patch size $1$, yielding $36$ tokens each of dimension $1$.
Tokens are linearly embedded to the hidden dimension and combined with 2D sinusoidal positional embeddings.
The noise level $\sigma$ is embedded via a Fourier feature projection followed by a two-layer MLP; these timestep embeddings modulate each transformer block via adaptive layer norm (adaLN).
There is no class conditioning (\texttt{num\_classes=0}) and the model does not predict the noise variance (\texttt{learn\_sigma=False}).

Model scales used in this work are summarized in Table~\ref{tab:dit_models}.
The primary model for all analyses is DiT-mini; DiT-nano, DiT-S, and DiT-B are used for model-scale ablations.

\begin{table}[h]
\centering\small
\begin{tabular}{lcccc}
\toprule
Model & $d_\text{model}$ & Depth & Heads & \#Params \\
\midrule
DiT-nano & 192  & 3  & 3  & ${\sim}2.2$M \\
DiT-mini & 384  & 6  & 6  & ${\sim}16.5$M \\
DiT-S    & 384  & 12 & 6  & ${\sim}32.5$M \\
DiT-B    & 768  & 12 & 12 & ${\sim}129.5$M \\
\bottomrule
\end{tabular}
\caption{DiT model scales. All use \texttt{mlp\_ratio=4}, \texttt{patch\_size=1}.}
\label{tab:dit_models}
\end{table}

\paragraph{EDM preconditioning.}
We adopt the EDM preconditioning of \citet{karras2022elucidatingDesignSp}.
Given a noisy input $\tilde{\mathbf{x}} = \mathbf{x} + \mathbf{n}$, $\mathbf{n}\sim\mathcal{N}(0,\sigma^2 I)$, the network $F_\theta$ produces a denoised estimate via:
\begin{equation}
    D_\theta(\tilde{\mathbf{x}}, \sigma) = c_\text{skip}(\sigma)\,\tilde{\mathbf{x}} + c_\text{out}(\sigma)\,F_\theta\!\left(c_\text{in}(\sigma)\,\tilde{\mathbf{x}},\; c_\text{noise}(\sigma)\right),
\end{equation}
where $c_\text{skip} = \sigma_\text{data}^2/(\sigma^2+\sigma_\text{data}^2)$, $c_\text{out} = \sigma\sigma_\text{data}/\sqrt{\sigma^2+\sigma_\text{data}^2}$, $c_\text{in} = 1/\sqrt{\sigma_\text{data}^2+\sigma^2}$, and $c_\text{noise} = \ln(\sigma)/4$.
We set $\sigma_\text{data}{=}1.0$ (consistent with data in $\{-1,+1\}^D$).

\paragraph{Loss and noise schedule.}
We use the EDM denoising score matching (DSM) objective:
\begin{equation}
    \mathcal{L} = \mathbb{E}_{\mathbf{x},\sigma,\mathbf{n}}\bigl[\lambda(\sigma)\,\|D_\theta(\mathbf{x}+\mathbf{n},\sigma) - \mathbf{x}\|^2\bigr],
\end{equation}
with loss weight $\lambda(\sigma) = (\sigma^2+\sigma_\text{data}^2)/(\sigma\,\sigma_\text{data})^2$.
Noise levels are drawn from a log-normal distribution: $\ln\sigma \sim \mathcal{N}(P_\text{mean},\, P_\text{std}^2)$ with $P_\text{mean}{=}{-}1.2$, $P_\text{std}{=}1.2$ (same as the EDM paper defaults for continuous data).
This distribution concentrates training at intermediate noise levels ($\sigma\approx e^{-1.2}{\approx}0.3$) while providing coverage from $\sigma_\text{min}{=}0.002$ to $\sigma_\text{max}{=}80$.

\paragraph{Optimizer and training schedule.}
All DiT experiments use Adam ($\beta_1{=}0.9$, $\beta_2{=}0.999$) with learning rate $\eta{=}10^{-4}$, batch size $256$, and $10^6$ gradient steps.
Weight decay is $0$ for baseline runs; AdamW with decoupled weight decay is used only in the optimization ablation (\S\ref{app:optim_ablation}).
No learning rate schedule or warmup is applied.
Training data is sampled once before training and held fixed throughout; the model sees each sample approximately $\lfloor 10^6 \times 256 / N \rfloor$ times.

\subsection{Compute resources}
\label{app:compute_resources}

Experiments were run on NVIDIA H100 and A100 GPUs with approximately 150GB available GPU memory.
The wall-clock times in Table~\ref{tab:compute_resources} give representative runtimes for the main model families under the training schedules used in this paper, measured from TensorBoard event timestamps or checkpoint file modification times.

\begin{table}[h]
\centering\small
\begin{tabular}{lccccc}
\toprule
Model family & Training steps & Wall-clock / run & \# Runs & Total GPU-hours \\
\midrule
DiT-nano & $10^6$ & ${\sim}9$ h  &   8 &  ${\sim}72$ h \\
DiT-mini & $10^6$ & ${\sim}8$ h  & 102 & ${\sim}816$ h (${\sim}34$ days) \\
DiT-B    & $10^6$ & ${\sim}48$ h &  12 & ${\sim}576$ h (${\sim}24$ days) \\
GPT-nano & $10^5$ & ${\sim}1$ h  &   8 &   ${\sim}8$ h \\
GPT-mini & $10^5$ & ${\sim}50$ min &  45 &  ${\sim}37$ h \\
GPT-B    & $10^5$ & ${\sim}5$ h  &  16 &  ${\sim}80$ h \\
\midrule
\textbf{Total} & & & \textbf{191} & ${\sim}\mathbf{1590}$ \textbf{GPU-h} (${\sim}\mathbf{66}$ \textbf{GPU-days}) \\
\bottomrule
\end{tabular}
\caption{Compute requirements. All runs used NVIDIA H100 or A100 GPUs. Wall-clock times are measured from TensorBoard event timestamps or checkpoint file modification times. Run counts cover all tasks (parity, exact-$K$, row-$K$, Latin square, sudoku, etc.) and represent a lower bound (excludes aborted or exploratory runs).}
\label{tab:compute_resources}
\end{table}

\subsection{Sampling and evaluation metrics}\label{app:eval_method}
Throughout training, at increasingly larger spacing of steps, we generate samples from a fixed set of 2048 initial noises, and evaluate them according to the following criterion. We used Heun's 2nd order deterministic sampler with 35 steps from $\sigma_{max}=80.0$ to $\sigma_{min}=0.002$ \citep{karras2022elucidatingDesignSp}. 

First, we evaluate how far the samples are from the boolean cube $\{-1,1\}^D$, as measured by the $\ell_{\infty}$ distance $d_{\ell_\infty} (\x)=\max_i||x_i|-1|$.
We call a sample \emph{invalid} when $d_{\ell_\infty} (\x)>\epsilon$, and calculated the invalid fraction for various thresholds $\epsilon \in \{0.1,0.01\}$.

Next, we binarize each element of the sample to $\{-1,1\}$ and evaluate the parity of the binarized sample $\bar \x$, at both group and sample level (Fig.~\ref{fig:dataset_eval_model_cmp}\textbf{A}).
The group parity accuracy is defined over a group of $G$ elements, with chance level $2^{-1}$.
For the sample parity accuracy, a prediction is correct only if the parities of all $D/G$ groups are correct,
for which the chance level is $2^{-D/G}$ (Fig.~\ref{fig:dataset_eval_model_cmp}B, dashed black). 

Further, we examine the \textit{memorization ratios} of groups and samples, which are the fractions of generated groups or samples that coincide with those in the training dataset.
If the model learns the true distribution (i.e., the uniform measure) on $(\mathcal{S}^+_{G})^{D/G}$, then the sample memorization ratio will be $N / (2^{G-1})^{D/G} = N\cdot 2^{-\frac{G-1}{G}D}$ (Fig.~\ref{fig:dataset_eval_model_cmp}B, dashed red); and group memorization ratio will be roughly $\min(2^{G-1},N\cdot D/G)/2^{G-1}$, while the specific value depends on the uniqueness of training groups, so it's computed empirically (Fig.~\ref{fig:dataset_eval_model_cmp}C, dashed red). 
If the model learns the uniform measure on the entire boolean cube, then the sample memorization will be $N\cdot 2^{-D}$ (Fig.~\ref{fig:dataset_eval_model_cmp}B, dashed blue); and group memorization ratio will be roughly $\min(2^{G-1},N\cdot D/G)/2^{G}$, with empirical values showed (Fig.~\ref{fig:dataset_eval_model_cmp}C, dashed blue).

\subsection{Task definitions and dataset generation}
\label{app:task_definitions}

\subsubsection{Exact-$K$ and row-$K$}
\label{app:exactk_rowk}

\begin{figure}[!htp]
    \centering
    \includegraphics[width=0.8\linewidth]{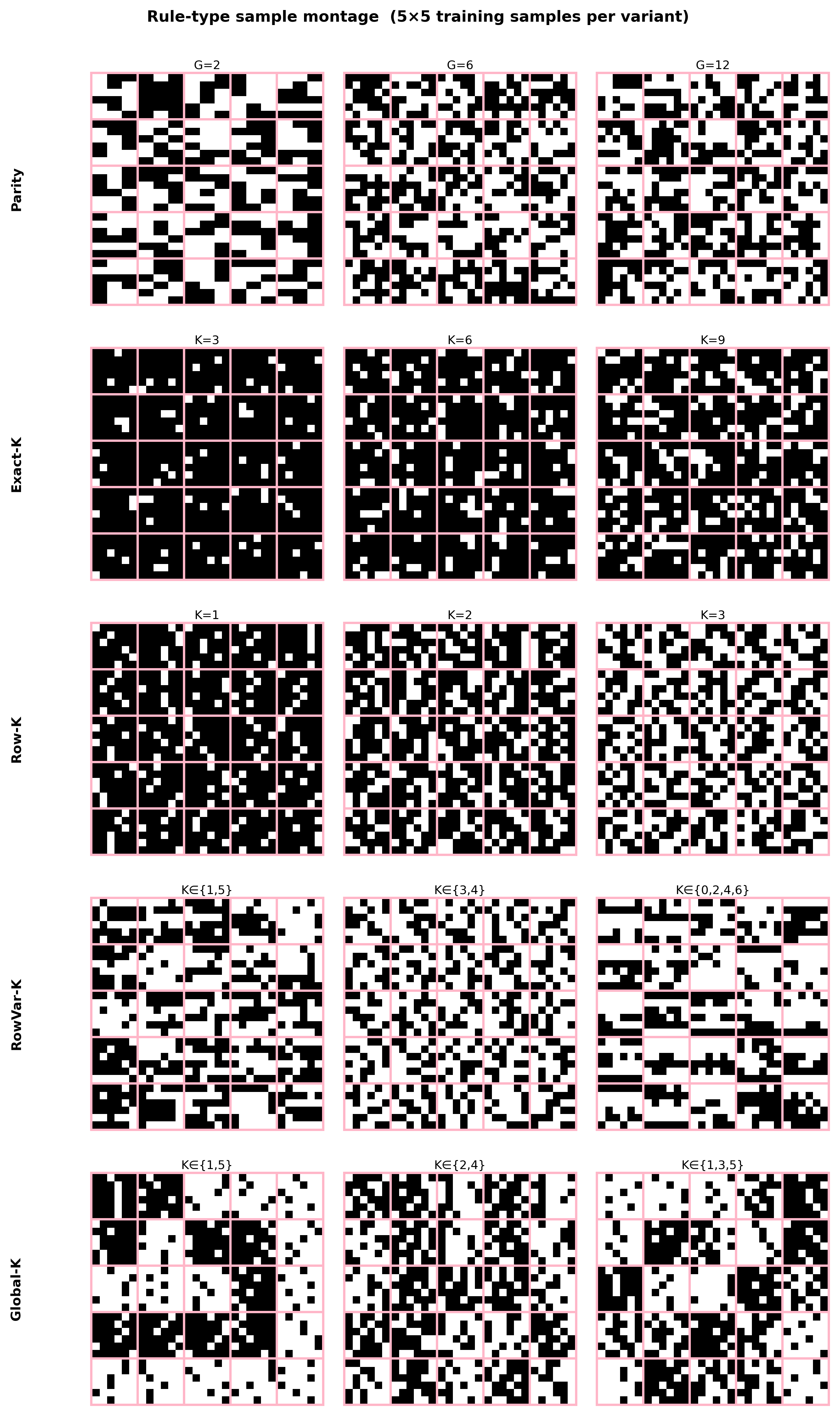}
    \caption{\textbf{Training-sample montage for binary rule families.}
    Each panel shows a $5{\times}5$ grid of representative $6{\times}6$ training samples for one rule variant.
    Rows correspond to group parity, exact-$K$, row-$K$, row-variable-$K$, and global-$K$ datasets.
    Columns vary the group size $G$, exact count $K$, or allowed count set $\mathcal{K}$.
    The montage illustrates how the same binary image space supports qualitatively different constraints: parity imposes high-order group interactions, exact-$K$ imposes a global count, row-$K$ imposes local row counts, row-variable-$K$ allows each row to choose from a set, and global-$K$ shares a single count choice across all rows in a sample.}
    \label{suppfig:rule_sample_montage}
\end{figure}

The binary rule datasets all use the same $6{\times}6$ image format, but differ in which subsets of bits are coupled by the constraint.
This shared visual format lets us compare rule-learning and memorization dynamics without changing the model architecture or sampler.

\paragraph{Exact-$K$ rule.}
A binary vector $\mathbf{x} \in \{-1,+1\}^D$ satisfies the \textit{exact-$K$} rule if exactly $K$ entries equal $+1$:
\begin{equation}
    \sum_{i=1}^{D} \mathbf{1}[x_i = +1] = K.
\end{equation}
The valid set has size $\binom{D}{K}$.
For $D{=}36$ this ranges from $\binom{36}{3}{=}7{,}140$ ($K{=}3$) to $\binom{36}{18}{\approx}9.1{\times}10^9$ ($K{=}18$).
Crucially, the constraint can be written as a single quadratic penalty $(\sum_i x_i - (2K{-}D))^2$, so the score (gradient) couples all bits through a \emph{degree-1} global sum.
This contrasts with group parity, whose score involves a degree-$(G{-}1)$ product, making exact-$K$ far easier to learn.

\paragraph{Exact-$K$ evaluation.}
After binarizing generated samples to $\{-1,+1\}$ (threshold $\epsilon{=}0.1$), a sample is \emph{rule-valid} if its count of $+1$ entries equals exactly $K$.
Sample accuracy and memorization ratio are computed identically to the parity case (\S\ref{app:eval_method}), with statistical memorization baseline $N/\binom{D}{K}$.

\paragraph{Exact-$K$ experiments.}
We train DiT-mini ($D{=}36$, $N{=}4096$) for $K\in\{3,4,6,8,9,12,18\}$.
All use the same protocol as the parity experiments (lr$=10^{-4}$, batch$=256$, $10^6$ steps, EDM sampler with 35 steps).
Across all $K$ values, a clear two-phase rule-then-memorize transition is observed, with rule learning occurring orders of magnitude faster than for comparable parity tasks, consistent with the lower-degree score argument.

\paragraph{Row-$K$ rule.}
We additionally study constraints on an $n{\times}n$ grid ($n{=}6$, $D{=}36$).
Three constraint variants are considered:
\begin{itemize}[leftmargin=*,itemsep=1pt]
    \item \textbf{Row-$K$ (fixed):} every row has exactly $K$ entries equal to $+1$.  Valid set: $\binom{n}{K}^n$.
    \item \textbf{Row-variable-$K$:} each row independently draws its count from a fixed set $\mathcal{K}$.  Valid set: $\bigl(\sum_{k\in\mathcal{K}}\binom{n}{k}\bigr)^n$.
    \item \textbf{Global-$K$:} a single $K\in\mathcal{K}$ is drawn per sample, shared across all rows.  Valid set: $\sum_{k\in\mathcal{K}}\binom{n}{k}^n$.
\end{itemize}
These variants isolate whether the model learns per-row structure, or must additionally infer which $K$ applies globally.

\paragraph{Row-$K$ evaluation.}
A generated grid is rule-valid if every row has the correct count of $+1$ entries ($=K$ for fixed, $\in\mathcal{K}$ for variable/global).
Sample accuracy is the fraction of grids where \emph{all} rows satisfy the constraint simultaneously.

\paragraph{Row-$K$ experiments.}
We train DiT-mini ($n{=}6$, $N{=}4096$) on: fixed $K\in\{2,3\}$; variable-$K$ with $\mathcal{K}\in\bigl\{\{1,5\},\{3,4\},\{0,2,4,6\},\{3,4,5,6\}\bigr\}$.
Scalar-encoded ($[-1,+1]$) and one-hot (with zero-mean normalization) representations are both evaluated.
As with exact-$K$, all row-$K$ variants learn the rule in far fewer steps than parity, confirming that degree-1 per-row sums are easily captured by the DiT score.

\subsubsection{Latin square and Sudoku}
\label{app:latinsq_sudoku}

\begin{figure}[!htp]
    \centering
    \includegraphics[width=0.8\linewidth,height=0.82\textheight,keepaspectratio]{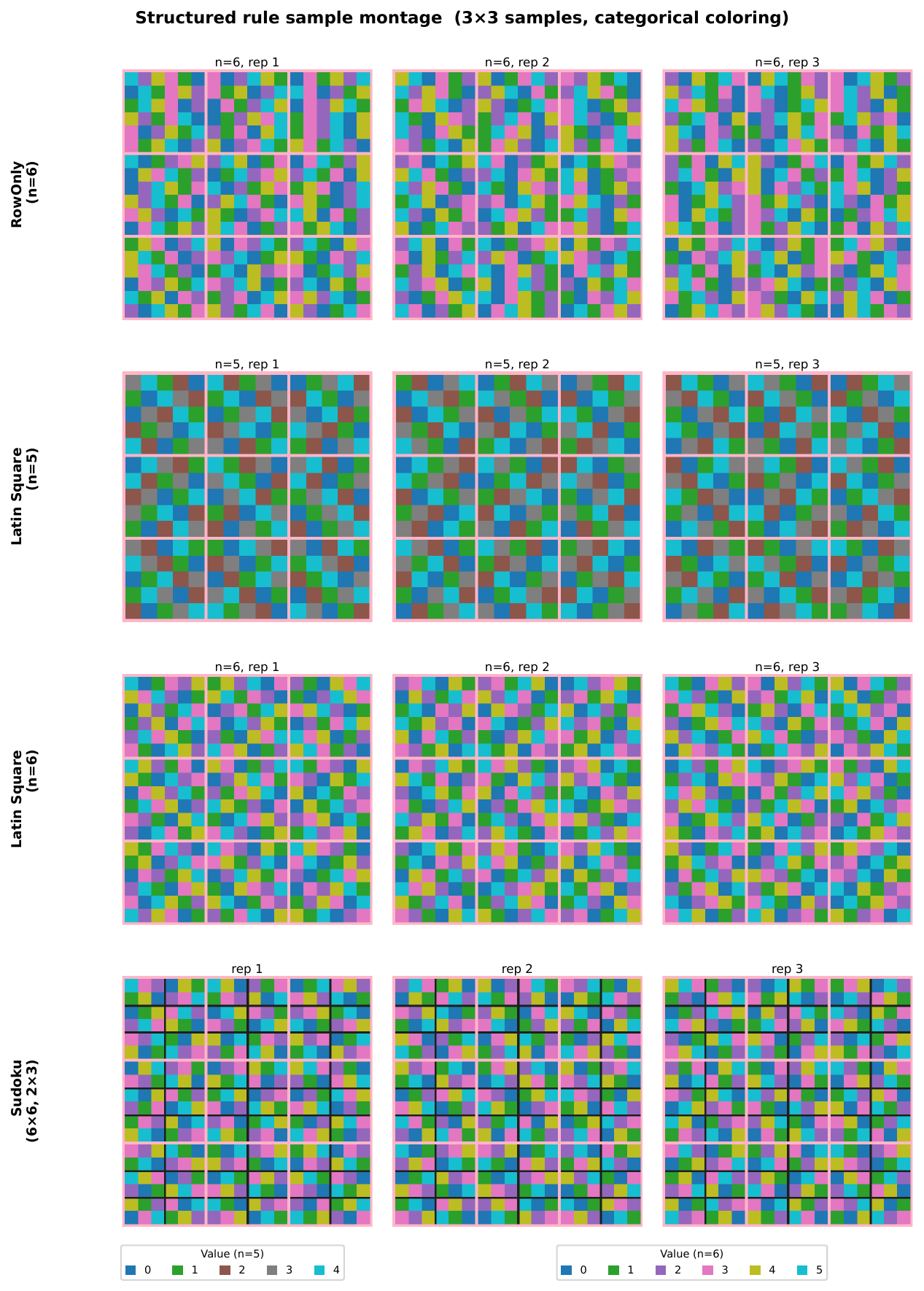}
    \caption{
    \textbf{Training-sample montage for structured categorical rule families.}
    Each row shows representative training samples from a different structured rule family, with three independently drawn samples shown per family.
    Rows correspond to RowOnly Latin Square ($n=6$), Latin Square ($n=5$), Latin Square ($n=6$), and Sudoku ($6{\times}6$ grid with $2{\times}3$ blocks).
    Colors denote categorical values: five colors for $n=5$ and six colors for $n=6$.
    The montage illustrates how the same grid-based categorical image space supports increasingly structured constraints, from row-wise value constraints to Latin-square row/column constraints and Sudoku-style row, column, and block constraints.
    }
    \label{suppfig:rule_sample_montage_latinsquare}
\end{figure}

\paragraph{Latin square rule.}
An $n{\times}n$ \textit{Latin square} is a grid filled with $n$ distinct symbols $\{0,\ldots,n{-}1\}$ such that each symbol appears exactly once in every row and exactly once in every column.
The number of valid $n{\times}n$ Latin squares is exactly $12, 576, 161{,}280, 812{,}851{,}200$ for $n=3,4,5,6$.
The row and column uniqueness constraints require each symbol to appear exactly once in each row and column, placing Latin squares structurally between row-$K$ (row-only, single-symbol) and more complex combinatorial rules.

\paragraph{Latin square row only rule}
We also considered a easier rule than Latin square, i.e. only considering the row uniqueness rule, namely, each row is a permutation of $n$, thus the number of valid $n{\times}n$ row only Latin squares samples is $(n!)^n$, i.e. 
$216,\;331{,}776,\;24{,}883{,}200{,}000,\;139{,}314{,}069{,}504{,}000{,}000$ for $n=3,4,5,6$.

\paragraph{Sudoku rule.}
A $6{\times}6$ \textit{Sudoku} (blocks $2{\times}3$) adds a third uniqueness constraint: each of the six $2{\times}3$ sub-blocks must also contain all six symbols exactly once.
This reduces the valid set from $812{,}851{,}200$ (Latin square) to $28{,}200{,}960$ (${\approx}3.5\%$ of Latin squares satisfy the block constraint).

\paragraph{Encoding.}
Two encodings are evaluated: (i) \textbf{scalar} — integer symbols linearly mapped to $[-1,+1]$ (for $n{=}6$: $\{0,\ldots,5\}\mapsto\{-1.0,-0.6,\ldots,+1.0\}$, input tensor shape $(n,n)$); (ii) \textbf{one-hot} — each cell encoded as a length-$n$ binary vector (input tensor shape $(n,n,n)$), optionally zero-mean normalized.

\paragraph{Evaluation.}
For scalar encoding, generated samples are snapped to the nearest valid symbol ($\epsilon{=}0.15$) and verified for row, column, and (for Sudoku) block uniqueness.
A sample is rule-valid only if all constraints are simultaneously satisfied.
For one-hot encoding, samples are decoded to integer grids before the same checks are applied.
Memorization ratio is computed as for parity: the fraction of generated grids that exactly match a training-set grid.

\paragraph{Experiments.}
We train DiT-mini and DiT-B on $n{=}5$ and $n{=}6$ Latin squares ($N{=}4096$, both encodings), and DiT-mini on $6{\times}6$ Sudoku ($N{=}4096$, scalar and zero-mean one-hot).
Latin squares are sampled via cyclic-base permutation (rows, columns, symbols), covering approximately $46\%$ of the $n{=}6$ isotopy classes; Sudoku grids are obtained by rejection-sampling Latin squares (approximately $29{\times}$ oversampling required).
Both tasks exhibit a clear rule-learning onset followed by a later memorization onset, demonstrating that the two-clock phenomenology extends to multi-valued, combinatorially structured rules.
\clearpage
\subsection{GPT architecture and training}
\label{app:gpt_training}

\paragraph{Architecture.}
We train three GPT model variants at different scales to assess model-size dependence, summarized in Table~\ref{tab:gpt_models}.
The default model used for all detailed analyses in this paper is GPT-mini.

\begin{table}[h]
\centering
\small
\begin{tabular}{lccccc}
\toprule
Model & $d_\text{model}$ & Layers & Heads & \#Params (approx.) \\
\midrule
GPT-nano & 384 & 3 & 6 & ${\sim}11$M \\
GPT-mini & 384 & 6 & 6 & ${\sim}22$M \\
GPT-B    & 768 & 12 & 12 & ${\sim}86$M \\
\bottomrule
\end{tabular}
\caption{GPT model variants used in parity learning experiments.}
\label{tab:gpt_models}
\end{table}

GPT-mini matches the scale of the DiT-mini used for diffusion experiments.
We additionally sweep dataset sizes $N \in \{4096, 8192, 16384, 32768\}$ for GPT-mini, and train GPT-nano and GPT-B at $N{=}4096$ and $N{=}32768$ to verify that the two-phase rule-then-memorize dynamics are robust across model capacity and data scale.

\paragraph{Tokenization.}
The binary parity sequences $\mathbf{x} \in \{{\pm}1\}^D$ are first mapped to $\{0,1\}^D$, then a start-of-sequence (SOS) token with index $2$ is prepended, yielding integer sequences of length $D{+}1{=}37$ from a vocabulary of size $3$ ($0, 1, \text{SOS}$).
The model is trained with next-token prediction cross-entropy loss, so at each position $k \in \{1,\ldots,D\}$ it predicts the $k$-th bit given the SOS token and bits $1$ through $k{-}1$.

\paragraph{Training setup.}
All GPT-mini baseline experiments use AdamW with learning rate $\eta{=}10^{-4}$, weight decay $\lambda{=}0.01$, batch size $256$, and $10^5$ gradient steps.
We save $40$ checkpoints spaced uniformly in log-step (from step ${\approx}500$ to $10^5$).
At each checkpoint, $N{=}2048$ sequences are sampled autoregressively at temperature $1.0$ and decoded back to $\{{\pm}1\}^D$; rule accuracy and memorization ratio are computed identically to the DiT evaluation (\S\ref{app:eval_method}).

\paragraph{Cross-entropy analysis across dataset splits.}
To dissect the rule-learning and memorization phases in the GPT, we evaluate the next-token prediction cross-entropy (CE) at each checkpoint on three held-out splits of $N{=}4096$ sequences each:
\begin{enumerate}
    \item \textbf{Training set}: the $N{=}4096$ samples seen during training.
    \item \textbf{Valid-novel test set}: rule-valid samples drawn from the same group-parity distribution, verified to not appear in the training set (Hamming distance $\geq 1$ to every training sample).
    \item \textbf{Boolean-cube samples}: uniformly random $\{0,1\}^D$ binary sequences with no parity structure, representing an unstructured null distribution.
\end{enumerate}
The valid-novel CE first \emph{decreases} when the model internalizes the parity rule, then \emph{increases} again as memorization progresses and the model overfits the training distribution---a signature of grokking-then-forgetting dynamics.
The boolean-cube CE serves as a ceiling reference; its divergence from the valid-novel curve marks the onset of rule learning.

\paragraph{Per-position cross-entropy.}
We additionally track the per-position CE: $\ell_k = -\mathbb{E}[\log p_\theta(x_k \mid x_1,\ldots,x_{k-1})]$ for $k \in \{1,\ldots,36\}$ on the valid-novel split.
Positions whose index is a multiple of the group size $G$ (i.e., $k \equiv 0 \pmod{G}$) are the \emph{last position in each group}; given the preceding $G{-}1$ bits of the group, this position is fully determined by the parity constraint.
Consequently, its CE is expected to collapse to $0$ first as the model learns the rule, providing a high-resolution probe of which structural positions are acquired earliest.

\clearpage
\section{Extended Results and Additional Analyses}
\label{app:extended_results}

\subsection{Extended evaluation results for DiT-Parity at training endpoint}
\begin{figure}[!htp]
    \centering
    \vspace{-8pt}
    \includegraphics[width=0.99\linewidth]{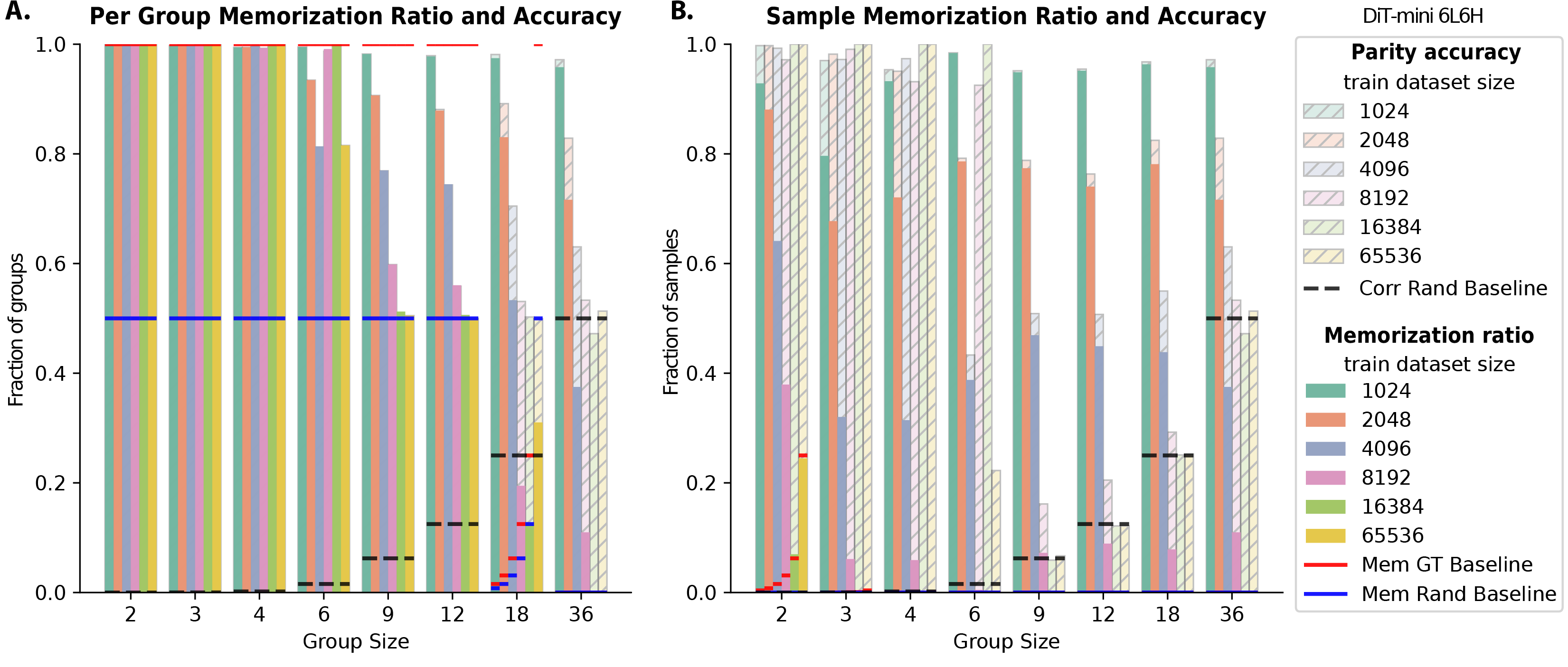}
    \vspace{-8pt}
    \caption{\textbf{Memorization and Creativity in Parity Learning across dataset scales.}
    Memorization ratio overlay on accuracy, for different group size and training dataset scale, at group level (\textbf{A.}) and sample level (\textbf{B.}), with similar format as Fig.~\ref{fig:dataset_eval_model_cmp}\textbf{B,C}.
Red solid line shows the memorization ratio of the ground truth distribution ($\mathcal P_G^+$ for groups, and $(\mathcal P_G^+)^{D/G}$ for samples); and blue solid line shows the memorization ratio of the chance distribution ($\mathcal U_G$ for groups and $\mathcal U_D$ for samples). Black dashed line shows the chance level accuracy.
}
    \vspace{-8pt}
    \label{suppfig:memorization_creativity_cmp}
\end{figure}

\clearpage
\subsection{Extended DiT learning curves}
\begin{figure}[!hbp]
    \centering
        \vspace{-8pt}
    \includegraphics[width=1.15\textwidth]{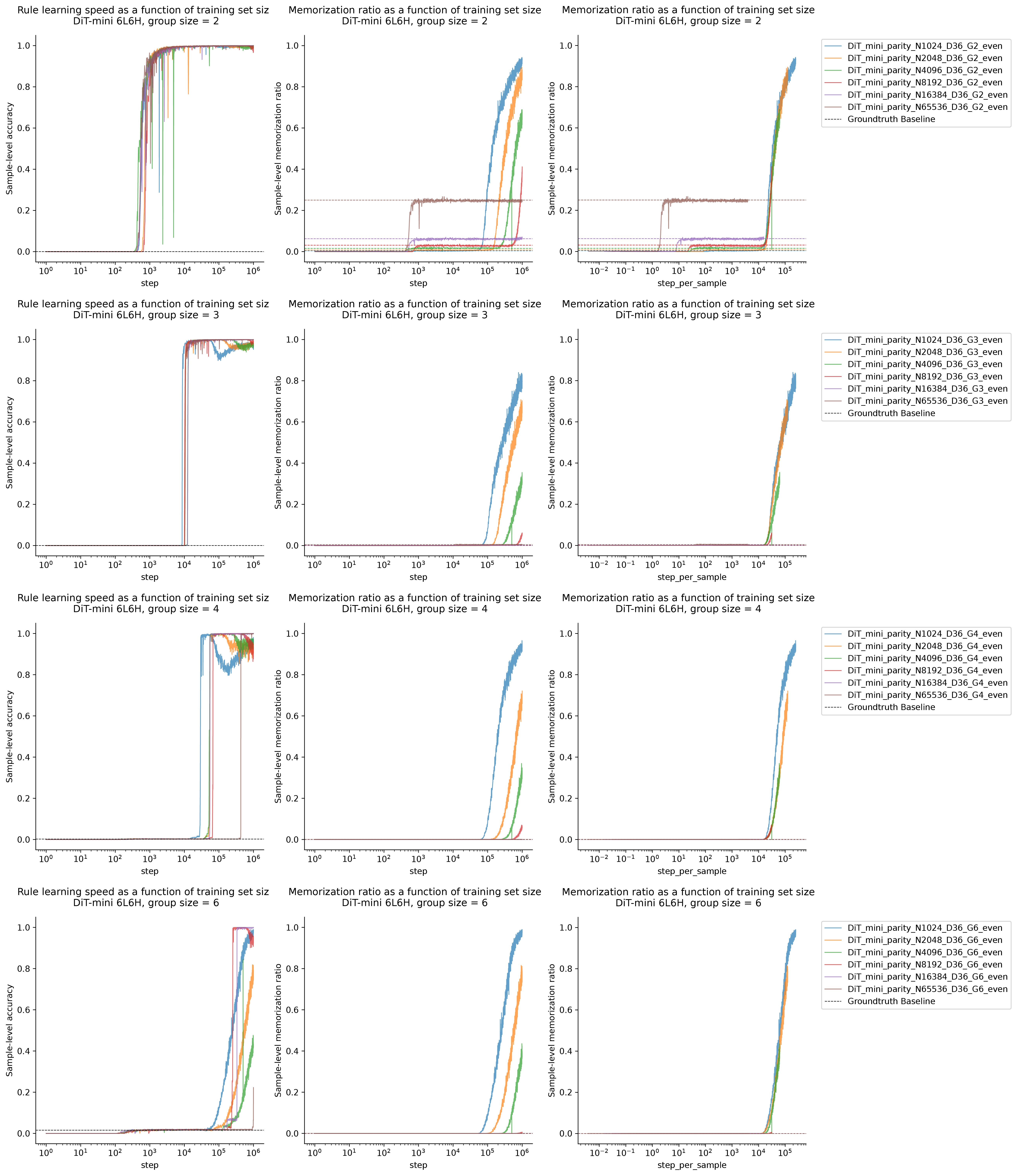}
        \vspace{-8pt}
    \caption{\textbf{Learning dynamics of rule acquisition and memorization across dataset size, $G=2,3,4,6$.} \textbf{Left.} Dynamics of sample parity accuracy across dataset scale, DiT-mini. \textbf{Mid. Right.} Dynamics of sample memorization ratio across dataset scales, the dynamics are plotted as a function of step (\textbf{Mid.}) and step per sample (step x batch size/ dataset size) (\textbf{Right.}). Colored dashed lines denotes the memorization ratio expected from the ground truth distribution.
    Per-$G$, per-$N$ extension of the trajectories aggregated in Fig.~\ref{fig:two_clocks_and_scaling}\textbf{G,H}.
    }
        \vspace{-8pt}
    \label{suppfig:scaling_law_dynamics_1}
\end{figure}

\begin{figure}[!hbp]
    \centering
        \vspace{-8pt}
    \includegraphics[width=1.15\textwidth]{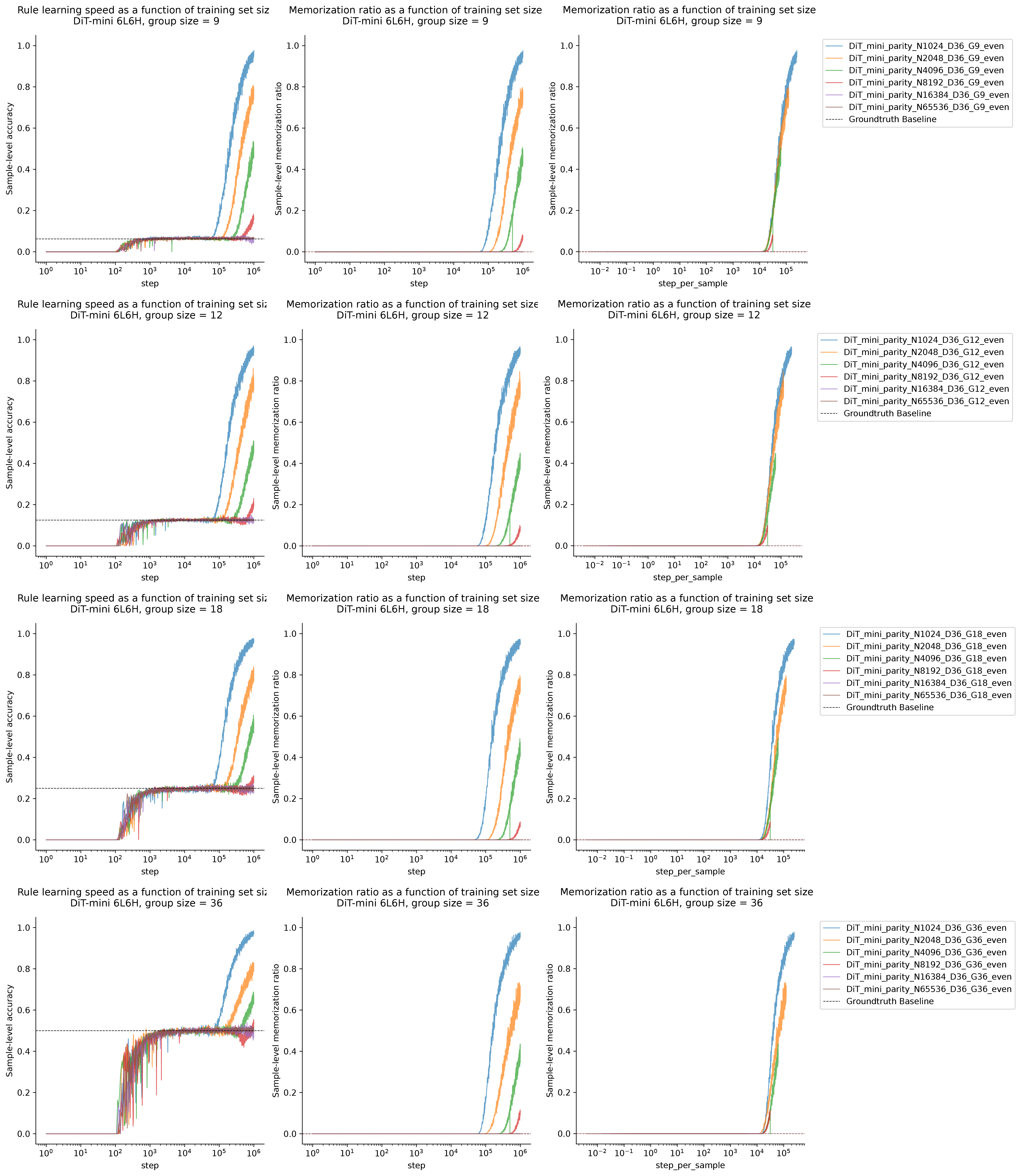}
        \vspace{-8pt}
    \caption{\textbf{Learning dynamics of rule acquisition and memorization across dataset size, $G=9,12,18,36$.}
    Similar format as Fig.~\ref{suppfig:scaling_law_dynamics_1}; per-$G$, per-$N$ extension of trajectories aggregated in Fig.~\ref{fig:two_clocks_and_scaling}\textbf{G,H}.
    }
        \vspace{-8pt}
    \label{suppfig:scaling_law_dynamics_2}
\end{figure}

\clearpage
\subsection{Extended memorization-time scaling for DiT and GPT} \label{app:tau_mem_scaling_extended}

\begin{figure}[!htp]
    \centering
    \includegraphics[width=\linewidth]{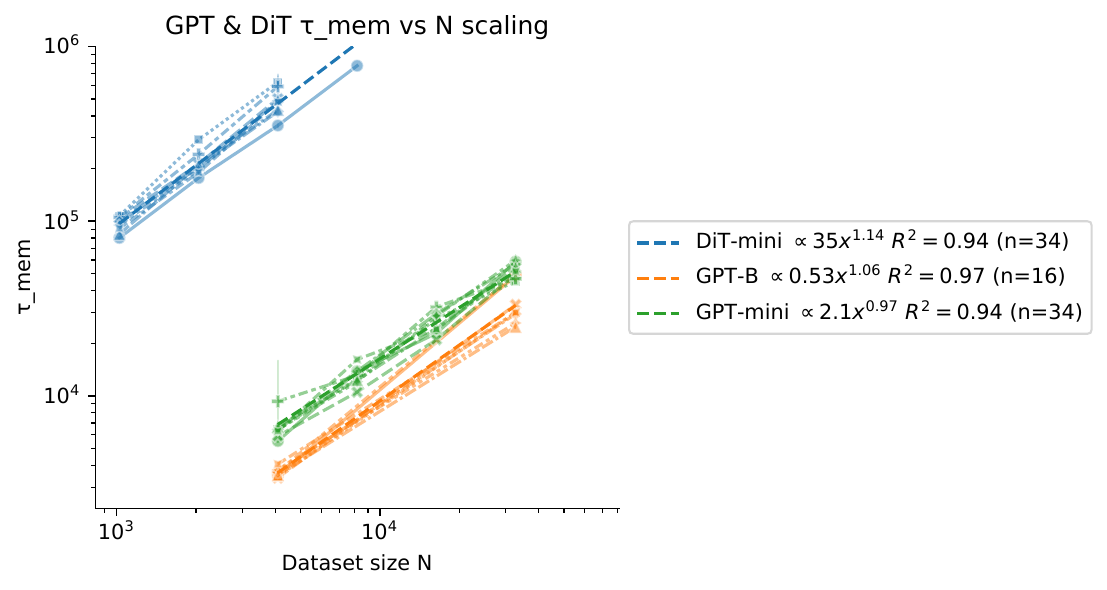}
    \caption{Memorization onset time $\tau_\mathrm{mem}$ as a function of training set size $N$, for all DiT and GPT model scales and group sizes $G$.
    Each point is one experiment; dashed lines show power-law fits $\tau_\mathrm{mem} \propto N^\alpha$ per model and $G$.
    The near-linear exponent $\alpha \approx 1$--$1.1$ is consistent across architectures and scales (see Table~\ref{tab:mem_scaling} for numerical values).
    DiT-mini exhibits systematically larger $\tau_\mathrm{mem}$ than GPT-mini at matched $N$, resulting a systematic upward shift of scaling curve. This suggests that the denoising objective is harder to memorize relative to next-token prediction. GPT-B further shifts the scaling downward, suggesting larger model exhibits faster memorization.}
    \label{suppfig:tau_mem_vs_N_scaling}
\end{figure}

Table~\ref{tab:mem_scaling} reports power-law fits $\tau_\mathrm{mem} \approx c \cdot N^\alpha$ for each model and group size, using the adaptive memorization threshold $\theta_\mathrm{mem} = 0.1 + N/\mathrm{support\_size}$ (see \S\ref{app:eval_method}).
Across models and all $G$ values pooled, the exponent $\alpha \approx 1.0$--$1.1$, indicating a near-linear scaling of memorization time with dataset size.

\begin{table}[h]
\centering\small
\caption{Power-law scaling of memorization onset $\tau_\mathrm{mem}$ with training set size $N$:
$\tau_\mathrm{mem} \approx c \cdot N^\alpha$.
Fit via log-log linear regression; $R^2$ and $n$ denote fit quality and number of data points.
$\tmem$ is measured with adaptive mem.\ threshold $= 0.1 + N/\mathrm{support\_size}$.}
\label{tab:mem_scaling}
\begin{tabular}{llrrrr}
\toprule
Model & $G$ & $c$ & $\alpha$ & $R^2$ & $n$ \\
\midrule
DiT-mini & \textbf{all} & 35.3 & 1.14 & 0.94 & 34 \\
 & $2$ & 45.8 & 1.08 & 1.00 & 5 \\
 & $3$ & 36.2 & 1.15 & 0.99 & 4 \\
 & $4$ & 19.0 & 1.25 & 0.99 & 4 \\
 & $6$ & 18.6 & 1.24 & 0.98 & 5 \\
 & $9$ & 70.2 & 1.05 & 1.00 & 4 \\
 & $12$ & 24.0 & 1.19 & 0.99 & 4 \\
 & $18$ & 26.4 & 1.17 & 1.00 & 4 \\
 & $36$ & 19.3 & 1.22 & 1.00 & 4 \\
\midrule
GPT-mini & \textbf{all} & 2.1 & 0.97 & 0.94 & 34 \\
 & $2$ & 0.6 & 1.10 & 0.99 & 4 \\
 & $3$ & 1.1 & 1.02 & 0.99 & 4 \\
 & $4$ & 2.2 & 0.96 & 0.99 & 4 \\
 & $6$ & 11.0 & 0.81 & 0.83 & 6 \\
 & $9$ & 1.0 & 1.05 & 1.00 & 4 \\
 & $12$ & 0.7 & 1.09 & 1.00 & 4 \\
 & $18$ & 1.4 & 1.01 & 1.00 & 4 \\
 & $36$ & 2.2 & 0.97 & 0.98 & 4 \\
\midrule
GPT-B & \textbf{all} & 0.5 & 1.06 & 0.97 & 16 \\
 & $2$ & 0.1 & 1.27 & 1.00 & 2 \\
 & $3$ & 0.4 & 1.10 & 1.00 & 2 \\
 & $4$ & 0.6 & 1.04 & 1.00 & 2 \\
 & $6$ & 0.1 & 1.25 & 1.00 & 2 \\
 & $9$ & 0.9 & 1.00 & 1.00 & 2 \\
 & $12$ & 1.7 & 0.93 & 1.00 & 2 \\
 & $18$ & 1.4 & 0.94 & 1.00 & 2 \\
 & $36$ & 1.4 & 0.96 & 1.00 & 2 \\
\bottomrule
\end{tabular}
\end{table}



\clearpage


\subsection{Dissecting rule learning and memorization dynamics}\label{app:dissection_extended}

\subsubsection{Per-sample training dynamics}\label{app:per_sample_dynamics_analysis}

For sampling, we used a fixed set of $2048$ initial Gaussian noise samples, and the deterministic Heun's sampler. Thus due to the continuous dependency of the sample on the PF-ODE parameters, individual sample trajectories are comparable over training time.
At each checkpoint, each generated sample is assigned to one of four mutually exclusive states (see \S\ref{app:eval_method} for definitions): \qinv{invalid (quantization-ambiguous)}, \rinv{invalid (rule-error)}, \nov{valid \& novel}, and \mem{valid \& memorized}.
This yields a per-sample state trajectory of shape $(T \times N)$ over the $T$ saved checkpoints.

\begin{figure}[!htp]
    \centering
    \includegraphics[width=\linewidth]{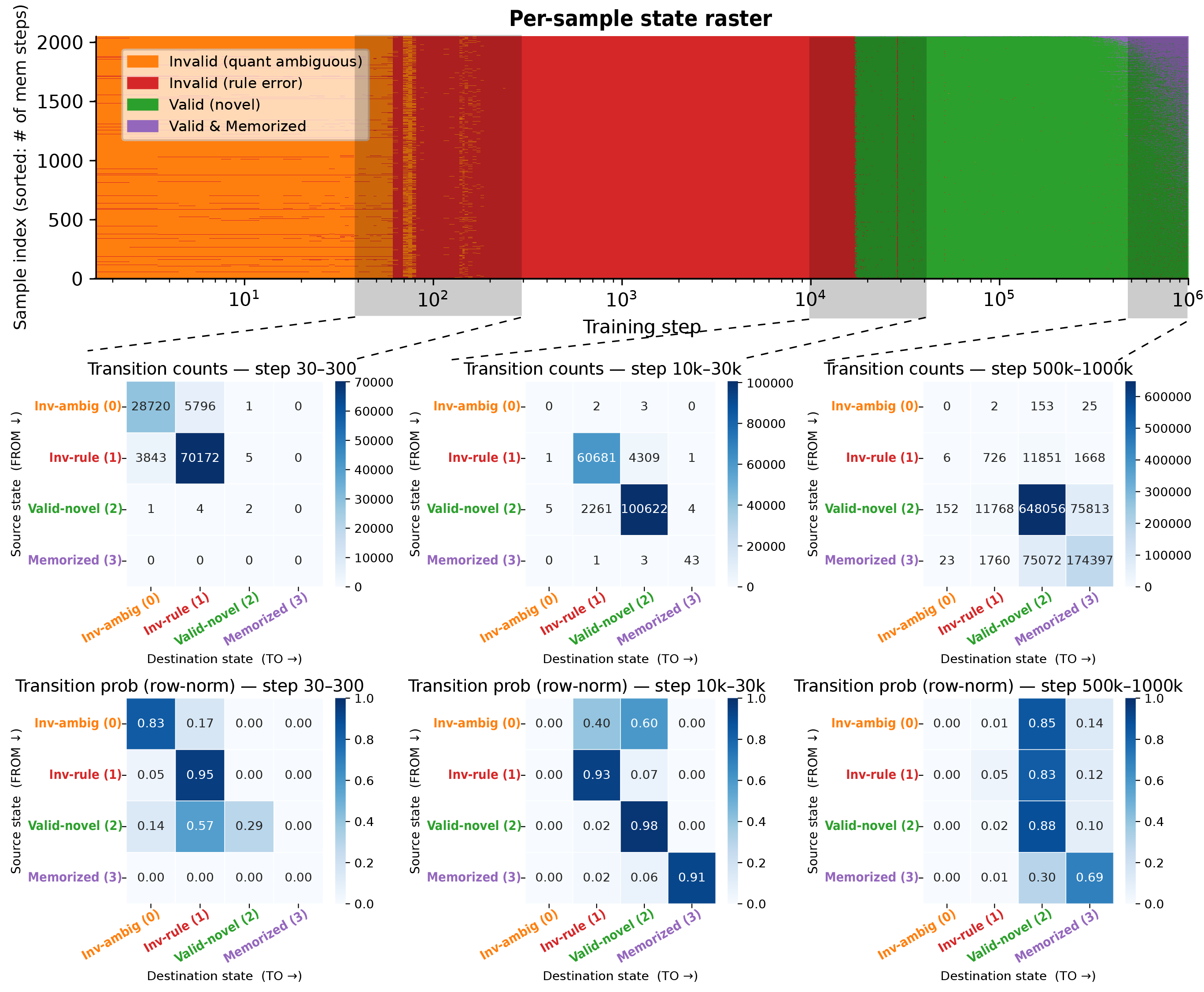}
    \caption{
\textbf{Per-sample state transitions reveal staged diffusion learning.}
\textbf{Top}: raster plot tracking the state of each generated sample across training, sorted by the step at which it first becomes memorized, same panel as Fig.~\ref{fig:diffusion_rule_mem_dissection}\textbf{A}. 
\textbf{Middle}: transition count matrices between sample states in three representative training windows: early quantization transition (steps 30--300), rule-learning transition (steps 10k--30k), and late memorization onset (steps 500k--1000k). Rows denote source states and columns denote destination states between consecutive evaluation checkpoints.
\textbf{Bottom}: row-normalized transition probabilities for the same windows. Early in training, samples primarily move from quantization-invalid to rule-invalid states; during rule learning, rule-invalid samples transition into the valid-novel state; late in training, valid-novel samples increasingly transition into the memorized state. Together, these dynamics show that memorization is preceded by a broad valid-but-novel phase rather than emerging directly from invalid generations.
}
    \label{suppfig:state_transition_markov}
\end{figure}

\paragraph{Transition matrix analysis.}
From consecutive checkpoint pairs $(t, t{+}1)$ we compute a $4{\times}4$ transition count tensor $\mathbf{T}(t)$, where $\mathbf{T}[t,i,j]$ is the number of samples in state $i$ at step $t$ that move to state $j$ at step $t{+}1$. 
Row-normalizing gives the empirical transition probability matrix. 
To characterize different training phases, we aggregate $\mathbf{T}(t)$ over key time windows—pre-rule-learning, rule-learning, and memorization—by summing counts within each window (Fig.~\ref{suppfig:state_transition_markov}). 
We also animate the heatmap over the full training trajectory to reveal how transition dynamics evolve continuously.

One salient observation from Fig.~\ref{suppfig:state_transition_markov} is that, at the late memorization phase, looking at the transition probability matrix (Markov matrix), we can see \rinv{invalid (rule-error)} is \textit{not an attractor state}, i.e. samples entering it will jump to \nov{valid \& novel}, and \mem{valid \& memorized} in the next model state. 
In contrast, \nov{valid \& novel} and \mem{valid \& memorized} are both attractor states, with some transition probability between them. 
We note that the valid $\to$ memorized transition slightly outnumbered the reverse direction causing the raw increase of memorized samples. 
This suggests the picture that the memorized samples are attractor states and tend to stay memorized, while the rule violation is transient state, and likely some "side effect" of the landscape change. 

\begin{figure}[!htp]
    \centering
    \includegraphics[width=\linewidth]{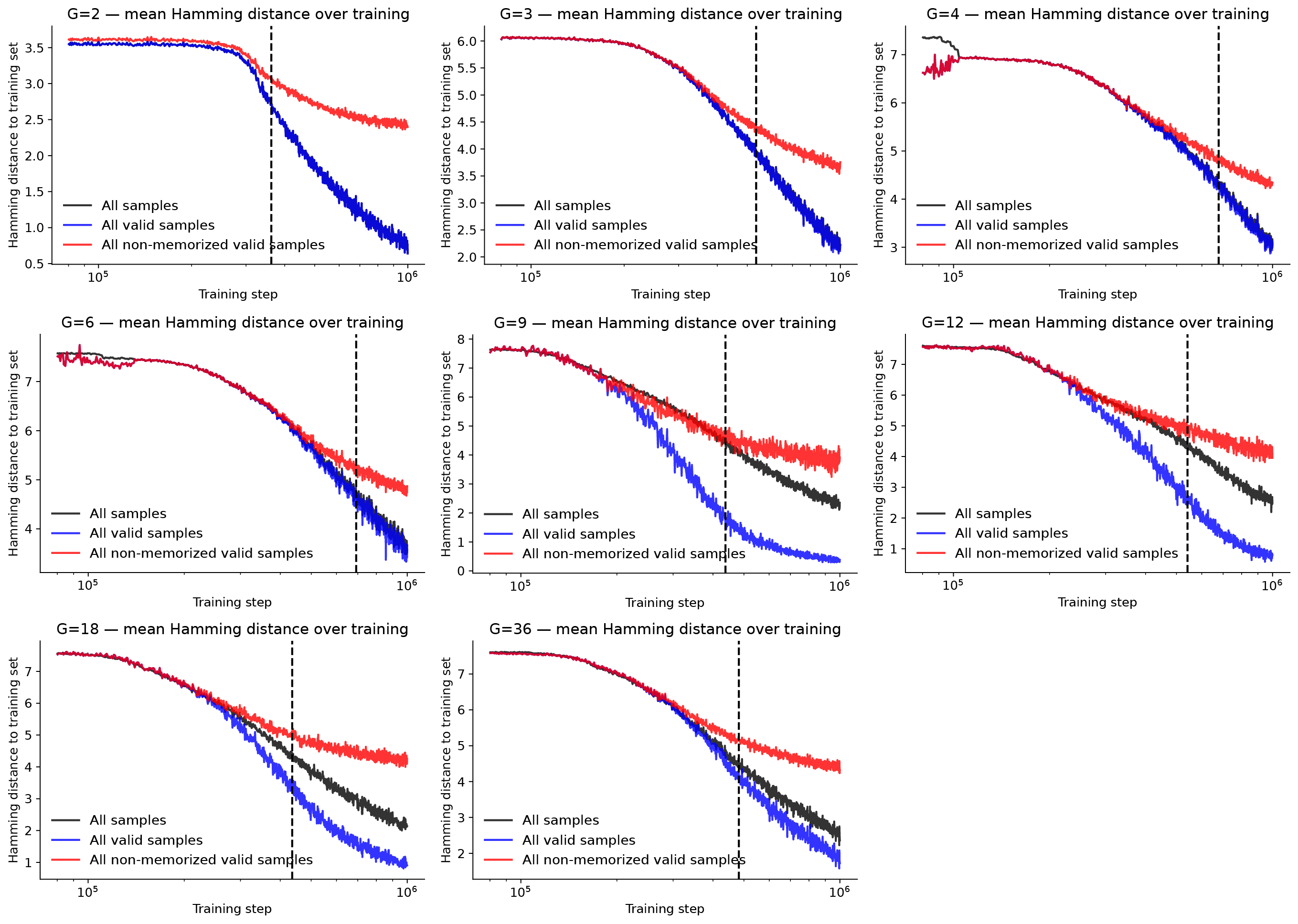}
\caption{
\textbf{Valid generations move progressively closer to the training set before memorization.}
Mean Hamming distance from generated samples to the closest training set sample over late training, shown separately for each parity group size $G$. 
Black curves show the average over all generated samples; blue curves show the average restricted to rule-valid samples; red curves show rule-valid but non-memorized samples.
Dashed vertical lines mark the memorization onset $\tmem$ with adaptive threshold memorization ratio 0.1. \\
As training proceeds, the nearest-neighbor Hamming distance of valid samples decreases sharply and continues to decline around $\tmem$.
Importantly, the red curves also decrease, showing that even non-memorized valid samples become increasingly close to training samples before exact memorization.
Thus, the late training phase is not an abrupt jump from novelty to memorization, but a gradual contraction of valid generations toward the empirical training set.
}
    \label{suppfig:haming_over_training}
\end{figure}

\paragraph{Hamming distance conditioned on sample type.}
Beyond the four states, for each sample, we also record its Hamming distance to the nearest training-set member. 
We then condition this statistic on specific sample types, (i) all samples, (ii) all rule valid samples, (iii) \nov{rule valid and novel samples} (i.e. excluding hamming 0 case); and then track statistics of these distance across training time. 
Across all $G$ rules, we find that the distance to training set of all three types decay at the late phase of training ($[1E5,1E6]$); and this decay is evident even for \nov{valid and novel} samples, showing that even before exact memorization, the generated samples get closer to the training point cloud in the late phase. 
Consistently, this decay of hamming distance precedes the onset of memorization $\tmem$ measured from sample level (Fig.~\ref{suppfig:haming_over_training}). 
Thus it shows that at the late phase, the training points have increasingly strong "attraction force", pulling the novel generations towards the ones closer to training data. 

\clearpage
\subsubsection{DSM loss by noise scale}\label{app:dsm_per_noise_scale}

We evaluate the DSM loss on three held-out data splits at each checkpoint:
\begin{enumerate} 
    \item \textbf{Training set}: the $N{=}4096$ samples used during training.
    \item \textbf{Valid-novel test set}: $N{=}4096$ rule-valid samples drawn from the same group-parity distribution but verified to be non overlapping with training sample (i.e.\ not memorizable). %
    \item \textbf{Boolean-cube samples}: $N{=}4096$ uniformly random $\{{\pm}1\}^D$ binary vectors with no parity constraint. 
\end{enumerate}
For model checkpoint we compute the unweighted DSM loss with respect to each split and evaluate over a fixed grid of noise levels $\sigma \in [\sigma_{min},\sigma_{max}] =[0.002, 80.0]$ (log-spaced, 50 levels), giving a noise-scale-vs-step matrix per data split. 

\begin{figure}[!htp]
    \centering
    \includegraphics[width=\linewidth]{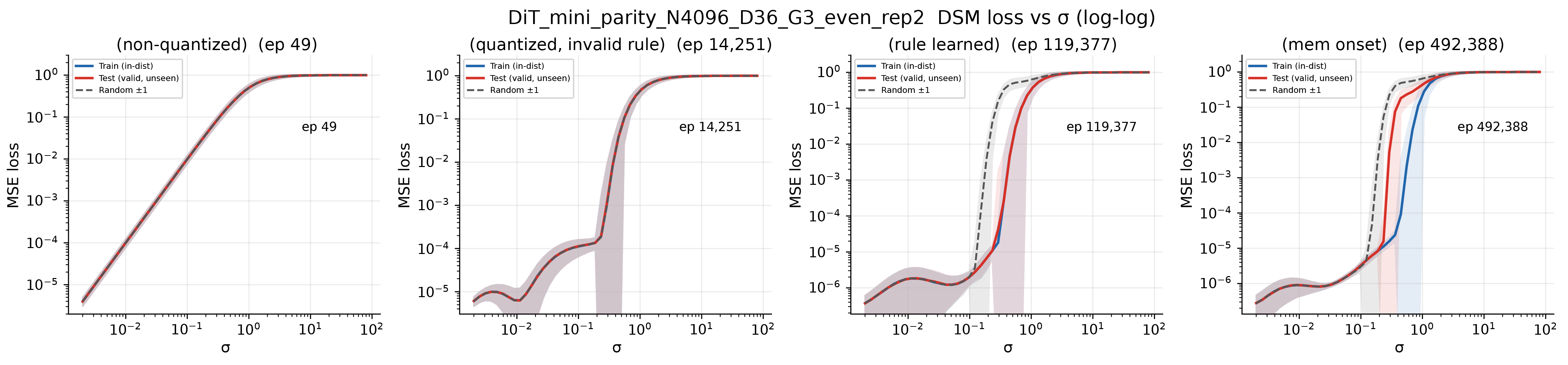}
    \caption{
\textbf{Noise-scale spectra at representative training phases.}
DSM loss as a function of $\sigma$ is shown at the four checkpoints: early non-quantized generation, quantized but rule-violating generation, rule-learned generation, and memorization onset (same as Fig.~\ref{fig:diffusion_rule_mem_dissection}\textbf{D}).
Train, held-out valid, and random Boolean-cube samples are nearly indistinguishable before rule learning.
After rule learning, the random Boolean-cube spectrum separates from train and held-out valid loss spectrum; at memorization onset, the training spectrum separates from the held-out valid spectrum.
Both separations are most pronounced at only a narrow range of noise scales $[0.1,2.0]$ for rule learning, $[0.2,2.0]$ for memorization, supporting the view that rule learning and memorization correspond to vector-field changes at a specific spatial and noising scale.
}
    \label{suppfig:dsm_per_sigma_critical_noise_scale}
\end{figure}

We visualize these matrices through several slices. 
We plot the full loss-vs-$\sigma$ spectrum (log–log scale) at selected checkpoints to reveal \emph{at which noise scales} the three splits diverge (Fig.~\ref{fig:diffusion_rule_mem_dissection}\textbf{C},~\ref{suppfig:dsm_per_sigma_critical_noise_scale}). These spectral view identifies the noise window that is critical for rule learning ($[0.1,2.0]$) and memorization ($[0.2,2.0]$). 

\begin{figure}[!htp]
    \centering
    \includegraphics[width=0.75\linewidth]{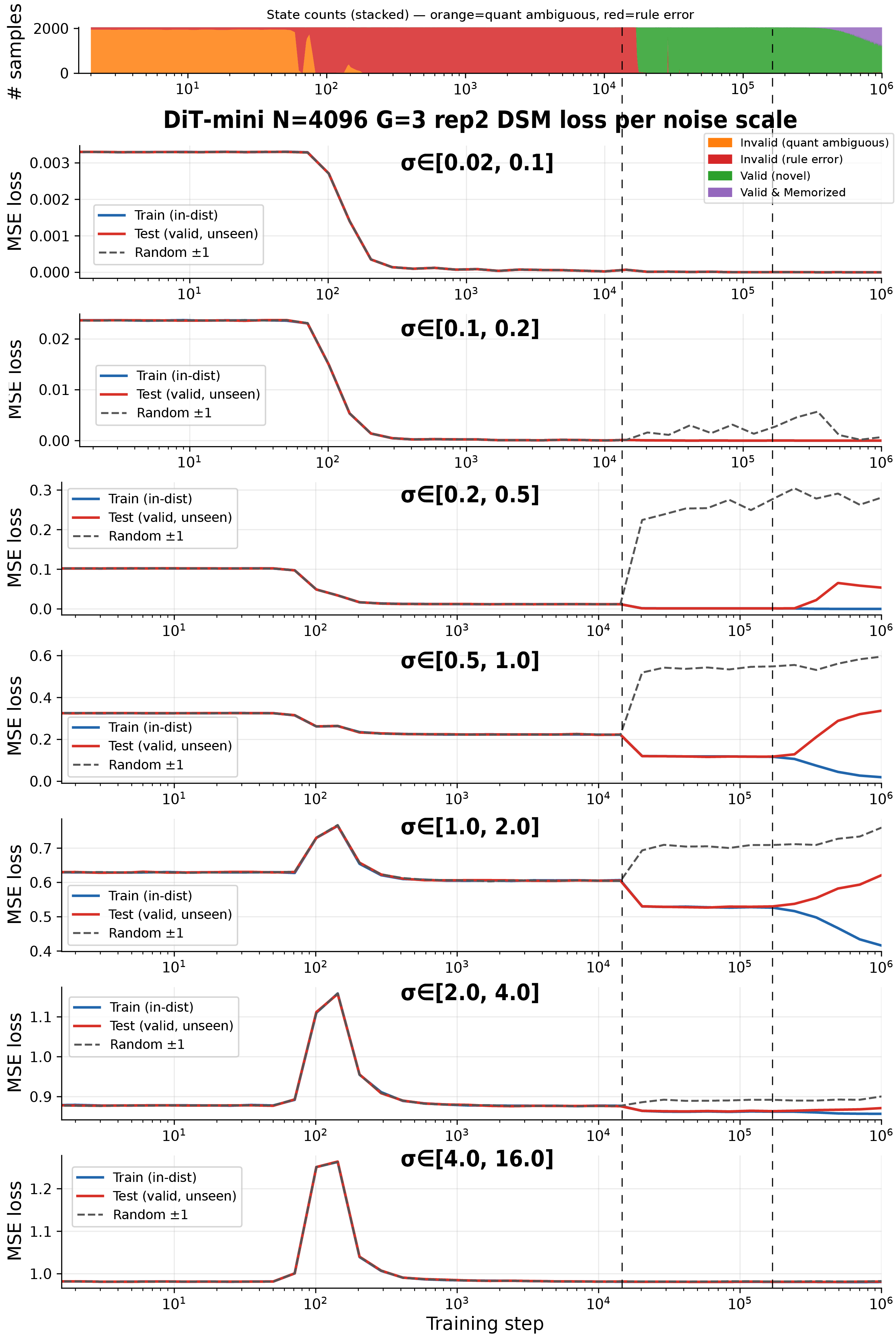}
    \caption{
\textbf{DSM loss dynamics across noise-scale bands.}
For DiT-mini trained on parity with $G{=}3$, $N{=}4096$ (run 2), we track DSM losses for training samples, held-out rule-valid samples, and random Boolean-cube samples, separately across logarithmic noise-scale bands.
The top panel shows the corresponding sample-state counts across training (Fig.~\ref{fig:diffusion_rule_mem_dissection}\textbf{A}).
Low-noise losses quickly collapse across splits as the model learns the Boolean cube, while intermediate noise scales show the parity-task relevant separations: random Boolean-cube loss separates from train/test around rule learning, and train/test losses separate around memorization. 
Vertical dashed lines mark train-random and train-test loss splitting times. Note that the DSM loss split times preceeds the sample level rule learning and memorization which is used to define $\trule$ and $\tmem$ in the main paper. 
This makes sense, since the change of vector field is immediately reflected in the DSM loss, and the samples are integrated result of vector field, thus the change is evident there later. 
In the future work, one could define $\trule$ and $\tmem$ according to DSM loss splits. 
}
    \label{suppfig:dsm_per_sigma_curves_overview}
\end{figure}

At these critical noise scales, we also plot the three DSM loss traces across training time to observe the onset of the splits (Fig.~\ref{fig:diffusion_rule_mem_dissection}\textbf{B},~\ref{suppfig:dsm_per_sigma_curves_overview}).  
The phase transitions identified from samples can also be identified from the loss: 
The rule-learning transition is identified as the step where the boolean-cube loss diverges from the train and valid-novel curves; the memorization onset is identified as the step where the training and valid-novel curves diverge from each other.
Notably, the DSM loss splits happen before the corresponding $\trule$ and $\tmem$ identified from samples. 

\begin{figure}[!htp]
    \centering
    \includegraphics[width=\linewidth]{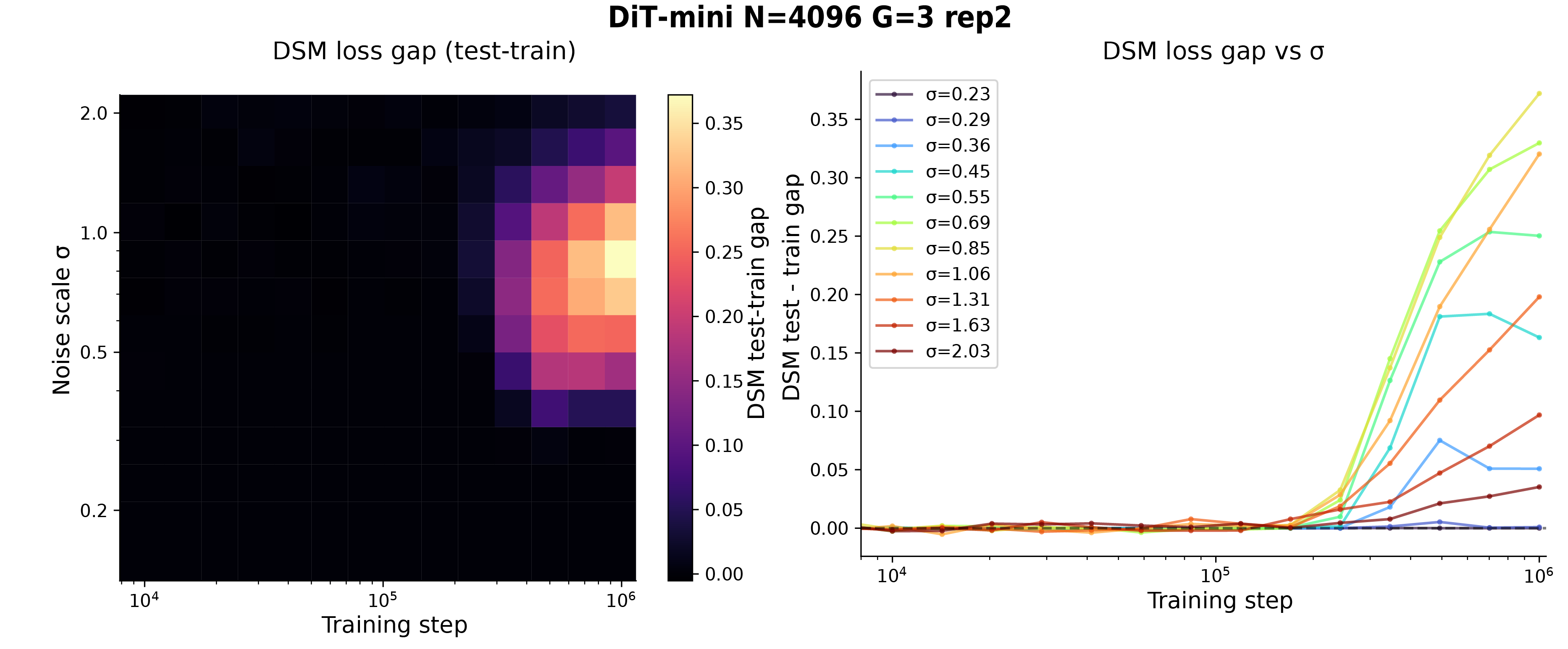}
    \caption{
\textbf{Onset of Train--test DSM loss gap zooming in the critical noise scales $[0.2,2.0]$.}
\textbf{Left}: heatmap of the DSM loss gap between held-out rule-valid samples and training samples as a function of training step and noise scale $\sigma$, focusing on the critical noise scale.
\textbf{Right}: the same gap plotted over training for representative fixed $\sigma$ values.
We note that the train-test gap first opens for noise scales around $\sigma\in[0.5,1.0]$, then this gap propagates and affects the lower and higher noise scales. In the end, the gap in $\sigma\in[0.5,1.0]$ is still the most pronounced. 
}
    \label{suppfig:dsm_per_sigma_heatmap_overview}
\end{figure}

Finally, visualizing the train-test difference of the DSM loss (i.e. generalization gap), we can trace the onset of the split on the noise-level $\times$ training step plane (Fig.~\ref{suppfig:dsm_per_sigma_heatmap_overview}). One can see the splits first happens in $\sigma\sim [0.5,1.0]$ then propagates to the lower and higher noise levels. 
We hypothesize that the attractor landscape first changes at $\sigma\sim1.0$, then due to the shared parametrization of denoiser, the change propagates to landscapes at nearby noise scales, causing the enlargement of the basins at $\sigma\sim0.5$. 

\clearpage
\subsubsection{Vector-field and attractor-basin analyses}\label{app:vector_field_basin}

\paragraph{Plane definition.}
All vector-field and landscape visualizations are performed in a 2D affine subspace of $\mathbb{R}^D$ anchored at a training sample $\mathbf{x}_a$.
Two companion anchor points define the spanning vectors:
$\mathbf{x}_b$ is obtained from $\mathbf{x}_a$ by flipping bits 0 and 1 simultaneously (Hamming-2 flip, producing a \emph{valid-novel} neighbor), and
$\mathbf{x}_c$ is obtained by flipping bit 0 alone (Hamming-1 flip, producing an \emph{invalid} neighbor).
The two plane vectors are $\mathbf{v}_{ab} = \mathbf{x}_b - \mathbf{x}_a$ and $\mathbf{v}_{ac} = \mathbf{x}_c - \mathbf{x}_a$, Gram--Schmidt orthonormalized.
A $50{\times}50$ grid spanning $\alpha, \beta \in [-1.75,\, 3.75]$ is constructed as
$\mathbf{x}(\alpha,\beta) = \mathbf{x}_a + \alpha\,\mathbf{v}_{ab} + \beta\,\mathbf{v}_{ac}$,
and the neural network denoiser $D_\theta(\mathbf{x},\sigma)$ is evaluated at each grid point at representative noise scales $\sigma$, then we can derive the score via Tweedie's formula $D_\theta(\mathbf{x},\sigma) = \mathbf{x} + \sigma^2 s_\theta(\mathbf{x},\sigma)$. 

We note that these four points form a face of the boolean hypercube, and their coordinates under the normal element-wise basis should be $(-1,-1),(-1,1),(1,-1),(1,1)$. 
The planes showed in Fig.~\ref{fig:diffusion_rule_mem_dissection}\textbf{D} uses $x_a,x_b$ line as the x-axis, thus the coordinate system is rotated and shifted, but not scaled. In this 2d plan, we know the exact coordinates of the four anchors $x_a=(0,0);x_b=(2\sqrt2,0),x_c=(\sqrt2,\sqrt2),x_d=(\sqrt2,-\sqrt2)$. 
This geometry also defines the critical distance of our parity datasets: the distance to nearest rule invalid sample is 2 (Hamming 1); the distance to the nearest valid and novel sample is (usually) $2\sqrt{2}$ (Hamming 2). 

\paragraph{Visualization.}
Each checkpoint at each noise scale $\sigma$ is displayed as a column of two rows:

\textbf{Top row — score magnitude and displacement field.}
The heatmap shows the $\ell_2$ norm of the full-dimensional score vector $s_\theta(\mathbf{x},\sigma)$ at each grid point, rendered in the \texttt{magma} colormap (bright = high score magnitude). Thus the darker color, i.e. close to zero score magnitude suggests attractor of the dynamics $D_\theta(\x,\sigma)=\x$. 
Overlaid white arrows show the 2D projection of the score vector $s_\theta(\mathbf{x},\sigma)$ onto the $(\mathbf{v}_{ab},\mathbf{v}_{ac})$ plane, subsampled for clarity.
Colored markers indicate the four anchor points: $\mathbf{x}_a$ (white, training sample), $\mathbf{x}_b$ (cyan, valid-novel Hamming-2 neighbor), $\mathbf{x}_c$ (yellow, invalid Hamming-1 neighbor), and $\mathbf{x}_d$ (orange, the mirror of $\mathbf{x}_c$ across $\mathbf{x}_a$).

\textbf{Bottom row — signed attraction along the $\mathbf{v}_{ab}$ axis.}
The heatmap shows the scalar projection of the denoiser onto $\mathbf{v}_{ab}$, i.e.\ $D_\theta(\mathbf{x},\sigma)\cdot \mathbf{v}_{ab}$, in a diverging \texttt{RdBu\_r} colormap. Due to the specific geometry of the four points on the face of hypercube, we can read out the target of attraction easily from the color: blue = attracted toward $\mathbf{x}_a$; red = attracted towards $\mathbf{x}_b$; white = attracted towards $\mathbf{x}_c$ or $\mathbf{x}_d$.
This row directly visualizes the basin and the boundary separating the $\mathbf{x}_a$ attractor from the $\mathbf{x}_b,\mathbf{x}_c,\mathbf{x}_d$ attractors. 

\begin{figure}[!htp]
    \centering
    \includegraphics[width=0.8\linewidth]{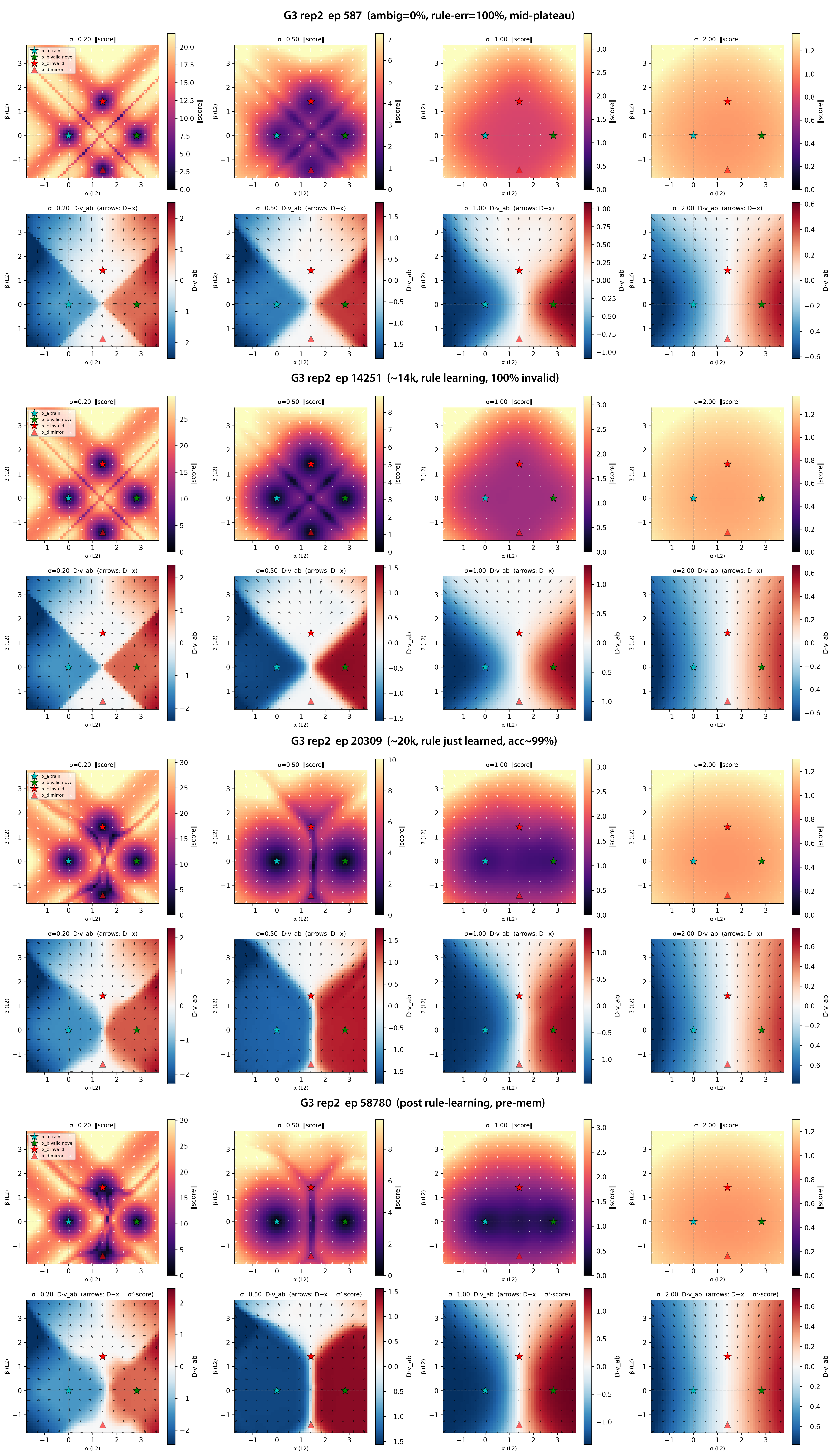}
    \caption{\textbf{Early vector-field landscape evolution.}
    Score magnitude (top row of each checkpoint) and signed denoising displacement along $\mathbf{v}_{ab}$ (bottom row) are shown for $\sigma\in\{0.20,0.50,1.00,2.00\}$.
    Before rule learning, the landscape first forms attractors around Boolean-cube vertices; after rule learning begins, the valid anchors $\mathbf{x}_a$ and $\mathbf{x}_b$ start to dominate the rule-violating anchors $\mathbf{x}_c$ and $\mathbf{x}_d$.}
    \label{suppfig:landscape_vector_field_early}
\end{figure}

\begin{figure}[!htp]
    \centering
    \includegraphics[width=\linewidth]{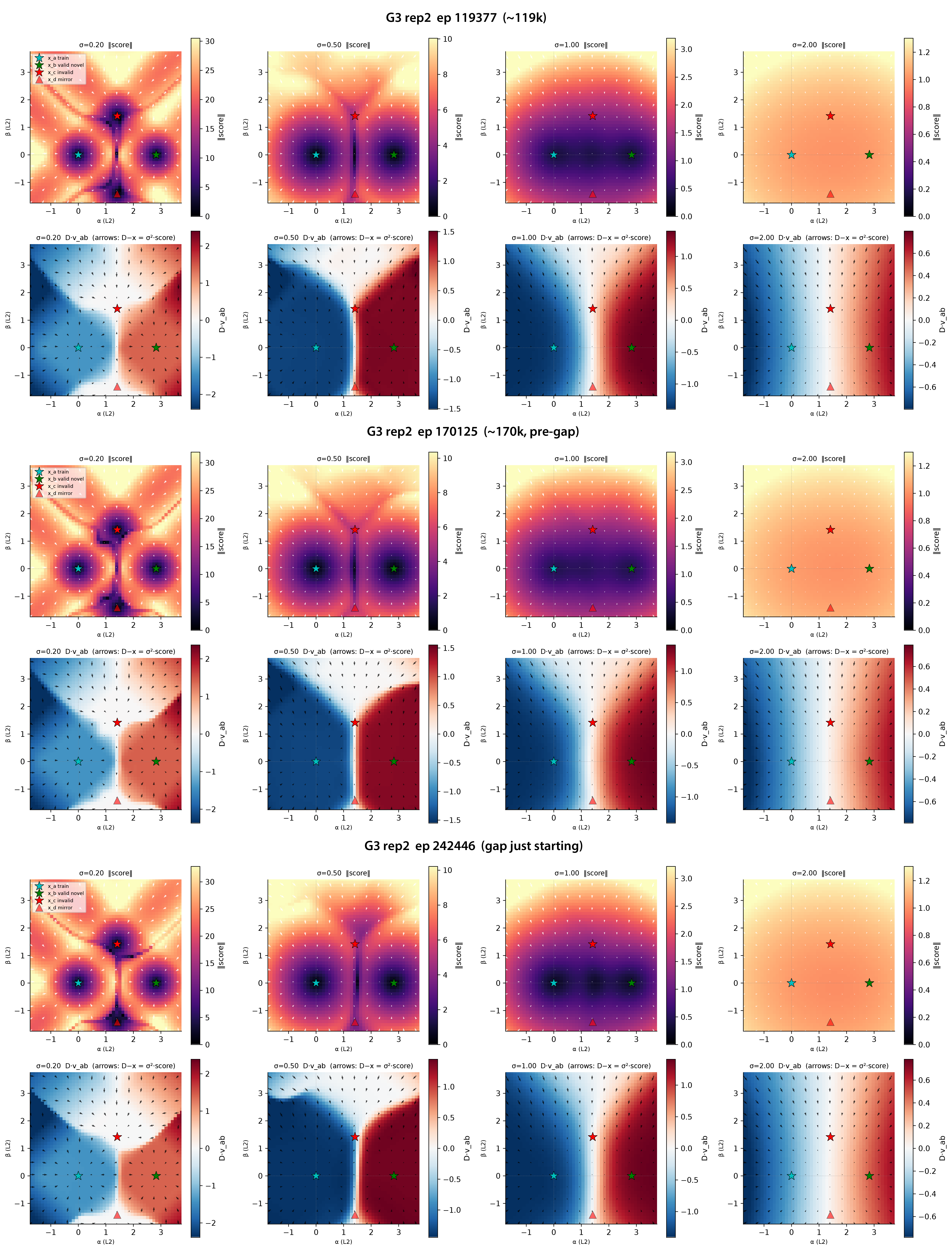}
    \caption{\textbf{Mid-training vector-field landscape evolution.}
    The same 2D slice and noise scales are tracked through the generalizing phase.
    The basin boundary between the training sample $\mathbf{x}_a$ and valid-novel neighbor $\mathbf{x}_b$ remains visible while the attraction toward $\mathbf{x}_a$ progressively strengthens, especially at intermediate noise scales.}
    \label{suppfig:landscape_vector_field_mid}
\end{figure}

\begin{figure}[!htp]
    \centering
    \includegraphics[width=\linewidth]{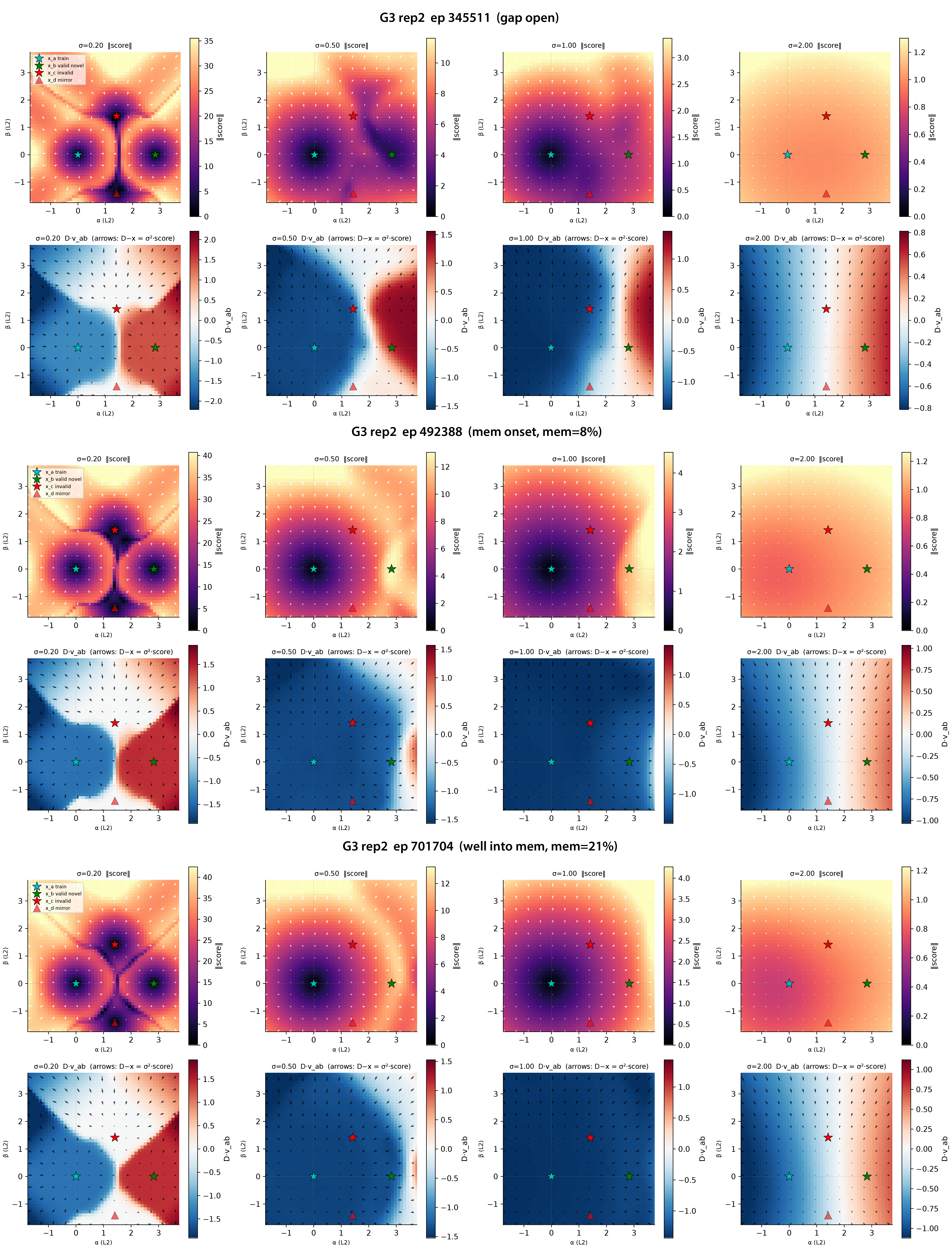}
    \caption{\textbf{Late vector-field landscape evolution.}
    During memorization onset and late memorization, the attraction basin of the training sample $\mathbf{x}_a$ expands toward $\mathbf{x}_b$ and eventually dominates much of the plane.
    The effect is strongest around the same intermediate noise scales where the DSM train--test gap opens.}
    \label{suppfig:landscape_vector_field_late}
\end{figure}

The same plane, grid, and anchor points are reused identically across all checkpoints (step\,$\approx$\,50, 14k, 119k, 492k for Fig.~\ref{fig:diffusion_rule_mem_dissection}\textbf{D}); grid evaluations are cached by a hash of the plane geometry and $\sigma$ value to avoid redundant forward passes.
A more densely sampled landscape view across the training process is shown in Figs.~\ref{suppfig:landscape_vector_field_early}--\ref{suppfig:landscape_vector_field_late}.

\paragraph{Attractor basin profile quantification.}

\begin{figure}[h]
    \centering
    \includegraphics[width=\linewidth]{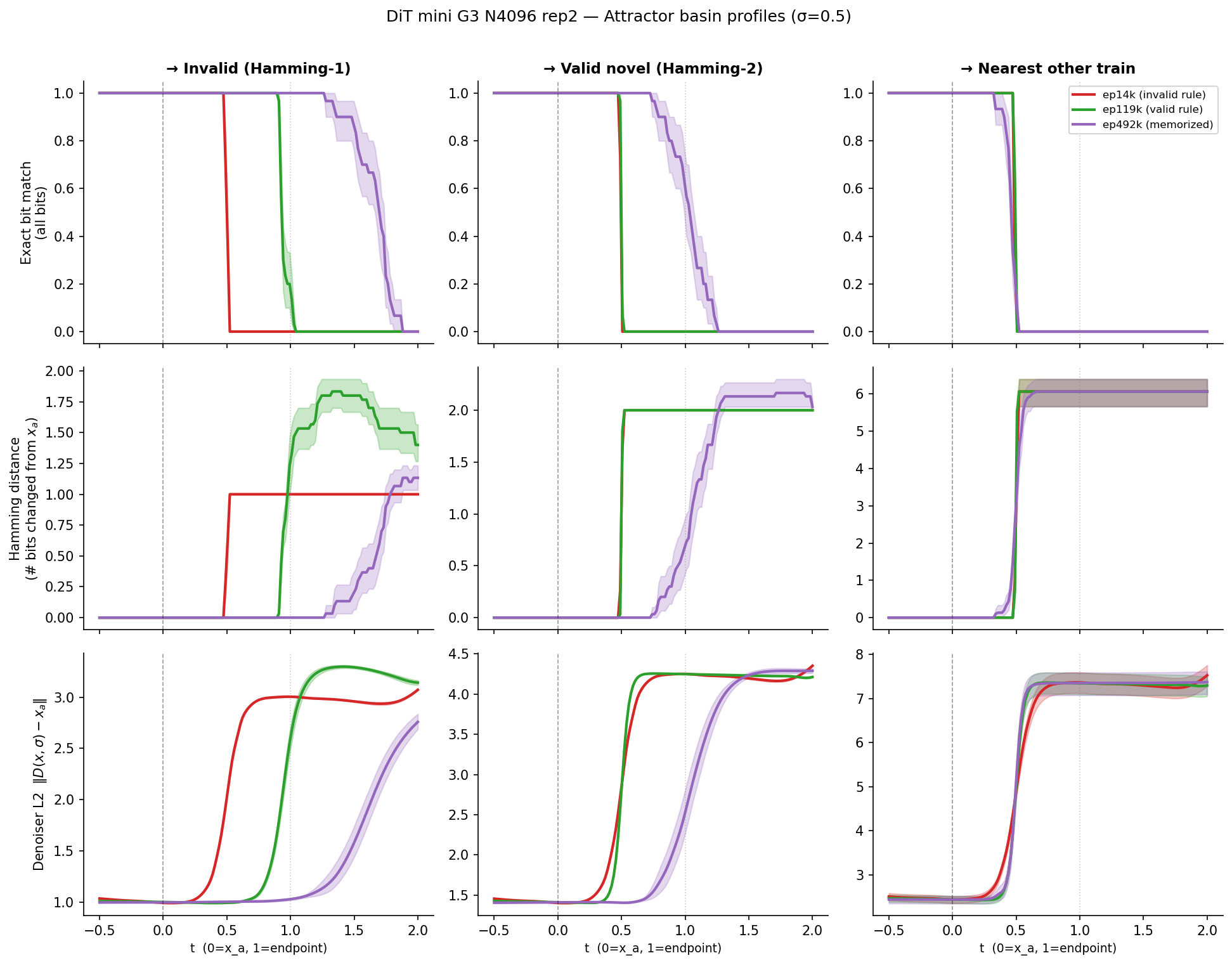}
    \caption{\textbf{Attractor-basin profiles across training phases.}
    For a DiT-mini model trained on group parity ($G{=}3$, $N{=}4096$), we probe 1D line segments from a training sample $x_a$ toward three endpoints: an invalid Hamming-1 neighbor (left), a valid-novel Hamming-2 neighbor (middle), and the nearest other training sample (right).
    Rows show exact-bit match to $x_a$, Hamming distance from $x_a$, and denoiser $\ell_2$ distance $\|D(x,\sigma)-x_a\|$ at $\sigma{=}0.5$.
    The basin around $x_a$ expands from the rule-error phase (14k) to the rule-learned phase (119k), and expands further at memorization onset (492k), especially toward valid-novel neighbors.}
    \label{suppfig:basin_profiles}
\end{figure}

To quantify the spatial extent of the attractor basin around a training sample $\mathbf{x}_a$, we evaluate the denoiser along 1D line segments through $\mathbf{x}_a$ in three fixed directions:
(i)~toward a Hamming-1 invalid neighbor $\mathbf{x}_\text{inv}$ (one bit flipped, breaking group parity);
(ii)~toward a Hamming-2 valid-novel neighbor $\mathbf{x}_\text{nov}$ (two bits within the same group flipped, parity preserved, not in the training set);
(iii)~toward the nearest other training sample $\mathbf{x}_\text{train}$ (by Hamming distance).
Each line is parameterized as $\mathbf{x}(t) = \mathbf{x}_a + t\,(\mathbf{x}_\text{end} - \mathbf{x}_a)$,
with $t$ ranging from $-0.5$ (behind $\mathbf{x}_a$) to $2.0$ (past the endpoint), sampled at $n{=}150$ uniformly spaced values.
At each point $\mathbf{x}(t)$ the denoiser output $D_\theta(\mathbf{x}(t),\sigma)$ is computed.
Three basin metrics are derived:
\begin{enumerate}
    \item \textbf{Exact match}: $\mathbf{1}[\operatorname{sign}(D_\theta(\mathbf{x}(t))) = \operatorname{sign}(\mathbf{x}_a)]$ — binary indicator that all output bits (after binarization) agree with $\mathbf{x}_a$; collapses to 0 when the denoiser is attracted to a different vertex.
    \item \textbf{Hamming distance}: $\sum_i \mathbf{1}[\operatorname{sign}(D_{\theta,i}) \neq \operatorname{sign}(x_{a,i})]$ — number of bits that disagree, a graded measure of basin exit.
    \item \textbf{L2 distance from start}: $\|D_\theta(\mathbf{x}(t)) - \mathbf{x}_a\|_2$ — Euclidean distance of the denoised output from the basin center.
\end{enumerate}
The procedure is repeated independently for $N_\text{samp}{=}30$ different anchor samples $\mathbf{x}_a$ drawn from the training set.
All three metrics are averaged across anchors, and bootstrap confidence intervals (5th--95th percentile of the bootstrap distribution of the mean, with $2000$ resamples) are shown as shaded bands.
Results are reported at $\sigma{=}0.5$ for three representative checkpoints: ep\,14k (rule-error phase), ep\,119k (generalization phase), and ep\,492k (memorization phase), in Fig.~\ref{suppfig:basin_profiles}.

\clearpage
\subsection{Autoregressive vs diffusion supplementary results}
\label{app:gpt_vs_dit}

\paragraph{Generation paradigm.}
GPT generates bits sequentially in a fixed left-to-right order, while DiT generates all $D$ bits jointly in a single diffusion process (conditioned on noise level $\sigma$).
This fundamental difference leads to distinct inductive biases: GPT can exploit causal structure (later bits can be predicted from earlier ones), whereas DiT treats all bit positions symmetrically.
In accordance with this difference, the GPT models employ causal attention masks, while DiT use all-to-all self attention. 

\paragraph{Structural consequence for parity learning.}
For even-parity groups of size $G$, the last bit of each group is fully determined by the preceding $G{-}1$ bits.
In the GPT, this determinism is directly exploited during next-token prediction: once the model learns the parity rule, the CE at positions $G, 2G, \ldots, D$ collapses to zero first, before other positions.
The DiT, by contrast, has no privileged ordering and learns to jointly enforce the constraint across all positions simultaneously, making its rule learning detectable as a collective change in the score field at a specific noise level $\sigma$ rather than a position-specific collapse. 

\paragraph{Shared qualitative dynamics.}
Despite these architectural differences, both models exhibit the same two-phase dynamics:
(i) an initial rule-learning transition in which validation-set performance improves sharply, and
(ii) a subsequent memorization transition in which training-set performance diverges from held-out performance.
The timescales differ: GPT-mini reaches both transitions within $10^5$ gradient steps, while DiT-mini requires up to $10^6$ steps due to the harder denoising objective and the continuous-noise parameterization.
Both models are compared under identical data conditions: $N{=}4096$, $D{=}36$, $G{=}6$.

\begin{figure}[h]
    \centering
    \includegraphics[width=\linewidth]{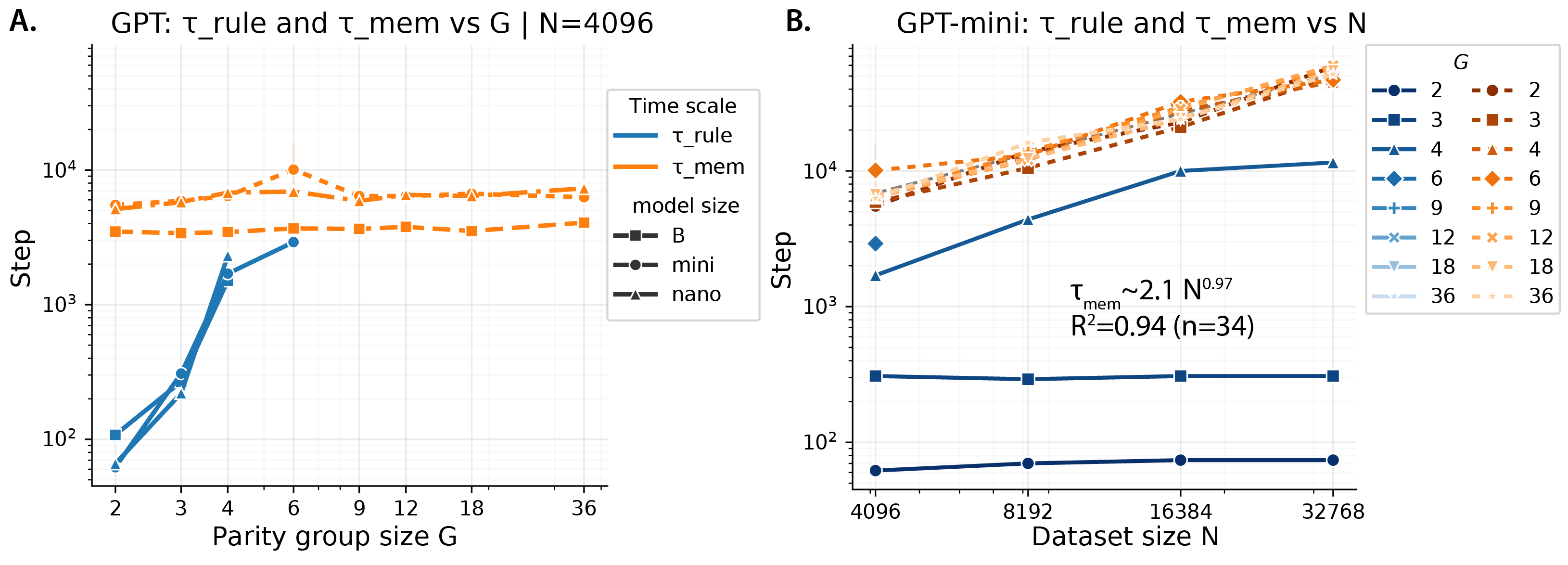}
    \caption{\textbf{Two clocks $\trule$ and $\tmem$ scaling relationships for GPT models — analogous layout to Fig.~\ref{fig:two_clocks_and_scaling}.}
    \textbf{A.}~$\trule$ (blue) and $\tmem$ (orange dashed) vs.~$G$ at $N{=}4096$ for GPT-nano, mini, and B; $\trule$ shown only for learnable $G$ where $\trule$ is smaller than $\tmem$. 
    \textbf{B.}~$(\textit{i})$~$\trule$ vs.~$N$ for learnable $G$. $(\textit{ii})$~$\tmem$ vs.~$N$ across all $G$, with power-law fit $\tmem \approx 2.1\,N^{0.97}$ ($R^2{=}0.94$, $n{=}34$, GPT-mini).
    }
    \label{suppfig:gpt_tau_rule_mem_scaling}
\end{figure}

Figure~\ref{suppfig:gpt_tau_rule_mem_scaling} shows the GPT analogue of Figure~\ref{fig:two_clocks_and_scaling}, confirming that the two-clock phenomenology holds for autoregressive models.

\clearpage
\subsection{Beyond Parity Rules}
\label{app:beyond_parity_rules}

\begin{figure}[!htp]
    \centering
    \includegraphics[width=\linewidth]{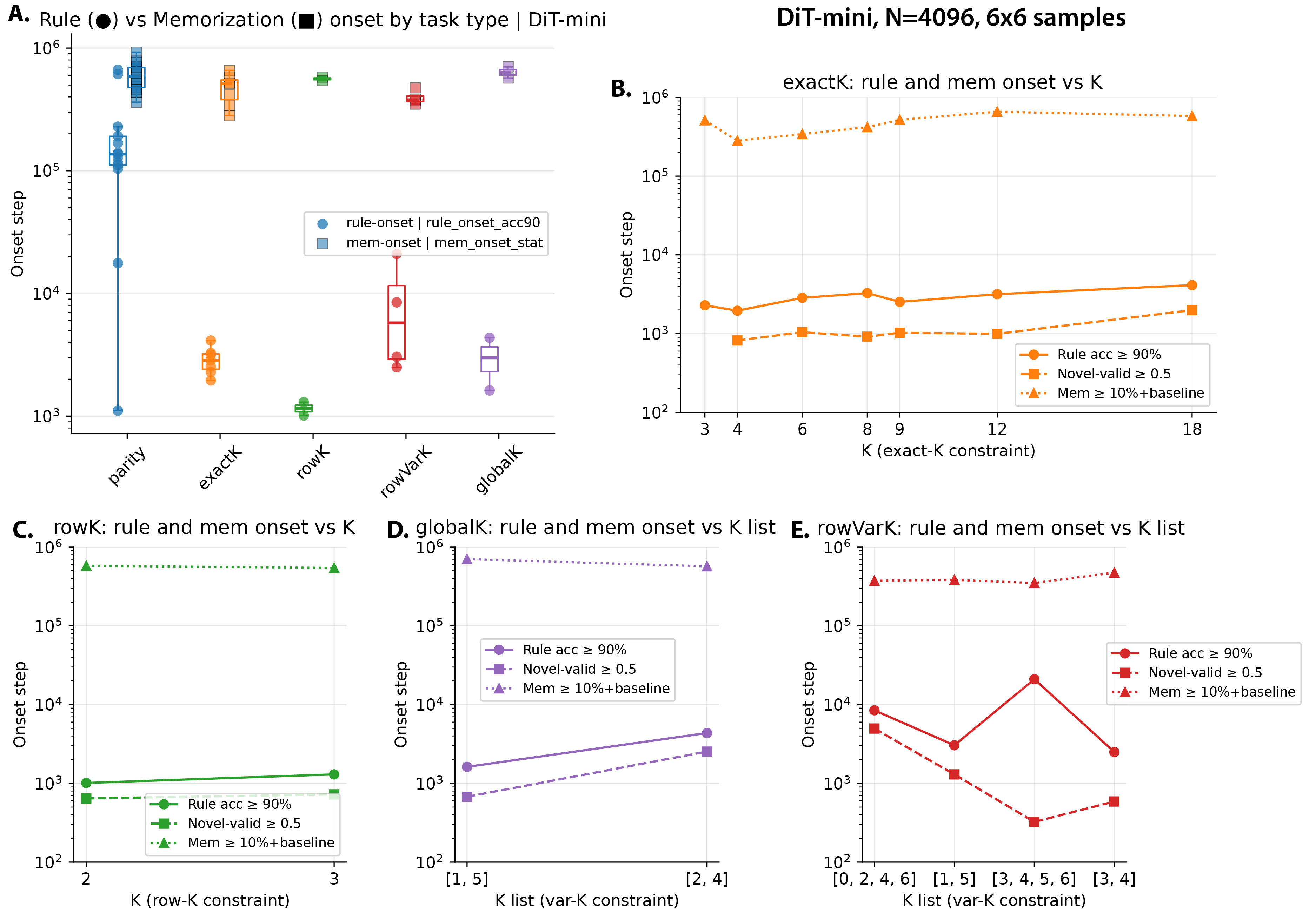}
    \caption{\textbf{Rule-learning and memorization timescales beyond parity.}
    DiT-mini models are trained on $6{\times}6$ binary samples with $N{=}4096$.
    \textbf{A.} Rule onset (circles; sample rule accuracy $\geq 90\%$) and memorization onset (squares; memorization ratio above baseline) across parity, exact-$K$, row-$K$, row-variable-$K$, and global-$K$ tasks.
    \textbf{B--E.} Onset times across different $K$ values or $K$ lists for exact-$K$, row-$K$, global-$K$, and row-variable-$K$ tasks, with novel-valid onset shown when available.
    Counting-style constraints are learned in $10^3$--$10^4$ steps, substantially earlier than their memorization onsets, while memorization remains concentrated near late training across task variants.}
    \label{suppfig:beyond_parity_tau_mem_rule_scaling}
\end{figure}

\paragraph{$\trule$ and $\tmem$ beyond parity rules: exact-$K$ and row-$K$}
For these alternative rules, we found similar two clock structure in their learning dynamics, a time point where rule gets learned and a time point where memorization starts. 
In general, the count based rules are much faster to learn than the parity rules (Fig.~\ref{suppfig:beyond_parity_tau_mem_rule_scaling}\textbf{A}). 

For simple count based rules, like exactK and rowK, the rule learning time $\trule$ is similar to the simplest parity rule $G=2$, saturating in $10^3$ steps. 
Increasing the number of requird active bits $K$ in exact-$K$ rule doesn't quite modulate the $\trule$ (Fig.~\ref{suppfig:beyond_parity_tau_mem_rule_scaling}\textbf{B}). 
This is stark contrast in changing $G$ for parity rules, which significantly delays the rule learning. 
Notably row-$K$, which is group level $K$ bits active rule, doesn't delay the learning, but make $\trule$ even smaller, potentially because fewer bits interact (Fig.~\ref{suppfig:beyond_parity_tau_mem_rule_scaling}\textbf{C}). 
These results support the energy-based interpretation above: exact-$K$ and row-$K$ constraints can be represented by degree-2 penalty on sum of weights, so their rule-learning clocks are much earlier than high-degree group parity. 
Rules with more complex $K$ structure (i.e. allowing $K$ to be chosen from a list) slightly delays $\trule$, but the general structure are consitent with 

Consistent with our main result, the memorization time scale for alternative rules is relatively insensitive to the specific rule and remains largely set by dataset size and model capacity (Fig.~\ref{suppfig:beyond_parity_tau_mem_rule_scaling}\textbf{A}). 
The resulting gap between early rule acquisition and late training-set fit produces an innovation window across these non-parity binary tasks.

\paragraph{Beyond parity rules and binary samples: SuDoKu and Latin Square}

Generalizing beyond rules applying to binary valued samples (boolean cube), we also examined more naturalistic rules such as Sudoku and Latin Square. 
These structured categorical tasks reveal a hierarchy of rule-learning difficulty that is qualitatively different from group parity.
In the parity task, increasing $G$ makes the local rule itself higher-order and eventually prevents rule acquisition, causing late performance to become memorization-driven.
Here, by contrast, the primitive constraints are simpler categorical uniqueness rules, but the difficulty arises from composing them across spatial axes and blocks.

RowOnly is learned rapidly, as it only requires satisfying row-wise constraints. The $\trule$ is comparable to easier parity rules. 
Latin Square and Sudoku are delayed because the model must coordinate row and column constraints, and the column-valid ratio rises much later than the row-valid ratio.
This suggests an anisotropic learning order: DiT first captures the more local or easier row structure before learning cross-row column consistency.
Sudoku, which adds block constraints, reaches high full-sample validity in this setting, but also shows a stronger late increase in sample memorization, especially in the scalar setting.

Thus, unlike high-$G$ parity where failure reflects difficulty of learning the rule, structured categorical tasks expose a different bottleneck: individually learnable rules are nested to constrain the globally consistent samples. 

One notable finding is that, comparing between encoding schemes, when one-hot encoding is used for these rules, the memorization ratio is much smaller, or i.e. memorization is delayed beyond this time horizon. This suggests the $\tmem$ depends also on the encoding of the input, i.e. the geometry of the samples in the space. This is consistent with the view that the basin needs to expand enough to devour nearby valid samples to cause increase of memorized samples. One hot encoding have a higher dimensional raw input space and likely larger distance between valid samples, thus memorization onset is later. 

\begin{figure}[!htp]
    \centering
    \includegraphics[width=\linewidth]{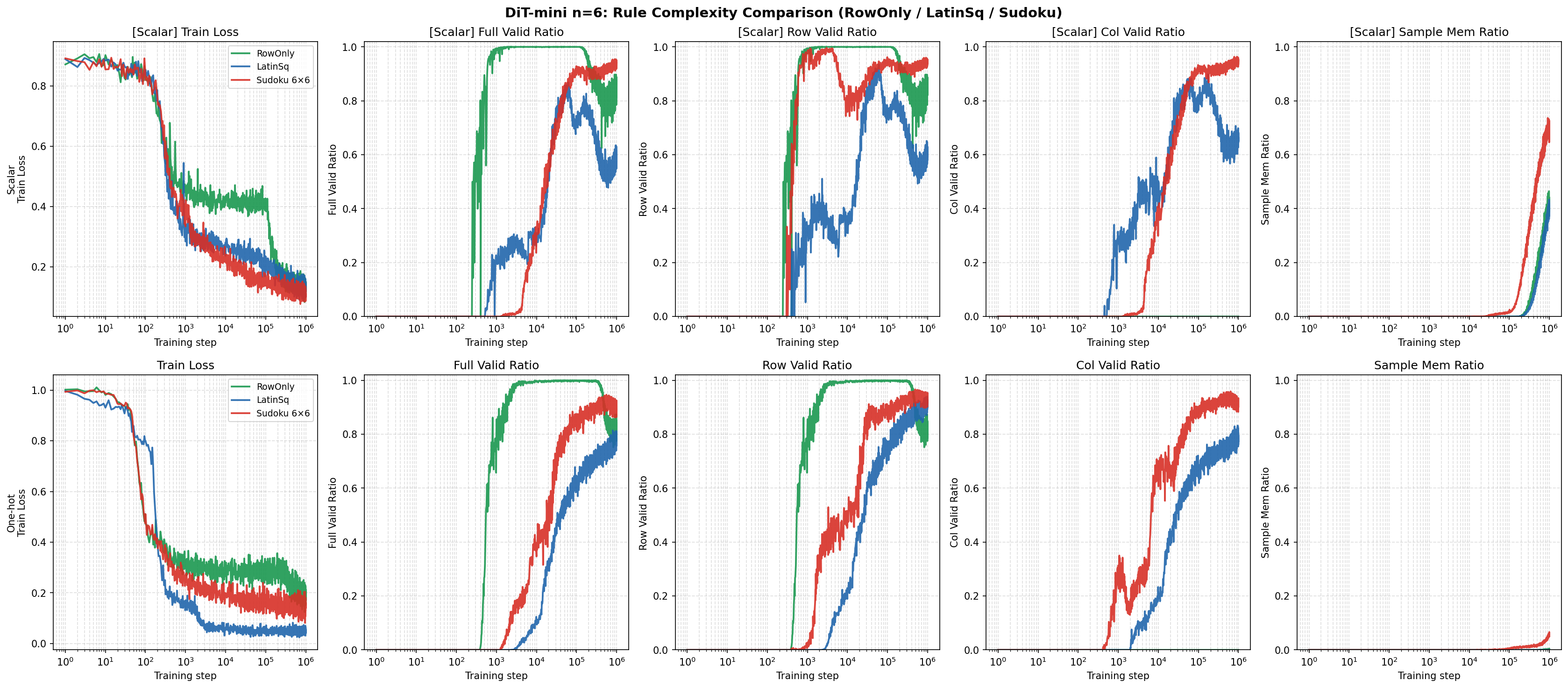}
    \caption{
    \textbf{Learning dynamics for structured categorical rule families: row only (Permutation), LatinSquare, Sudoku.}
    Training trajectories of DiT-mini on three $n=6$ structured categorical datasets: RowOnly (row permutation), Latin Square, and $6{\times}6$ Sudoku with $2{\times}3$ blocks. 
    Columns show train loss, full-sample valid ratio, row-valid ratio, column-valid ratio, and sample memorization ratio over training.
    The top row shows training results with scalar encoding (input shape $1\times 6\times6$), while the bottom row shows training results with one-hot encoding (input shape $6\times 6\times6$).
    RowOnly requires only row-wise constraints; Latin Square requires both row and column constraints; Sudoku adds block constraints on top of row and column constraints.
    Across tasks, RowOnly is learned earliest, reaching high row and full-sample validity around $10^3$ steps, and the rule learning transition is sharp, similar to parity learning. 
    However, Latin Square and Sudoku exhibit delayed and more gradual validity increase, with column-level validity emerging later than row-level validity. 
    Sudoku reaches high full-sample validity more reliably than Latin Square in this run, but also shows the strongest late memorization. 
    Notably, one-hot training generally shows much lower memorization over the same horizon.
    }
    \label{suppfig:beyond_parity_latinsq_sudoku}
\end{figure}

\begin{figure}[t]
    \centering
    \includegraphics[width=\textwidth]{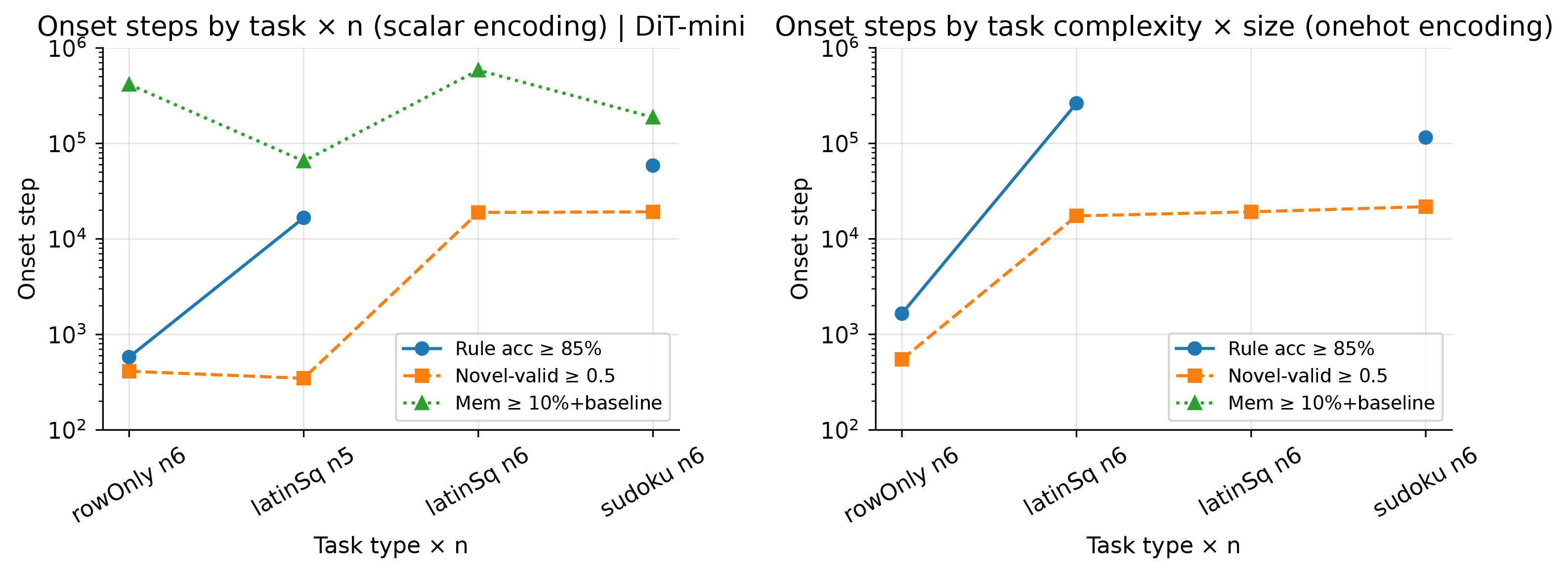}
    \caption{
    \textbf{Onset times for rule learning, novel valid generation, and memorization in structured categorical tasks.}
    We compare DiT-mini training dynamics across RowOnly, Latin Square, and Sudoku rule families.
    \textbf{Left:} scalar encoding.
    \textbf{Right:} one-hot encoding.
    For each task, we measure the first training step at which the model reaches rule accuracy $\geq 85\%$ (blue circles), novel-valid fraction $\geq 0.5$ (orange squares), and memorization ratio at least $10\%$ above the corresponding baseline (green triangles). Note we used a slightly different threshold for $\trule$ since some task didn't reach 90\% accuracy.
    Onset times are plotted on a log scale.
    Across tasks, rule learning and novel-valid generation occur earliest for RowOnly rule and are delayed for more nested constrained rules such as Latin Square and Sudoku.
    Memorization emerges substantially later than rule learning in the scalar setting, and $\tmem$ is comparable across rule complexity, but is earlier for $n=5$ case where the board is smaller. 
    In the one-hot setting memorization does not happen or reach the threshold within the plotted training horizon for most tasks. 
    These results extend the two-clock picture beyond parity: structured categorical rules also exhibit separable timescales for rule acquisition, novel valid generation, and memorization, with the absolute onsets depending on task structure and encoding. 
    }
    \label{suppfig:beyond_parity_latinsq_sudoku_tau_rule_mem}
\end{figure}
\clearpage
\subsection{Optimization ablations}
\label{app:optim_ablation}

\begin{figure}[!htp]
    \centering
    \vspace{-10pt}
    \includegraphics[width=0.80\linewidth]{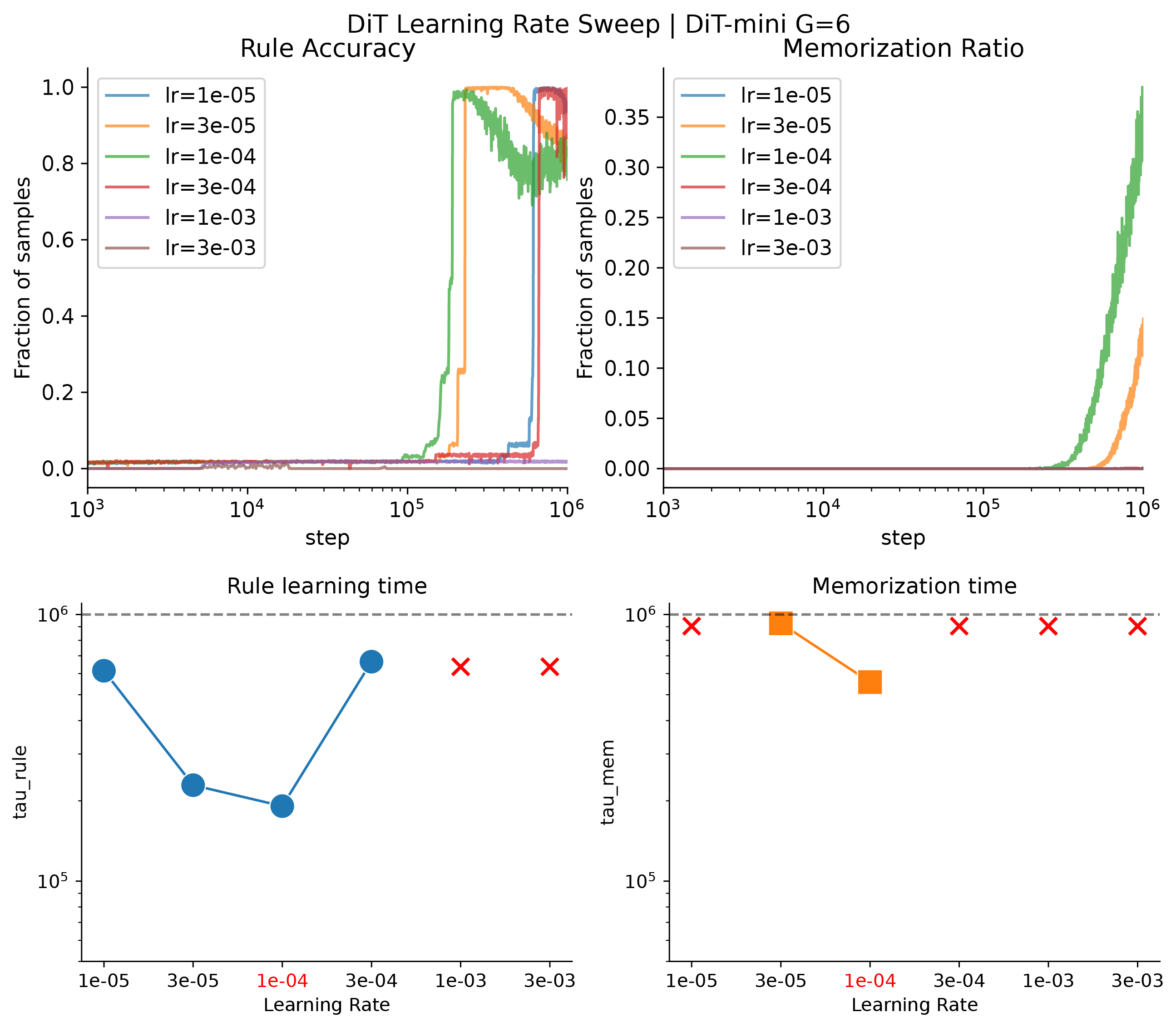}
    \vspace{-10pt}
    \caption{
    \textbf{Learning-rate sweep for DiT-mini on group parity with $G=6$.}
    We train DiT-mini models with learning rates ranging from $10^{-5}$ to $3{\times}10^{-3}$ and track rule accuracy, memorization ratio, rule-learning time $\trule$, and memorization time $\tmem$.
    Moderate learning rates enable rule learning, with $\mathrm{lr}=10^{-4}$ producing the earliest reliable rule-learning transition, followed by $\mathrm{lr}=3{\times}10^{-5}$ and $10^{-5}$.
    Larger learning rates fail to acquire the rule within the training budget, while $\mathrm{lr}=3{\times}10^{-4}$ reaches high rule accuracy only very late.
    Memorization is observed primarily for the learning rates that successfully learn the rule, with $\mathrm{lr}=10^{-4}$ showing the strongest late memorization.
    Red crosses denote runs for which the corresponding onset time was not reached within the training horizon; gray dashed lines mark the training budget.
    }
    \label{suppfig:dit_lr_sweep}
\end{figure}
\begin{figure}[!htp]
    \centering
    \vspace{-10pt}
    \includegraphics[width=0.80\linewidth]{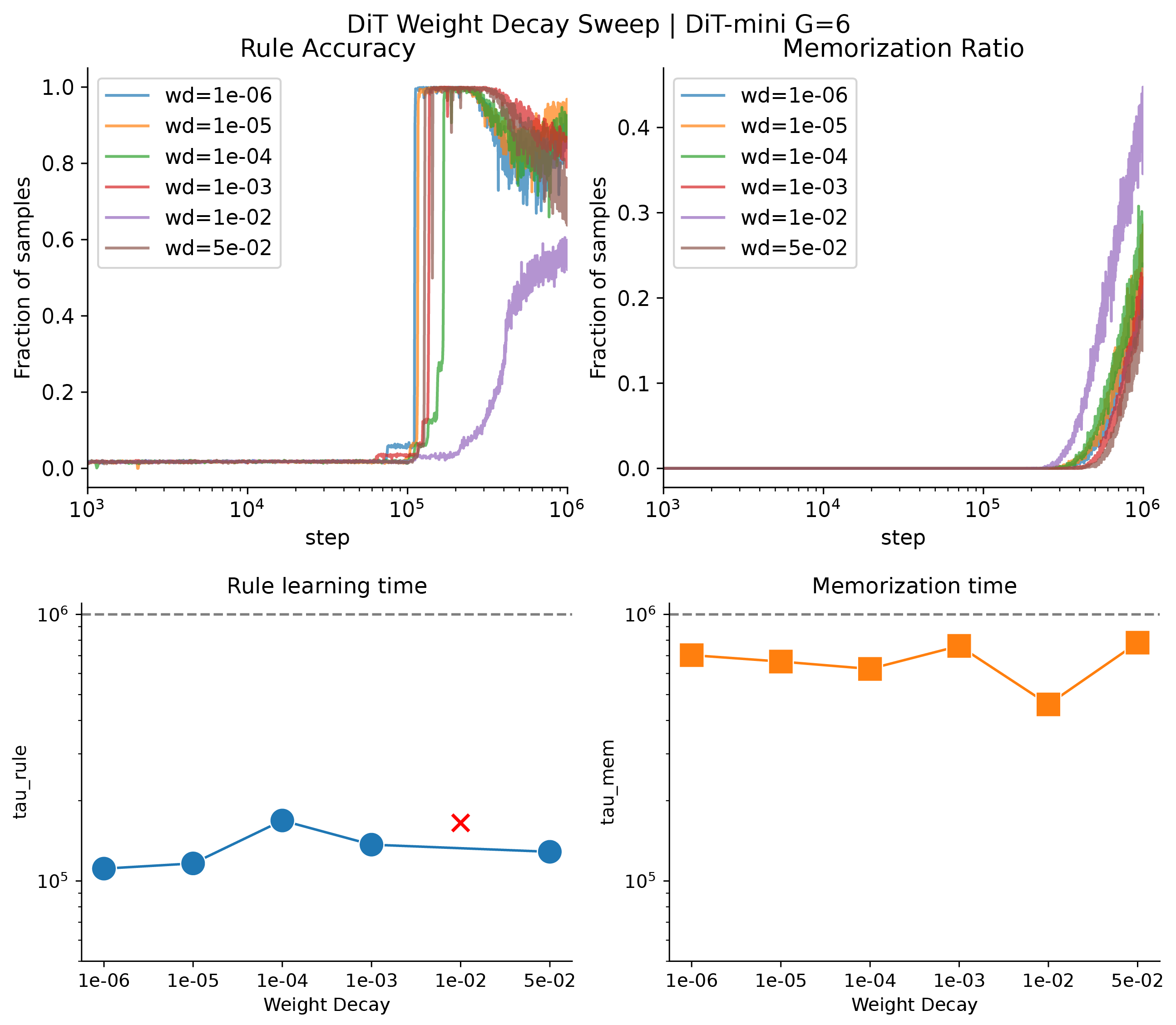}
    \vspace{-10pt}
    \caption{
    \textbf{Weight-decay sweep for DiT-mini on group parity with $G=6$.}
    We train DiT-mini models across a range of weight decay values while keeping the learning rate fixed.
    Rule learning occurs robustly across small to moderate weight decay values, with $\trule$ remaining on the order of $10^5$ steps.
    Very large weight decay delays and weakens rule acquisition, as seen for $\mathrm{wd}=10^{-2}$, which does not reach the rule-learning criterion within the training budget.
    In contrast, memorization onset $\tmem$ is comparatively insensitive to weight decay over most of the tested range, remaining near late training for successful runs.
    These results indicate that DiT rule learning at $G=6$ is more sensitive to learning rate than to moderate weight decay.
    Red crosses denote missing onset times; gray dashed lines mark the training budget.
    }
    \label{suppfig:dit_wd_sweep}
\end{figure}
\begin{figure}[!htp]
    \centering
    \vspace{-10pt}
    \includegraphics[width=0.80\linewidth]{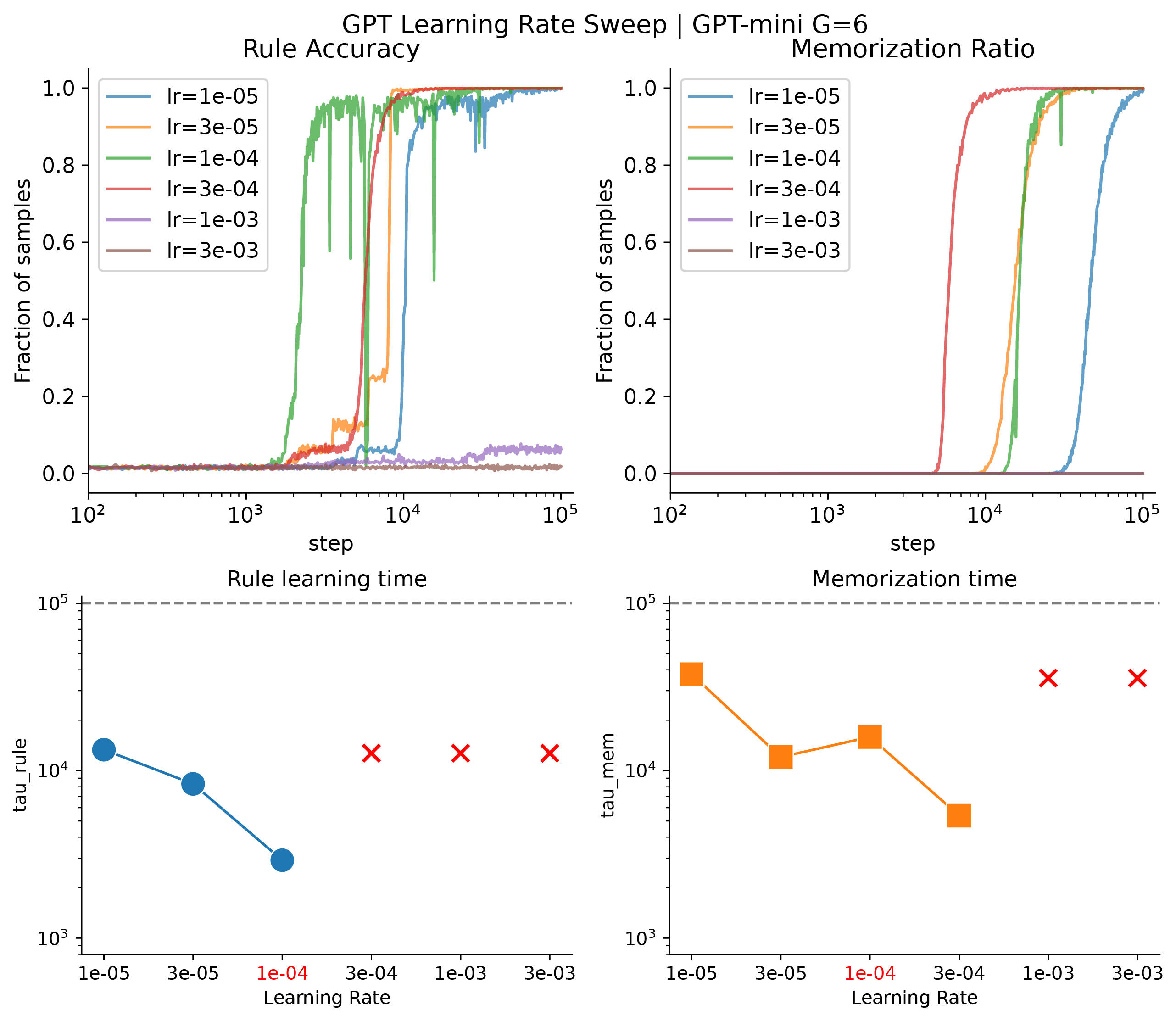}
    \vspace{-10pt}
    \caption{
    \textbf{Learning-rate sweep for GPT-mini on group parity with $G=6$.}
    We train GPT-mini models with learning rates from $10^{-5}$ to $3{\times}10^{-3}$ and measure rule accuracy, memorization ratio, $\trule$, and $\tmem$.
    Rule learning is strongly learning-rate dependent: intermediate learning rates, especially $\mathrm{lr}=10^{-4}$ and $3{\times}10^{-4}$, reach high rule accuracy rapidly, whereas larger learning rates fail to learn the rule.
    Memorization follows successful rule learning and occurs substantially earlier in GPT than in DiT, reaching high memorization ratios for the learning rates that learn the task.
    This sweep shows that GPT has a narrower stable learning-rate window, but within that window both rule learning and memorization proceed rapidly.
    Red crosses denote runs that did not reach the corresponding onset criterion within the training budget; gray dashed lines mark the training budget.
    }
    \label{suppfig:gpt_lr_sweep}
\end{figure}
\begin{figure}[!htp]
    \centering
    \vspace{-10pt}
    \includegraphics[width=0.80\linewidth]{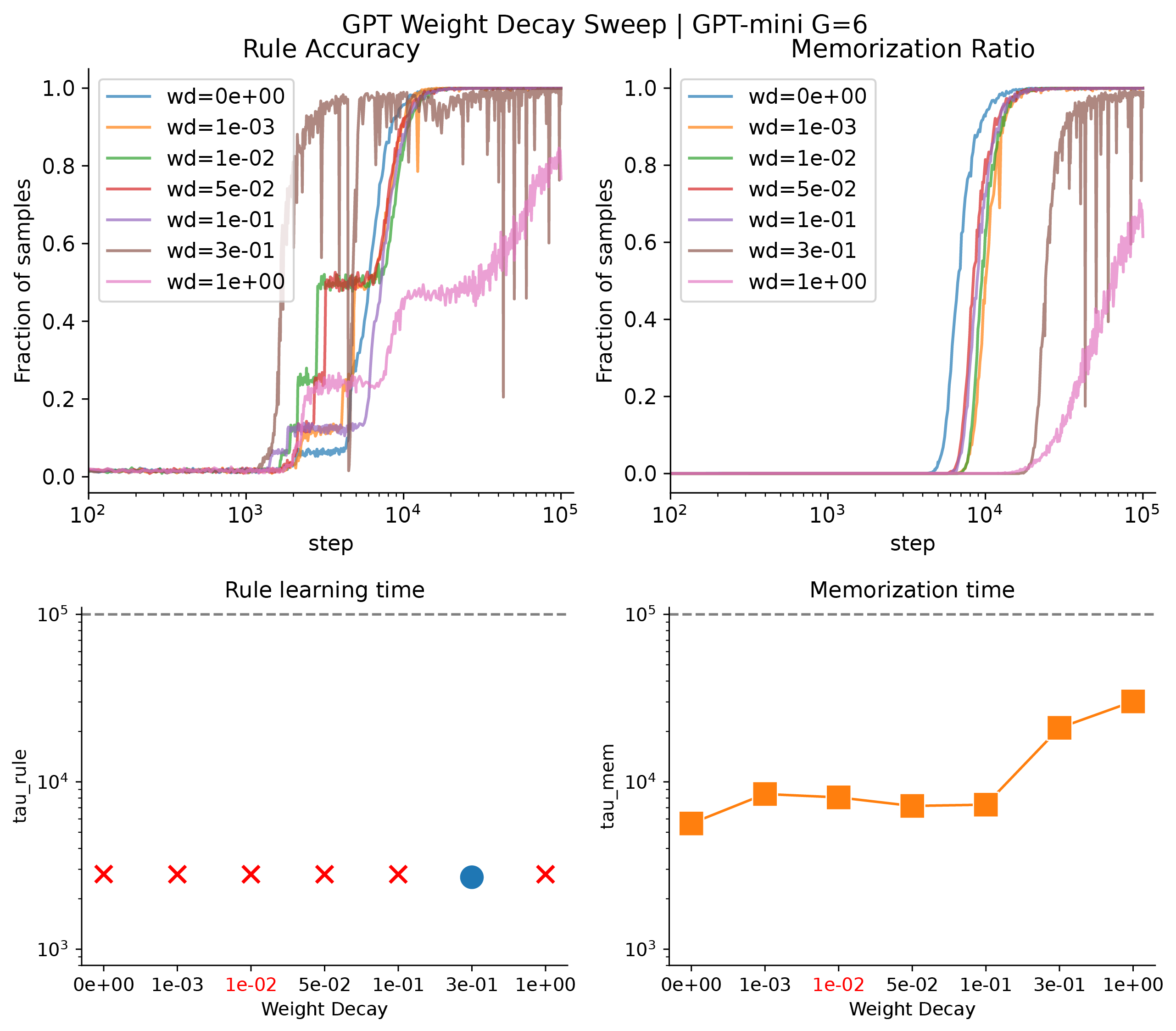}
    \vspace{-10pt}
    \caption{
    \textbf{Weight-decay sweep for GPT-mini on group parity with $G=6$.}
    We train GPT-mini models across weight decay values from $0$ to $1$ and track rule accuracy, memorization ratio, $\trule$, and $\tmem$.
    Across most weight decay values, GPT-mini rapidly reaches high rule accuracy, indicating that rule learning is robust to moderate regularization.
    However, memorization time is strongly affected by large weight decay: increasing weight decay delays $\tmem$, with $\mathrm{wd}=0.3$ and $\mathrm{wd}=1$ producing substantially later memorization.
    Strong regularization can therefore slow memorization without necessarily preventing rule acquisition, although very large weight decay also destabilizes or weakens the rule-accuracy trajectories.
    Red crosses denote onset times that were not reached within the measured window; gray dashed lines mark the training budget.
    }
    \label{suppfig:gpt_wd_sweep}
\end{figure}

We systematically ablate the optimization parameters, i.e. learning rate and weight decay for both the DiT and GPT models, keeping all other hyperparameters at their respective baselines.
Onset times for rule learning and memorization are detected using a \emph{sustained threshold} criterion: the first checkpoint at which the metric exceeds the threshold for $5$ consecutive evaluation points, to avoid counting transient spikes.
The thresholds are: Sample Accuracy $> 0.90$ for rule learning, and Sample Memorization Ratio $> 0.10$ for memorization, rule onset time later than memorization time are discarded since these are usually due to memorization. 

\paragraph{DiT learning rate sweep.}
We sweep $\eta \in \{10^{-5}, 3{\times}10^{-5}, 10^{-4}, 3{\times}10^{-4}, 10^{-3}, 3{\times}10^{-3}\}$ using the Adam optimizer (no weight decay), $10^6$ gradient steps, $N{=}4096$, $G{=}6$, $D{=}36$ (Fig.~\ref{suppfig:dit_lr_sweep}). 
The baseline is $\eta{=}10^{-4}$ (Adam, $\lambda{=}0$). 
Very low learning rates ($10^{-5}$) fail to reach the rule-learning transition within $10^6$ steps; very high rates ($10^{-3}$--$3{\times}10^{-3}$) destabilize training.
The intermediate range $\eta \in [3{\times}10^{-5}, 3{\times}10^{-4}]$ reliably produces both transitions. 
Interestingly, the value used in main experiment $10^{-4}$ yields the earliest rule learning onset and the memorization onset, showing our setting is well suited for learning the task. Though tuning up or down the learning rate can indeed delay the memorization onset. 

\paragraph{DiT weight decay sweep.}
We sweep $\lambda \in \{10^{-6}, 10^{-5}, 10^{-4}, 10^{-3}, 10^{-2}, 5{\times}10^{-2}\}$ using AdamW at fixed $\eta{=}10^{-4}$, $10^6$ steps, all other settings identical to the DiT baseline (Fig.~\ref{suppfig:dit_wd_sweep}). 
The DiT baseline uses $\lambda{=}0$ (Adam), so this sweep explores whether $L_2$ regularization slows or suppresses memorization. We didn't find strong effect of weight decay on $\trule$ or $\tmem$ in the range we explored. 

\paragraph{GPT learning rate sweep.}
We sweep $\eta \in \{10^{-5}, 3{\times}10^{-5}, 10^{-4}, 3{\times}10^{-4}, 10^{-3}, 3{\times}10^{-3}\}$ using AdamW with fixed weight decay $\lambda{=}0.01$, batch size $256$, $10^5$ gradient steps, $N{=}4096$, $G{=}6$ (Fig.~\ref{suppfig:gpt_lr_sweep}). 
We found that the baseline setting $10^{-4}$ is also the fastest to reach the rule transition, and the memorization time. 
Increasing learning rate shortens both $\trule$ and $\tmem$ until a certain point $10^{-3}$ where it won't learn or memorize. 

\paragraph{GPT weight decay sweep.}
We sweep $\lambda \in \{0, 10^{-3}, 10^{-2}, 5{\times}10^{-2}, 3{\times}10^{-1}, 1, 10\}$ using AdamW at fixed $\eta{=}10^{-4}$, all other settings identical to the GPT-mini baseline (Fig.~\ref{suppfig:gpt_wd_sweep}).
We find generally increasing the weight decay delays the memorization time $\tmem$ although rule learning is also affected by it. 
We find that a moderate weight decay of $\lambda{=}3{\times}10^{-1}$ maximizes the temporal gap between rule-learning onset and memorization onset, consistent with the grokking literature~\citep{power2022grokking}, where weight decay suppresses memorization more than it delays rule generalization.
At very large $\lambda$ ($\lambda \geq 1$) the effective learning signal is attenuated and both transitions slow, reducing the observable separation.

\clearpage
\subsection{Stability across repeated runs}\label{app:run_stability}

\begin{figure}[!htp]
\centering
\includegraphics[width=0.88\linewidth]{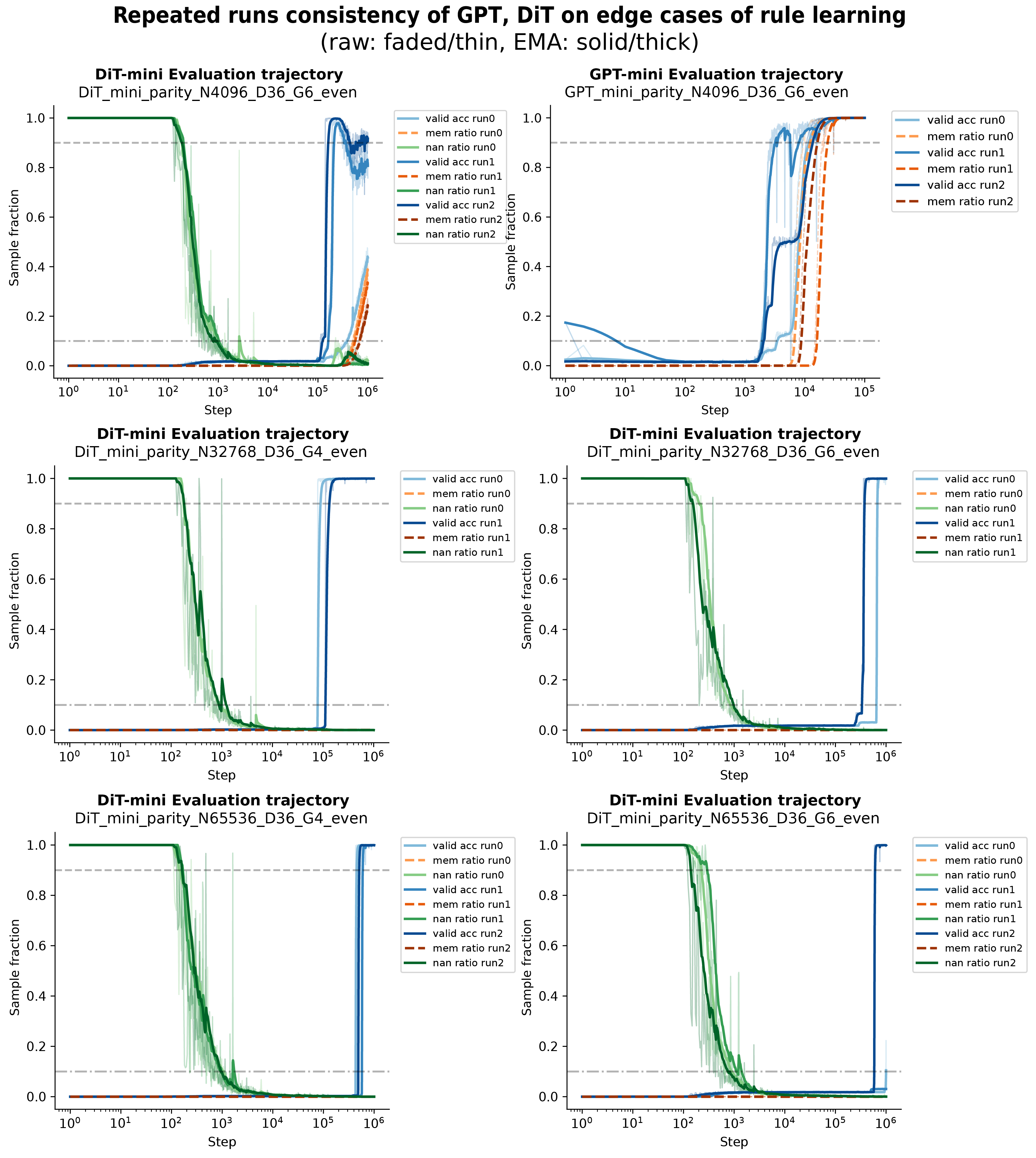}
\caption{\textbf{Cross-run consistency of GPT and DiT near the parity learnability frontier.}
Evaluation trajectories across independent runs for matched GPT-mini and DiT-mini configurations, varying dataset size $N$ and group size $G$.
Different shades of the same hue denote different seeds; thin faded curves show raw trajectories and thick curves show EMA-smoothed trajectories.
Blue denotes rule accuracy, orange dashed denotes memorization ratio, and green denotes the NaN/invalid ratio.
Horizontal dash-dot and dashed lines mark the $0.1$ and $0.9$ thresholds used to define transition onsets.
\textbf{Top row:} $N{=}4096$, $G{=}6$, a data-limited setting near the learnability frontier.
Both architectures show substantial run-to-run variability in the timing of rule learning.
For DiT, two of three runs undergo rule learning around $10^5$ steps, whereas one run fails to reach the $0.9$ threshold and instead shows a concurrent rise in memorization.
For GPT, rule-learning onset varies over roughly $2{\times}10^3$--$2{\times}10^4$ steps, with one run exhibiting a transient plateau near $0.5$ accuracy before completing the transition.
\textbf{Middle and bottom rows:} larger datasets ($N{=}32768$ and $N{=}65536$) at $G{=}4$ and $G{=}6$.
For $G{=}4$, rule-learning transitions become sharp and tightly clustered across seeds, and the earlier NaN-ratio decay phase near $10^2$--$10^3$ steps is highly reproducible.
For the harder $G{=}6$ case, variability remains: even at $N{=}65536$, only one of three DiT runs reaches successful rule learning.
Together, these results indicate that cross-run instability is concentrated in the timing and success of rule acquisition near the $G{\approx}6$ learnability frontier, whereas the earlier Boolean-cube quantization phase and the later memorization dynamics are comparatively stable across seeds.}
\label{fig:cross_run_stability}
\end{figure}

\clearpage

\section{Broader Impacts}
\label{app:broader_impacts}

This work is a foundational study of rule learning and memorization in controlled generative models, and does not introduce deployed systems, human-subject data, or high-risk datasets.
Its positive impact is to provide diagnostics for when generative models learn abstract constraints versus merely fit finite training sets, which may help improve reliability in structured generation, scientific modeling, and reasoning systems.
The main negative risk is indirect: methods that improve rule-conforming generation could eventually strengthen generative systems in domains where misuse is possible, and overinterpreting transient rule-learning behavior could lead to misplaced trust in models that later memorize or violate constraints.
We mitigate these risks by focusing on synthetic tasks, reporting failure modes alongside successes, and emphasizing evaluation over deployment.

\section{LLM Usage}
\label{app:llm_usage}

Generative AI tools, including large language models, were used during manuscript preparation and research workflow support.
Their use included visualizing results for submission, implementing standard methods, editing text for grammar and word choice, data processing and filtering, facilitating or running experiments, and drafting sections of the paper.
All scientific claims, experimental choices, analyses, and final manuscript content were reviewed and controlled by the authors.



\end{document}

%% file: math_commands.tex
\usepackage{amsmath,amsfonts,bm}

\newcommand{\trule}{\tau_{\text{rule}}}
\newcommand{\tmem}{\tau_{\text{mem}}}

\def\eqref#1{equation~\ref{#1}}

\def\1{\bm{1}}

\DeclareMathAlphabet{\mathsfit}{\encodingdefault}{\sfdefault}{m}{sl}
\SetMathAlphabet{\mathsfit}{bold}{\encodingdefault}{\sfdefault}{bx}{n}



%% file: related_work.tex
\section{Related work}
\label{sec:related_work}

\paragraph{Rule learning in Diffusion models} Prior work has demonstrated that diffusion models learn diverse rules with mixed success.
\cite{wang_diverse_2024} showed that unconditional diffusion models can learn to generate according to only some of the rules in RAVEN's progression matrices encoded as integer arrays. In particular, rules corresponding to logical operations (AND, OR, XOR) over sets of attributes were difficult to learn. Similarly, \cite{Han2025CanDiff} examined rule learning in the pixel space, showing that diffusion models can learn the coarse proportional relationship between bars and shadows length
but not the precise rule specified in the training set.
These prior works prompt this study, which examines what kind of binary rules can be learned.

\paragraph{Memorization and Creativity in Diffusion models}
The question of when diffusion models are able to generate genuinely novel samples is relevant both for understanding these tools, and for preventing these models from reproducing potentially sensitive training data.
From the rule learning perspective, the model that truly learns the rule should not simply regurgitate the training set, but learn the data manifold underlying it.
From a score-matching perspective, however, exactly matching the score of the empirical data distribution implies that the reverse process will simply reproduce training samples \citep{kamb2024analytict,li2024goodscoredoeslead,wang_unreasonable_2024}. Yet high-quality diffusion models routinely generate images that are not identical copies of any training example, raising the question of what mechanisms enable this generalization.

A striking empirical demonstration of generalization comes from \citet{Kadkhodaie2024Generalization} and \citet{zhang2023emergence}: when two diffusion models are trained on disjoint splits of the same dataset, both models—given sufficient data—produce highly consistent samples from matched noise seeds, and these samples differ from the nearest training images in either split. This cross-split consistency indicates that the models converge onto a shared generative structure (i.e. data manifold) rather than memorizing their respective training sets.

One line of work attributes this convergence to the fact that well-trained diffusion models behave surprisingly similarly to a linear-score diffusion model, which captures only the first two moments of the data distribution \citep{wang_unreasonable_2024,li2024understanding}. Under this view, cross-split consistency follows naturally from the largely shared Gaussian summary statistics of the two splits, which induce a shared linear score structure \citep{wang2026RMT_Diffusion}.

Beyond linear structure, a complementary line of work emphasizes architectural inductive biases. \citet{kamb2024analytict} show that when the score network is a simple convolutional neural network, its locality and translation equivariance favor patch-wise composition, producing globally novel samples that remain locally consistent "mosaics." \citet{lukoianov2026locality} further demonstrate that the effective receptive field governing this locality is itself determined by image statistics and the linear (Wiener-filter) score. \citet{Wang2025Analytical} generalize this picture, showing that different architectural constraints induce different distributional approximations: linear networks learn the Gaussian approximation, while circular convolutional networks learn the stationary Gaussian process approximation. \citet{finn2025originscreativity} extend the analysis to attention-based architectures, providing evidence that a final self-attention layer promotes global consistency across distant regions, organizing locally plausible features into coherent global layouts.

Complementary theoretical work probes why well-trained diffusion models generalize despite memorization pressures \citep{biroli2024dynamicalregimes,Vastola2025Generalization,2502_19499v1}. Taken together, these results suggest that departures from exact empirical-score fitting—mediated by both architectural and training-dynamical inductive biases—are what allow diffusion models to avoid pure memorization while maintaining visual plausibility \citep{ambrogioni2023searchdispersedmemoriesgenerative}.

In this work, we study memorization and generalization from a dynamical point of view, in a setting where the underlying data distribution is tractable, allowing us to characterize these effects precisely. 

\paragraph{Learning parity}
Many results in this paper focus on parity learning, a versatile testbed that has been widely adopted for understanding both the representational and learning aspects of neural networks~\citep{hahn20,bhattamishra2022simplicity,glasgow24,abbe2024generalization,abbe25init,shoshani2025hardness}.
The difficulty depends on the number of bits that the parity is defined over, where more bits require a higher boolean sensitivity or larger weight norms in the case of neural networks.
For Transformers specifically, learning parity requires growing the MLP norms~\citep{liu2022shortcut,hahn2024sensitive} and the use of normalization layers~\citep{hahn20,yao2021self,chiang2022overcoming}.
Even when a network is sufficiently expressive, parity is computationally challenging to learn~\citep{Kearns98,barak2022hidden,edelman2023pareto,wen2024sparse,kim25}.
In this work, we explore parity learning from a generative modeling perspective, leveraging this well studied problem to characterize how much modern generative modeling framework, in particular diffusion, can learn these underlying structures.

\paragraph{Score complexity and spectral bias.}
As discussed in \Cref{sec:beyond_parity}, 
the parity score naturally decomposes into a local term and a global multiplicative term $\prod_{j=1}^G x_j$, whose polynomial degree grows with $G$. In the Fourier/Walsh-Hadamard basis, higher-degree interactions correspond to higher-frequency components, and it is well established that neural networks exhibit a \emph{spectral bias}, fitting low-frequency components before high-frequency ones \citep{canatar2021,Wang2025Analytical}. This framework offers a natural explanation for our learning dynamics: small-$G$ components emerge early in training, while large-$G$ components appear only much later—if they appear at all. When the latter are not learned from data, accuracy improvements in late training tend to come from memorization rather than genuine rule acquisition. A more formal score-complexity analysis could help predict the point at which models shift from generalizing to overfitting, and explain how architectural constraints shape this transition.

\paragraph{Larger group size complicates learning}
It is well known that higher-degree polynomials (i.e., larger $G$) is more computationally challenging to learn, where the required number of steps grows exponentially in the polynomial degree~\citep{barak2022hidden,abbe2023sgd,damian2025generative}.
We observe similar phenomenon in our experiments (Fig.~\ref{fig:two_clocks_and_scaling}\textbf{C,E}), where rules with higher $G$ tend to require more gradient steps to learn.
For transformer specifically, it has also been observed that higher-degree or more global functions are harder to learn~\citep{bhattamishra2022simplicity,abbe2024far,hahn2024sensitive,vasudeva2024simplicity}.

\paragraph{More depth eases learning}
Although one layer suffices to express the parity function for a fixed $G$~\citep{hahn20,liu2022shortcut}, we observe that a greater depth leads to better learning empirically.
One intuition is that representing the optimal score function of the ground-truth parity distribution requires the network to reach certain Lipschitzness~\citep{hahn20},
and using more layers means that the required per-layer Lipschitzness
\footnote{The Lipschitzness comes from both the weight norms and the scaling introduced by normalization layers~\citep{yao2021self,hahn2024sensitive}.}
is smaller, which can be easier to reach in learning.
This intuition is also consistent with our finding that rule learning often precedes memorization,
since the Lipschitzness of the memorizing score function is usually higher than that of the score function of the ground truth rule distribution; the former requires a larger weight norm, as also noted in ~\cite{montanari2025dynamical}.
Further, it is consistent with our observation that deeper DiTs tend to memorize more (Fig.~\ref{fig:dataset_eval_model_cmp}\textbf{B,C}), as depth facilitates the growth of Lipschitzness.

%% file: iclr2025_conference.bib
@article{li2024understanding,
  title={Understanding generalizability of diffusion models requires rethinking the hidden gaussian structure},
  author={Li, Xiang and Dai, Yixiang and Qu, Qing},
  journal={Advances in neural information processing systems},
  volume={37},
  pages={57499--57538},
  year={2024}
}

@article{lukoianov2026locality,
  title={Locality in image diffusion models emerges from data statistics},
  author={Lukoianov, Artem and Yuan, Chenyang and Solomon, Justin and Sitzmann, Vincent},
  journal={Advances in Neural Information Processing Systems},
  volume={38},
  pages={95121--95157},
  year={2026}
}

@article{wang2026RMT_Diffusion,
  title={A Random Matrix Theory Perspective on the Consistency of Diffusion Models},
  author={Wang, Binxu and Zavatone-Veth, Jacob and Pehlevan, Cengiz},
  journal={arXiv preprint arXiv:2602.02908},
  year={2026}
}

@article{darcet2023ViTregister,
  title={Vision transformers need registers},
  author={Darcet, Timoth{\'e}e and Oquab, Maxime and Mairal, Julien and Bojanowski, Piotr},
  journal={arXiv preprint arXiv:2309.16588},
  year={2023}
}

@article{yu2024repa,
  title={Representation alignment for generation: Training diffusion transformers is easier than you think},
  author={Yu, Sihyun and Kwak, Sangkyung and Jang, Huiwon and Jeong, Jongheon and Huang, Jonathan and Shin, Jinwoo and Xie, Saining},
  journal={arXiv preprint arXiv:2410.06940},
  year={2024}
}

@article{power2022grokking,
  title={Grokking: Generalization beyond overfitting on small algorithmic datasets},
  author={Power, Alethea and Burda, Yuri and Edwards, Harri and Babuschkin, Igor and Misra, Vedant},
  journal={arXiv preprint arXiv:2201.02177},
  year={2022}
}

@article{Kearns98,
    author = {Kearns, Michael},
    title = {Efficient noise-tolerant learning from statistical queries},
    year = {1998},
    issue_date = {Nov. 1998},
    publisher = {Association for Computing Machinery},
    address = {New York, NY, USA},
    volume = {45},
    number = {6},
    issn = {0004-5411},
    url = {https://doi.org/10.1145/293347.293351},
    doi = {10.1145/293347.293351},
    journal = {J. ACM},
    month = nov,
    pages = {983–1006},
    numpages = {24}
}

@inproceedings{kim25,
  author    = {Juno Kim and Taiji Suzuki},
  title     = {Transformers Provably Solve Parity Efficiently with Chain of Thought},
  booktitle = {The Thirteenth International Conference on Learning Representations, {ICLR} 2025, Singapore, April 24-28, 2025},
  publisher = {OpenReview.net},
  year      = {2025},
  url       = {https://openreview.net/forum?id=n2NidsYDop},
  bibsource = {dblp computer science bibliography, https://dblp.org}
}

@article{wen2024sparse,
  title     = {From Sparse Dependence to Sparse Attention: Unveiling How Chain-of-Thought Enhances Transformer Sample Efficiency},
  author    = {Kaiyue Wen and Huaqing Zhang and Hongzhou Lin and Jingzhao Zhang},
  journal   = {International Conference on Learning Representations},
  year      = {2024},
  doi       = {10.48550/arXiv.2410.05459},
  bibSource = {Semantic Scholar https://www.semanticscholar.org/paper/48e6f754c53f7a1c253576f8e18e33bbed0d546f}
}

@article{barak2022hidden,
  title     = {Hidden Progress in Deep Learning: SGD Learns Parities Near the Computational Limit},
  author    = {B. Barak and Benjamin L. Edelman and Surbhi Goel and S. Kakade and Eran Malach and Cyril Zhang},
  journal   = {Neural Information Processing Systems},
  year      = {2022},
  doi       = {10.48550/arXiv.2207.08799},
  bibSource = {Semantic Scholar https://www.semanticscholar.org/paper/f5f5616f39493566a9d502f611adcc8f1ceb394e}
}

@article{edelman2023pareto,
  title   = {Pareto Frontiers in Neural Feature Learning: Data, Compute, Width, and Luck},
  author  = {Benjamin L. Edelman and Surbhi Goel and Sham Kakade and Eran Malach and Cyril Zhang},
  year    = {2023},
  journal = {arXiv preprint arXiv: 2309.03800}
}

@article{hahn20,
  title={Theoretical limitations of self-attention in neural sequence models},
  author={Hahn, Michael},
  journal={Transactions of the Association for Computational Linguistics},
  volume={8},
  pages={156--171},
  year={2020},
  publisher={MIT Press One Rogers Street, Cambridge, MA 02142-1209, USA journals-info~…}
}

@article{hahn2024sensitive,
  title   = {Why are Sensitive Functions Hard for Transformers?},
  author  = {Michael Hahn and Mark Rofin},
  year    = {2024},
  journal = {arXiv preprint arXiv: 2402.09963}
}

@article{liu2022shortcut,
  title={Transformers learn shortcuts to automata},
  author={Liu, Bingbin and Ash, Jordan T and Goel, Surbhi and Krishnamurthy, Akshay and Zhang, Cyril},
  journal={arXiv preprint arXiv:2210.10749},
  year={2022}
}

@article{chiang2022overcoming,
  title   = {Overcoming a Theoretical Limitation of Self-Attention},
  author  = {David Chiang and Peter Cholak},
  year    = {2022},
  journal = {arXiv preprint arXiv: 2202.12172}
}

@article{yao2021self,
  title={Self-attention networks can process bounded hierarchical languages},
  author={Yao, Shunyu and Peng, Binghui and Papadimitriou, Christos and Narasimhan, Karthik},
  journal={arXiv preprint arXiv:2105.11115},
  year={2021}
}

@inproceedings{glasgow24,
  author    = {Margalit Glasgow},
  title     = {{SGD} Finds then Tunes Features in Two-Layer Neural Networks with near-Optimal Sample Complexity: {A} Case Study in the {XOR} problem},
  booktitle = {The Twelfth International Conference on Learning Representations, {ICLR} 2024, Vienna, Austria, May 7-11, 2024},
  publisher = {OpenReview.net},
  year      = {2024},
  url       = {https://openreview.net/forum?id=HgOJlxzB16},
  bibsource = {dblp computer science bibliography, https://dblp.org}
}

@article{shoshani2025hardness,
  title   = {Hardness of Learning Fixed Parities with Neural Networks},
  author  = {Itamar Shoshani and Ohad Shamir},
  year    = {2025},
  journal = {arXiv preprint arXiv: 2501.00817}
}

@article{abbe2024far,
  title     = {How Far Can Transformers Reason? The Globality Barrier and Inductive Scratchpad},
  author    = {Emmanuel Abbe and Samy Bengio and Aryo Lotfi and Colin Sandon and Omid Saremi},
  journal   = {Neural Information Processing Systems},
  year      = {2024},
  bibSource = {Semantic Scholar https://www.semanticscholar.org/paper/c2f048c4523877dd55227aedbd19b6e095703e2d}
}

@article{mou2023t2i0adapter0,
  title     = {T2I-Adapter: Learning Adapters to Dig out More Controllable Ability for Text-to-Image Diffusion Models},
  author    = {Chong Mou and Xintao Wang and Liangbin Xie and Jing Zhang and Zhongang Qi and Ying Shan and Xiaohu Qie},
  journal   = {AAAI Conference on Artificial Intelligence},
  year      = {2023},
  doi       = {10.48550/arXiv.2302.08453},
  bibSource = {Semantic Scholar https://www.semanticscholar.org/paper/58842cdca3ea68f7b9e638b288fc247a6f26dafc}
}

@article{chen25multilayer,
  journal = {2025 IEEE 66th Annual Symposium on Foundations of Computer Science (FOCS)},
  pages   = {2631-2653},
  doi     = {10.1109/FOCS63196.2025.00136},
  title   = {Theoretical limitations of multi-layer Transformer},
  year    = {2025},
  author  = {Lijie Chen and Binghui Peng and Hongxun Wu}
}

@article{montanari2025dynamical,
  title   = {Dynamical Decoupling of Generalization and Overfitting in Large Two-Layer Networks},
  author  = {Andrea Montanari and Pierfrancesco Urbani},
  year    = {2025},
  journal = {arXiv preprint arXiv: 2502.21269}
}

@article{vasudeva2024simplicity,
  title   = {Simplicity Bias of Transformers to Learn Low Sensitivity Functions},
  author  = {Bhavya Vasudeva and Deqing Fu and Tianyi Zhou and Elliott Kau and Youqi Huang and Vatsal Sharan},
  year    = {2024},
  journal = {arXiv preprint arXiv: 2403.06925}
}

@article{abbe2023sgd,
  title     = {SGD learning on neural networks: leap complexity and saddle-to-saddle dynamics},
  author    = {E. Abbe and Enric Boix-Adserà and Theodor Misiakiewicz},
  journal   = {Annual Conference Computational Learning Theory},
  year      = {2023},
  doi       = {10.48550/arXiv.2302.11055},
  bibSource = {Semantic Scholar https://www.semanticscholar.org/paper/59e3b8ae1e119e2f48c8e64ecb32229e45ffcc01}
}

@article{damian2025generative,
  title   = {The Generative Leap: Sharp Sample Complexity for Efficiently Learning Gaussian Multi-Index Models},
  author  = {Alex Damian and Jason D. Lee and Joan Bruna},
  year    = {2025},
  journal = {arXiv preprint arXiv: 2506.05500}
}

@article{bhattamishra2022simplicity,
  title     = {Simplicity Bias in Transformers and their Ability to Learn Sparse Boolean Functions},
  author    = {S. Bhattamishra and Arkil Patel and Varun Kanade and Phil Blunsom},
  journal   = {Annual Meeting of the Association for Computational Linguistics},
  year      = {2022},
  doi       = {10.48550/arXiv.2211.12316},
  bibSource = {Semantic Scholar https://www.semanticscholar.org/paper/fb6d75a4f3b1af2058f59957116c178a47b56f05}
}

@misc{ambrogioni2023searchdispersedmemoriesgenerative,
      title={In search of dispersed memories: Generative diffusion models are associative memory networks}, 
      author={Luca Ambrogioni},
      year={2023},
      eprint={2309.17290},
      archivePrefix={arXiv},
      primaryClass={stat.ML},
      url={https://arxiv.org/abs/2309.17290}, 
}

@article{abbe2024generalization,
  title   = {Generalization on the unseen, logic reasoning and degree curriculum},
  author  = {Abbe, Emmanuel and Bengio, Samy and Lotfi, Aryo and Rizk, Kevin},
  journal = {Journal of Machine Learning Research},
  volume  = {25},
  number  = {331},
  pages   = {1-58},
  year    = {2024}
}

@inproceedings{abbe25init,
  author    = {Emmanuel Abbe and Elisabetta Cornacchia and Jan Hazla and Donald Kougang{-}Yombi},
  title     = {Learning High-Degree Parities: The Crucial Role of the Initialization},
  booktitle = {The Thirteenth International Conference on Learning Representations, {ICLR} 2025, Singapore, April 24-28, 2025},
  publisher = {OpenReview.net},
  year      = {2025},
  url       = {https://openreview.net/forum?id=OuNIWgGGif},
  bibsource = {dblp computer science bibliography, https://dblp.org}
}

@misc{li2024goodscoredoeslead,
      title={A Good Score Does not Lead to A Good Generative Model}, 
      author={Sixu Li and Shi Chen and Qin Li},
      year={2024},
      eprint={2401.04856},
      archivePrefix={arXiv},
      primaryClass={cs.LG},
      url={https://arxiv.org/abs/2401.04856}, 
}

@misc{bonnaire2025memorize,
      title={Why Diffusion Models Don't Memorize: The Role of Implicit Dynamical Regularization in Training}, 
      author={Tony Bonnaire and Raphaël Urfin and Giulio Biroli and Marc Mézard},
      year={2025},
      eprint={2505.17638},
      archivePrefix={arXiv},
      primaryClass={cs.LG},
      url={https://arxiv.org/abs/2505.17638}, 
}

@misc{biroli2024dynamicalregimes,
  title        = {Dynamical Regimes of Diffusion Models},
  author       = {Biroli, Giulio and Bonnaire, Tony and De Bortoli, Valentin and M\'ezard, Marc},
  year         = {2024},
  eprint       = {2402.18491},
  archivePrefix= {arXiv},
  primaryClass = {cs.LG},
  url          = {https://arxiv.org/abs/2402.18491},
}

@article{dodson2026two,
  title={Two Calm Ends and the Wild Middle: A Geometric Picture of Memorization in Diffusion Models},
  author={Dodson, Nick and Gao, Xinyu and Wang, Qingsong and Wang, Yusu and Wan, Zhengchao},
  journal={arXiv preprint arXiv:2602.17846},
  year={2026}
}

@misc{finn2025originscreativity,
      title={Origins of Creativity in Attention-Based Diffusion Models}, 
      author={Emma Finn and T. Anderson Keller and Manos Theodosis and Demba E. Ba},
      year={2025},
      eprint={2506.17324},
      archivePrefix={arXiv},
      primaryClass={cs.LG},
      url={https://arxiv.org/abs/2506.17324}, 
}

@misc{kamb2024analytict,
      title={An analytic theory of creativity in convolutional diffusion models}, 
      author={Mason Kamb and Surya Ganguli},
      year={2024},
      eprint={2412.20292},
      archivePrefix={arXiv},
      primaryClass={cs.LG},
      url={https://arxiv.org/abs/2412.20292}, 
}

@misc{Peebles2023Scalable,
  title = {Scalable {{Diffusion Models}} with {{Transformers}}},
  author = {Peebles, William and Xie, Saining},
  year = {2023},
  month = mar,
  number = {arXiv:2212.09748},
  eprint = {2212.09748},
  primaryclass = {cs},
  publisher = {arXiv},
  doi = {10.48550/arXiv.2212.09748},
  urldate = {2025-07-22},
  archiveprefix = {arXiv}
}

@article{zhang2023emergence,
  title = {The Emergence of Reproducibility and Consistency in Diffusion Models},
  author = {Zhang, Huijie and Zhou, Jinfan and Lu, Yifu and Guo, Minzhe and Shen, Liyue and Qu, Qing},
  year = {2023},
  journal = {arXiv preprint arXiv:2310.05264},
  eprint = {2310.05264},
  archiveprefix = {arXiv}
}

@inproceedings{prabhudesai2025diffusion,
  title     = {Diffusion Beats Autoregressive in Data-Constrained Settings},
  author    = {Mihir Prabhudesai and Mengning Wu and Amir Zadeh and Katerina Fragkiadaki and Deepak Pathak},
  booktitle = {The Thirty-ninth Annual Conference on Neural Information Processing Systems},
  year      = {2025},
  url       = {https://openreview.net/forum?id=W5Ht05jF4c}
}

@article{ni2025diffusion,
  title   = {Diffusion Language Models are Super Data Learners},
  author  = {Jinjie Ni and Qian Liu and Longxu Dou and Chao Du and Zili Wang and Hang Yan and Tianyu Pang and Michael Qizhe Shieh},
  year    = {2025},
  journal = {arXiv preprint arXiv: 2511.03276}
}

@article{karras2022elucidatingDesignSp,
    title = {Elucidating the design space of diffusion-based generative models},
    journal = {arXiv preprint arXiv:2206.00364},
    author = {Karras, Tero and Aittala, Miika and Aila, Timo and Laine, Samuli},
    year = {2022},
}

@misc{peebles_scalable_2023,
    title = {Scalable {Diffusion} {Models} with {Transformers}},
    url = {http://arxiv.org/abs/2212.09748},
    doi = {10.48550/arXiv.2212.09748},
    urldate = {2025-07-22},
    publisher = {arXiv},
    author = {Peebles, William and Xie, Saining},
    month = mar,
    year = {2023},
    note = {arXiv:2212.09748 [cs]},
}

@misc{wang_diverse_2024,
    title = {Diverse capability and scaling of diffusion and auto-regressive models when learning abstract rules},
    url = {http://arxiv.org/abs/2411.07873},
    doi = {10.48550/arXiv.2411.07873},
    urldate = {2024-12-02},
    publisher = {arXiv},
    author = {Wang, Binxu and Shang, Jiaqi and Sompolinsky, Haim},
    month = nov,
    year = {2024},
    note = {arXiv:2411.07873},
}

@article{wang_unreasonable_2024,
    title = {The unreasonable effectiveness of gaussian score approximation for diffusion models and its applications},
    issn = {2835-8856},
    url = {https://openreview.net/forum?id=I0uknSHM2j},
    journal = {Transactions on Machine Learning Research},
    author = {Wang, Binxu and Vastola, John},
    year = {2024},
}

@misc{Han2025CanDiff,
  type = {Preprint},
  title = {Can {{Diffusion Models Learn Hidden Inter-Feature Rules Behind Images}}?},
  author = {Han, Yujin and Han, Andi and Huang, Wei and Lu, Chaochao and Zou, Difan},
  year = {2025},
  month = feb,
  number = {2502.04725v1},
  eprint = {2502.04725v1},
  publisher = {arXiv},
  urldate = {2025-08-30},
  archiveprefix = {arXiv}
}

@misc{Wang2025Analytical,
  type = {Preprint},
  title = {An {{Analytical Theory}} of {{Spectral Bias}} in the {{Learning Dynamics}} of {{Diffusion Models}}},
  author = {Wang, Binxu and Pehlevan, Cengiz},
  year = {2025},
  month = mar,
  number = {2503.03206v2},
  eprint = {2503.03206v2},
  publisher = {arXiv},
  urldate = {2025-08-30},
  archiveprefix = {arXiv}
}

@misc{Vastola2025Generalization,
  type = {Preprint},
  title = {Generalization through Variance: How Noise Shapes Inductive Biases in Diffusion Models},
  author = {J. Vastola, John},
  year = {2025},
  month = apr,
  number = {2504.12532v1},
  eprint = {2504.12532v1},
  publisher = {arXiv},
  urldate = {2025-08-30},
  archiveprefix = {arXiv}
}

@misc{2502_19499v1,
    type = {Preprint},
    title = {On the {Interpolation} {Effect} of {Score} {Smoothing}},
    url = {http://arxiv.org/abs/2502.19499v1},
    urldate = {2025-08-30},
    publisher = {arXiv},
    author = {Chen, Zhengdao},
    month = feb,
    year = {2025},
}

@article{canatar2021,
    title = {Spectral bias and task-model alignment explain generalization in kernel regression and infinitely wide neural networks},
    volume = {12},
    issn = {2041-1723},
    url = {https://doi.org/10.1038/s41467-021-23103-1},
    doi = {10.1038/s41467-021-23103-1},
    number = {1},
    journal = {Nature Communications},
    author = {Canatar, Abdulkadir and Bordelon, Blake and Pehlevan, Cengiz},
    month = may,
    year = {2021},
    pages = {2914},
}

@inproceedings{Kadkhodaie2024Generalization,
    title = {Generalization in diffusion models arises from geometry-adaptive harmonic representation},
    url = {https://openreview.net/forum?id=ANvmVS2Yr0},
    booktitle = {The twelfth international conference on learning representations},
    author = {Kadkhodaie, Zahra and Guth, Florentin and Simoncelli, Eero P and Mallat, Stéphane},
    year = {2024},
}
